\documentclass{SCIYA2023enOL}
\online
\begin{document}

\ensubject{fdsfd}

\ArticleType{ }
\Year{2022}
\Month{January}%
\Vol{65}
\No{1}
\BeginPage{1} %
\DOI{ }

\title[]{A method for escaping limit cycles in training GANs}
{A method for escaping limit cycles in training GANs}

\author[1,2]{Keke Li}{{20229002@cqnu.edu.cn, likeke135@163.com}}
\author[1,$\ast$]{Xinmin Yang}{{xmyang@cqnu.edu.cn}}

\AuthorMark{Keke Li}

\AuthorCitation{Keke Li, Xinmin Yang}

\address[1]{National Center for Applied Mathematics in 
	Chongqing, Chongqing Normal University, \\ Chongqing {\rm 
	401331}, 
	China}
\address[2]{School of Mathematical Sciences, University of Electronic 
Science and Technology of China, \\ Chengdu {\rm 611731}, Sichuan, 
China}

\abstract{This paper mainly conducts further research to alleviate 
the issue of limit cycling behavior in training generative 
adversarial networks (GANs) through the proposed predictive 
centripetal acceleration algorithm (PCAA). 
Specifically, we first derive the upper and lower bounds on the 
last-iterate convergence rates of PCAA for the general bilinear game, 
with the upper bound notably improving upon previous results. 
Then, we combine PCAA with the adaptive moment estimation algorithm 
(Adam) to propose PCAA-Adam, a practical approach for training 
GANs. 
Finally, we validate the effectiveness of the proposed 
algorithm through experiments conducted on bilinear games, 
multivariate Gaussian distributions, and the CelebA dataset, 
respectively.}



\maketitle

\section{Introduction}
As of now, generative adversarial networks (GANs) \cite{GAN2014} have 
emerged as a prominent and highly researched topic in the field of 
deep learning. They hold a crucial position, particularly when it 
comes to artificial intelligence content generation. 
Furthermore, they have wide 
applications in areas such as image generation and editing 
\cite{crowson2022vqgan}, video generation 
\cite{skorokhodov2022style}, 3D generation 
\cite{chan2022efficient, cai2022pix}, music generation 
\cite{Jesse2019gan}, and privacy protection \cite{QuZL2020}. For 
more  applications, please refer to 
\cite{Radford2015,Brock2018,Lvxs2021,Vondrick2016,FangH2020,
	AlawiehL2020,QuZL2020,yangyaxiang2020}.
The GANs framework  consists of two main components: the generator 
and the discriminator. 
The generator aims to generate samples that 
resemble real data, while the discriminator tries to distinguish 
between samples generated by the generator and real data. 
These two components compete with each other through adversarial 
training, ultimately aiming to generate realistic samples. 
The objective function for adversarial training in GANs 
can be formulated as follows:
\begin{equation*}  
	\min _{G} \max _{D} V(G, D) :=\mathbb{E}_{x \sim 
		\mathbb{P}_{\text {data 
	}}}\big[\log D(x)\big]+\mathbb{E}_{z \sim 
		\mathbb{P}_{\text{noise}}(z)}\big[\log 
	\big(1-D\big(G(z)\big)\big)\big],
\end{equation*}
where $G$, $D$, $x$ and $z$ represent the generator network, the 
discriminator network,  real data 
samples drawn from the data distribution $\mathbb{P}_{\text{data}}( 
x)$, and noise samples drawn from the noise 
distribution $\mathbb{P}_{\text{noise}}(z)$, respectively. 
The main goal of adversarial training is to find the optimal 
generator $G$ that minimizes the 
discriminator's ability to distinguish between real and generated 
samples, while the discriminator $D$ aims to maximize its ability to 
correctly classify between real and generated samples.
For a more detailed understanding of the objective function in GANs, 
please refer to \cite{GAN2014, goodfellow2016nips, 
wang2020mathematical}. 

Although GANs have many advantages 
(see \cite{GAN2014, goodfellow2016nips}) and widespread practical 
applications (see \cite{Farajzadeh2022gan}), they still face numerous 
challenges both in theory and practical applications (see 
\cite{goodfellow2016nips, odena2019open, Saxena2021generative, 
Chae2023open}). One of the most significant challenges is that GANs 
are hard to train (see \cite{goodfellow2016nips, Leina2020, 
grnarova2021generative, Chae2023two, Chae2023open}), and standard 
optimization methods often fail to converge to reasonable solution 
(see \cite{mertikopoulos2019optimistic}). 
And it has been pointed out in \cite{daskalakis2018training, 
	mertikopoulos2019optimistic} that for practical training of GANs, 
changes in algorithm outputs are needed for corresponding analysis, 
and the usual average-iterate convergence is no longer applicable. 
Therefore, it is necessary to consider the last-iterate convergence 
of algorithms.
However, training GANs using gradient descent ascent (GDA) often 
shows oscillations and is susceptible to limit cycling behavior 
in the context of last-iterate convergence (see 
\cite{daskalakis2018training, peng2020training, Berard2020A, 
shen2020learning, pethick2023open}).
Moreover, Lin and Jordan \cite{lin2020gradient} found that for 
convex-concave 
games, although \cite{nedic2009subgradient} established the 
average-iterate convergence of GDA, GDA may converge to a limit 
cycle or even diverge in general in the context of last-iterate 
convergence.
Therefore, the training of GANs has been described as ``notoriously 
hard" in  
\cite{mescheder2017numerics, chavdarova2021taming, 
grnarova2021generative}, highlighting the difficulties associated 
with this task. In response to these challenges, researchers have 
recently proposed new algorithms based on GDA to tackle issues 
encountered during GANs training, including 
problems like oscillations and limit cycling behavior 
\cite{daskalakis2018training, peng2020training, Li2023training}. 
Especially, Heish \cite{hsieh2020convergence} identified how to 
utilize optimization techniques to eliminate limit cycling behavior 
in minimax games as an open question, which has attracted researchers 
to further investigate.
Indeed, recently there have been some advances in proposing methods 
based on GDA to alleviate the limit cycling behavior in training  
GANs.
In particular, Daskalakis et al. \cite{daskalakis2018training} were 
the first to attempt to address the issue of limit cycling behavior 
in training  GANs by introducing optimistic gradient descent ascent 
(OGDA). Initially, they analyzed the convergence rate of OGDA on the 
the bilinear game and later integrated OGDA with Adam 
\cite{kingma2014adam} to propose the Optimistic-Adam algorithm for 
practical training of GANs. Their numerical experiments with 
Wasserstein GAN \cite{arjovsky2017wasserstein} illustrated the 
superior performance of Optimistic-Adam compared to Adam.
Gidel et al. \cite{gidel2019a} initially demonstrated that OGDA is 
equivalent to storing and re-using the extrapolated gradient for 
extrapolation. They then employed variational inequality tools and, 
assuming monotonicity and compactness and convexity, established the 
convergence of their proposed ``extrapolation from past" algorithm in 
the average-iteration sense. Finally, they introduced an Adam variant 
incorporating extrapolation steps for the training of GANs.
Mertikopoulos et al. \cite{mertikopoulos2019optimistic} emphasized 
that adversarial learning, especially concerning GANs, still lacks a 
comprehensive understanding, and training GANs is widely known to be 
challenging. 
Building on the research in 
\cite{daskalakis2018training,gidel2019a}, Mertikopoulos et al. 
\cite{mertikopoulos2019optimistic} interpreted the 
optimistic method 
as an ``extra gradient" step, extrapolating the training process 
along the current gradient. They established a convergence theory for 
OGDA, considering dependency and Lipschitz assumptions, and further 
introduced an alternative version of Optimistic-Adam, showcasing its 
effectiveness through experiments with DCGAN (see \cite{Radford2015}).
Moreover, Peng et al. \cite{peng2020training} highlighted that 
training GANs often encounters cyclic behaviors in the iterations. 
Drawing an analogy with uniform circular motion, they proposed two 
algorithms, simultaneous centripetal acceleration (Grad-SCA) and 
alternating centripetal acceleration (Grad-ACA), to mitigate the 
occurrence of limit cycling behavior. Under suitable assumptions, 
they derived linear convergence rates for Grad-SCA and Grad-ACA in 
the bilinear game setting. Additionally, they validated the 
effectiveness of their proposed RMSprop-ACA algorithm through 
experiments involving Gaussian kernel learning with GANs.
Given the current state of research on GANs training algorithms, 
directly studying the convergence of training algorithms for GANs is 
extremely challenging \cite{odena2019open, Saxena2021generative, 
salimans2016improved}. 
Therefore, proposing new algorithms that guarantee convergence on 
simplified GANs models and using them to improve the quality of 
generated samples in practical applications is of great significance 
\cite{daskalakis2018training, ryu2019ode, liang2019interaction, 
gidel2019a, mertikopoulos2019optimistic, peng2020training, 
chavdarova2021taming, he2021age}. 
Taking this into consideration, Li et al. \cite{Li2023training} 
built upon the open question presented in \cite{hsieh2020convergence} 
and carried out preliminary investigations in conjunction with GANs. 
They introduced the predictive centripetal acceleration (PCA) 
algorithm, which is based on an intuitive geometric interpretation 
of the approximate centripetal hypothesis. 
The algorithm demonstrated last-iterate linear convergence on the 
bilinear game with full-rank square matrices.
Additionally, they integrated this algorithm with Adam to propose the 
PCA-Adam algorithm for the practical training of GANs.

In spite of Li et al. \cite{Li2023training} has obtained late-iterate 
linear convergence results for the bilinear game with the matrix $A$ 
is a full-rank square matrix, we are curious about the applicability 
of the algorithm proposed in \cite{Li2023training} to more general 
bilinear game, where the matrix $A$ is not necessarily square.
Moreover, we are not satisfied with the linear convergence results 
achieved in \cite{Li2023training}, and we wonder if these results can 
be improved. 
Furthermore, in order to adapt the algorithms proposed for specific 
GANs to be used for training general GANs, \cite{Li2023training} 
combines the proposed algorithm with Adam  
and introduces PCA-Adam. 
However, we have noticed that the choice of the exponent for the 
exponential decay rate in PCA-Adam does not make it evident how 
PCA-Adam can straightforwardly reduce to Extra-Adam 
\cite{mertikopoulos2019optimistic} and Adam.
Therefore, with this purpose in mind, we aim to further investigate 
and explore 
the following questions:  
{\it
Q1: Does GDA also diverge when used to solve the general 
bilinear game?
Q2: Does PCAA also converge for the general bilinear game? If 
it converges, where does the algorithm converge to? 
Q3: How fast is the convergence rate? Has the convergence 
rate improved compared to the convergence rate in related literature?
Q4: How can we combine the proposed algorithm with Adam in 
such a way that the resulting new algorithm can simultaneously reduce 
to both Extra-Adam and Adam? What are the actual effects of the new 
Adam algorithm proposed in comparison to these two algorithms?}

Addressing these questions both theoretically and experimentally 
constitutes the main research work of this paper.
Specifically, firstly, we obtained the result of the PPCA iteration 
format with a zero projection term for the general bilinear game, 
which geometrically illustrates the divergence of GDA when used to 
solve the general bilinear game. Furthermore, based on this result, 
we can 
derive the iteration format for PCAA. 
Secondly, we theoretically proved that if PCAA   
converges for the general bilinear game in last-iterate sense, the 
iteration sequence will eventually converge to Nash equilibrium 
point.
Then, we demonstrated that PCAA exhibits linear convergence with a 
complexity of $\mathcal{O}(\kappa \log(1/\epsilon))$ for the general 
bilinear game, which improves upon and refines the complexity result 
of $\mathcal{O}(\kappa^{-2} \log(1/\epsilon))$ obtained in 
\cite{liang2019interaction} and \cite{Li2023training} for the 
bilinear game with a full-rank matrix.
Moreover, we want to 
emphasize that although the complexity result of PCAA is similar to 
the literatures \cite{mishchenko2020revisiting, mokhtari2020unified}, 
the linear convergence results we obtained is superior to the results 
obtained in \cite{mishchenko2020revisiting} and 
\cite{mokhtari2020unified}. 
Furthermore, we have obtained a lower complexity bound of PCAA for 
the general bilinear game.
Finally, we directly combined PCAA with Adam and used the same 
exponential decay rate for both prediction and update steps in 
PCAA-Adam. This ensures that PCAA-Adam can easily reduce to 
Extra-Adam and Adam when certain parameter values are chosen. 
Furthermore, numerical experiments on the CelebA dataset validated 
the effectiveness of  PCAA-Adam for training GANs compared to 
Extra-Adam and Adam. 
 
\subsection*{Contributions}

\begin{itemize}
	\item  We demonstrate its last-iterate linear convergence on the
	general bilinear game and provide examples to showcase its 
	effectiveness in alleviating cyclic behavior. Additionally, we 
	provided a lower complexity bound of PCAA for solving the general 
	bilinear game.
	
	\item We have identified the reasons for the divergence of GDA in 
	solving the general bilinear game from a geometrically intuitive 
	perspective. Additionally, we analyzed the reasons for 
	non-convergence from the perspective of convergence theory.
	
	\item We combined the proposed algorithm PCAA with Anderson 
	mixing and introduced PCAA-AM to accelerate the computational 
	efficiency of PCAA in solving the general bilinear game. 
	
	\item We combined the proposed algorithm PCAA with Adam and 
	introduced PCAA-Adam for practical training of GANs. Experimental 
	results on the CelebA dataset illustrated that PCAA-Adam 
	outperforms the comparison algorithms such as Adam and Extra-Adam 
	in training GANs.
\end{itemize}

\section{Preliminaries}\label{section2}

Throughout this paper, unless otherwise specified, let 
${\prod}_{(\theta)}{(\phi)} ,\; \theta, \phi 
\in \mathbb{R}^n$ denote the projection of  vector $\phi$ onto  
vector  $\theta$, $\nabla V( \cdot)$ denote the gradient of function 
$V$, respectively. 
A  two-player zero-sum game is a game between two players, with one 
player's payoff function is denoted by $V$ and the other player's 
payoff function is $-V$, where $V: \Theta \times \Phi  \rightarrow 
\mathbb{R}$ with $\Theta \times \Phi  \subseteq \mathbb{R}^{m} \times 
\mathbb{R}^{n}$. 
The function $V$ maps the actions took by both players $(\theta, 
\phi) \in \Theta \times \Phi$ to a real value, representing the gain 
of $\phi$-player and the loss of $\theta$-player. 
We call $\phi$ player, who tries to maximize the 
payoff function $V$,the max-player, and $\theta$-player the 
min-player. In the most classical setting, a two-player zero-sum game 
has the following form:
\begin{equation*}  
	\min _{\theta \in \Theta} \max _{\phi \in \Phi} V(\theta, \phi).
\end{equation*}
In this paper, we focus on the differentiable two-player game, i.e.,  
payoff functions $V(\cdot, \cdot)$ is differentiable on  $\Theta 
\times \Phi \subseteq \mathbb{R}^{m} \times \mathbb{R}^{n}$.

\begin{definition}[See \cite{jin2020local}]\label{D2.1} 
	Point $(\theta^{\star}, \phi^{\star})$ is a  saddle point of $V$, 
	if for any $(\theta, \phi) \in \Theta \times \Phi$, we have
	\begin{equation*} 
		V(\theta^{\star}, \phi) \leq V(\theta^{\star}, \phi^{\star}) 
		\leq V(\theta, \phi^{\star}).
	\end{equation*}
\end{definition}

\begin{definition}[{See \cite{jin2020local}}]\label{DD2.2}
	Point $(\theta^*, \phi^*)$ is referred to as a local saddle point 
	(or local Nash equilibrium point) of the function $V(\cdot, 
	\cdot)$ if there exists $\delta > 0$ such that for any $(\theta, 
	\phi) \in \Theta \times \Phi$ satisfying $\|\theta - \theta^*\|_2 
	\leq 
	\delta$ and $\|\phi - \phi^*\|_2 \leq \delta$, the following 
	inequality holds:
	\begin{equation*}\label{eq3.1.1}
		V(\theta^*, \phi) \leq V(\theta^*, \phi^*) \leq V(\theta, 
		\phi^*).
	\end{equation*}
\end{definition}

\begin{definition}[See \cite{abernethy2021last}]\label{def1}
	Point $(\theta, \phi) \in \Theta \times \Phi 
	$ is a critical point of the differentiable  $V$, if $\nabla 
	V(\theta, \phi) = 0$.
\end{definition}

\section{Predictive centripetal acceleration algorithm}

\subsection{Related algorithm}

\begin{figure}[h!] 
	\subfloat[$V(\theta,\phi)=\theta \phi + 
	\epsilon(\theta^2 - \phi^2)$]{\includegraphics[width=220 pt, 
	height =200 pt]{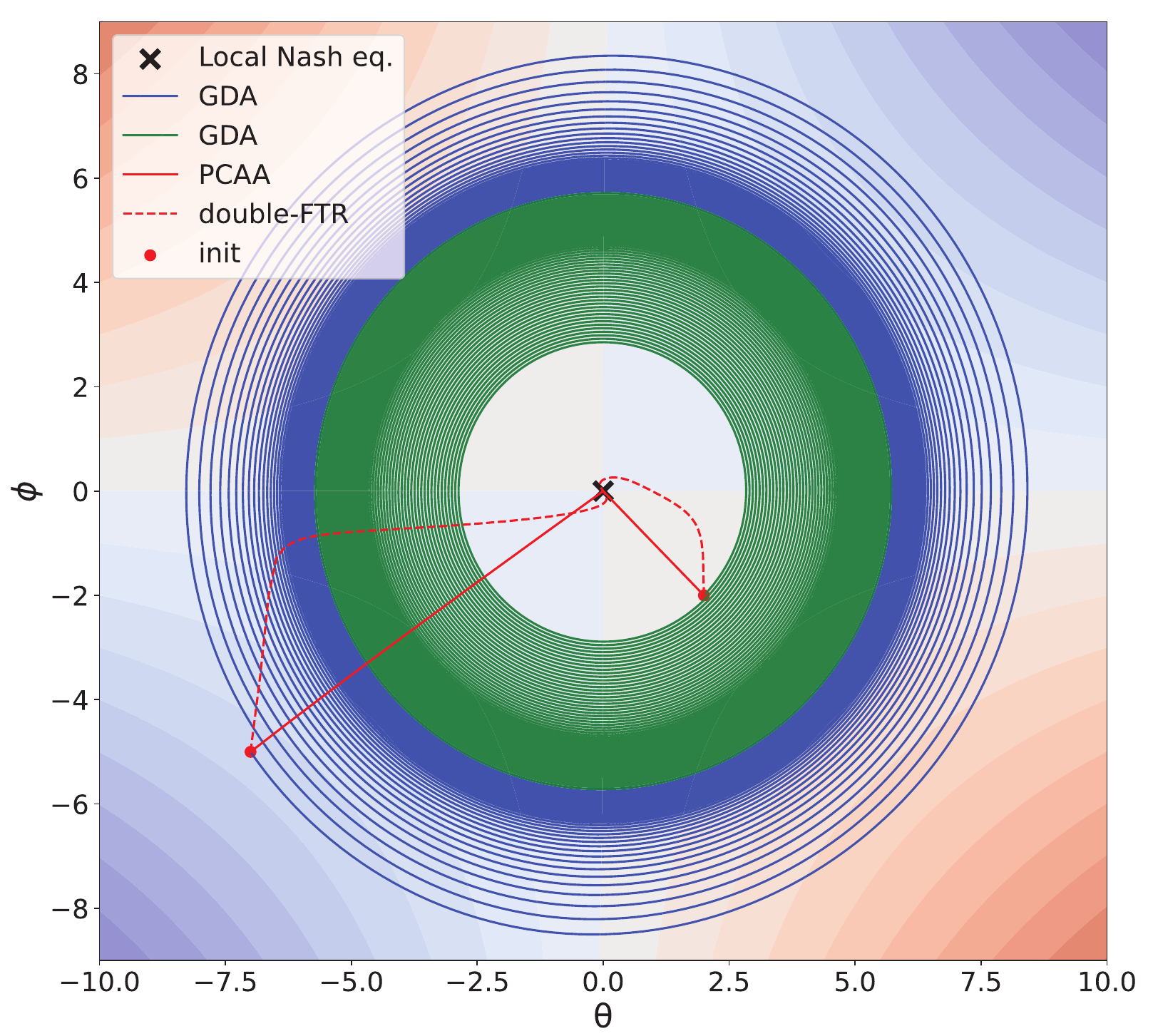} } 
	\subfloat[$V(\theta,\phi)= 2 \theta^2-50\theta 
	\phi- \phi^2$]{ \includegraphics[width=220 pt, 
		height =200 pt]{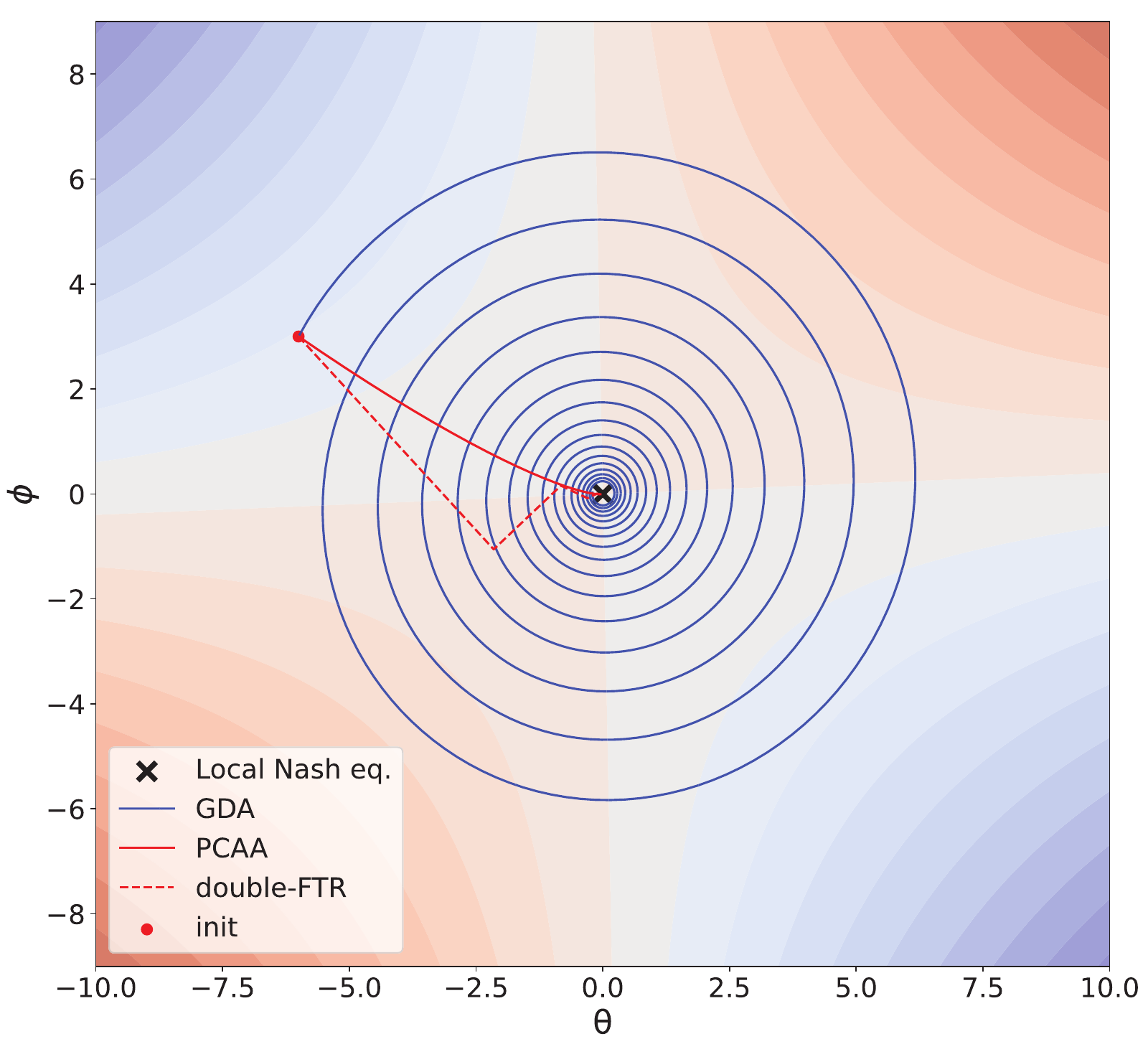}}
	\caption{Two examples of issues with GDA in finding Nash 
		equilibrium in zero-sum games, where the double-FTR algorithm 
		can 
		be found in \cite{bao2022finding}. (a) GDA 
		converges
		to a limit cycle (we use $\epsilon = 0.0001,\gamma = \beta= 
		0.01, 
		\alpha =0  $) instead of the unique Nash equilibrium $(0, 
		0)$. 
		(b) The convergence process of GDA exhibits oscillations, 
		manifested as a slow spiral convergence towards the Nash 
		equilibrium.}
	\label{exPPCA1}
\end{figure}

A natural approach for players in a differentiable zero-sum game to 
find a local Nash equilibrium is to use gradient-based search 
algorithms. The simplest gradient-based algorithm for solving 
differentiable zero-sum games is the gradient descent 
ascent (GDA) \cite{mescheder2017numerics, peng2020training, 
	bailey2020finite}  and its variants such as alternating gradient 
descent ascent (AGDA)  \cite{mescheder2017numerics, peng2020training, 
	bailey2020finite}  algorithm, 
extra gradient algorithm (EG) 
\cite{korpelevich1976extragradient}, 
and optimistic gradient descent ascent algorithm (OGDA) 
\cite{daskalakis2018training}, among others. These types of 
algorithms aim to iteratively update the players' strategies based on 
gradients, with the goal of converging to a local Nash equilibrium. 
First, let's review the iterative formats of GDA and AGDA, where 
$\alpha$ is the step size parameter.

 \begin{equation}\label{GDA}
	\begin{aligned}  
		\theta_{t+1}  =\theta_{t}-\alpha \nabla_{\theta} 
		V(\theta_{t}, 
		\phi_{t}),  	\quad	\phi_{t+1}  =\phi_{t}+\alpha 
		\nabla_{\phi}V(\theta_{t}, \phi_{t}). 
	\end{aligned}\tag{GDA}
\end{equation}
\begin{equation}\label{AGDA}
	\begin{aligned} 
		\theta_{t+1}  =\theta_{t}-\alpha \nabla_{\theta} 
		V(\theta_{t}, 
		\phi_{t}), \quad 		\phi_{t+1}  =\phi_{t}+\alpha 
		\nabla_{\phi} V(\theta_{t+1}, \phi_{t}). 
	\end{aligned} \tag{AGDA}
\end{equation}
In the following, we review a variant of EG called 
modified predictive method, proposed by Liang and Stokes in
\cite{liang2019interaction}, which allows for simultaneous gradient 
updates. In this variant, referred to as MPM in this article, the 
parameters $\gamma$ and $\beta$ represent the step sizes for the 
prediction step and the update step, respectively.
\begin{equation}\label{MPM}
	\begin{aligned}
		&\mathrm{Predictive ~ step:} ~
		\theta_{t+1 / 2}  =\theta_{t}-\gamma \nabla_{\theta} 
		V(\theta_{t}, \phi_{t}), \qquad
		\phi_{t+1 / 2}  =\phi_{t}+\gamma \nabla_{\phi} 
		V(\theta_{t}, \phi_{t});  \\
		&\mathrm{gradient ~ step:}~
		\theta_{t+1}  =\theta_{t}-\beta \nabla_{\theta} 
		V(\theta_{t+1 / 2}, \phi_{t+1 / 2}); \quad
		\phi_{t+1}  =\phi_{t}+\beta \nabla_{\phi} V(\theta_{t+1 
			/ 2}, \phi_{t+1 / 2}).
	\end{aligned} \tag{MPM}
\end{equation}
Clearly, in algorithm \eqref{MPM}, if $\gamma = \beta$, then MPM 
reduces to  EG,  which was proposed 
by Korpelevich in \cite{korpelevich1976extragradient}.
Then, we recall a variant of OGDA called simultaneous centripetal 
Acceleration (Grad-SCA) and its alternating version Grad-ACA, which 
were proposed by Peng and Dai et al. in \cite{peng2020training}, 
aiming to alleviate the issue of limit 
cycling behavior in training GANs. The parameters $\alpha_1, 
\alpha_2, 
\beta_1, \beta_2$ are step size-related parameters. Obviously, in 
Grad-SCA, when the parameters 
$\alpha_1=\alpha_2=\beta_1=\beta_2$,  Grad-SCA  reduces 
to OGDA. 
\begin{equation}\label{Grad-SCA}
	\begin{aligned} 
		&G_{\theta} =\nabla_{\theta} V(\theta_{t}, 
		\phi_{t} )+\frac{\beta_{1}}{\alpha_{1}}\big(\nabla_{\theta}
		V (\theta_{t}, \phi_{t} )-\nabla_{\theta} 
		V (\theta_{t-1}, \phi_{t-1} )\big),  
		&\theta_{t+1}  =\theta_{t}-\alpha_{1} G_{\theta}, \\ 
		&G_{\phi} =\nabla_{\phi} V (\theta_{t}, 
		\phi_{t} )+\frac{\beta_{2}}{\alpha_{2}}\big(\nabla_{\phi}
		V (\theta_{t}, \phi_{t} )-\nabla_{\phi} 
		V (\theta_{t-1}, \phi_{t-1} )\big),  
		&\phi_{t+1} =\phi_{t}+\alpha_{2} G_{\phi}.
	\end{aligned} \tag{Grad-SCA}
\end{equation}
\begin{equation}\label{Grad-ACA}
	\begin{aligned} 
		&G_{\theta} =\nabla_{\theta} V (\theta_{t}, 
		\phi_{t} )+\frac{\beta_{1}}{\alpha_{1}}\big(\nabla_{\theta}
		V (\theta_{t}, \phi_{t} )-\nabla_{\theta} 
		V (\theta_{t-1}, \phi_{t-1} )\big),  
		&\theta_{t+1}  =\theta_{t}-\alpha_{1} G_{\theta}, \\ 
		&G_{\phi} =\nabla_{\phi} V (\theta_{t+1}, 
		\phi_{t} )+\frac{\beta_{2}}{\alpha_{2}}\big(\nabla_{\phi}
		V (\theta_{t+1}, \phi_{t} )-\nabla_{\phi} 
		V (\theta_{t}, \phi_{t-1} )\big), 
		&\phi_{t+1} =\phi_{t}+\alpha_{2} G_{\phi}. 
	\end{aligned} \tag{Grad-ACA}
\end{equation}
Finally, we present an example to visually demonstrate the 
oscillation and limit cycle problems that exist when using GDA to 
solve two-player zero-sum games. Please refer to Figure \ref{exPPCA1} 
for details.

In response to the issue of limit cycling behavior and inspired by 
the work in \cite{daskalakis2018training}, Peng et al. 
\cite{peng2020training} proposed the SCA and ACA methods to alleviate 
such behavior. Especially, 
they provided an explanation of the fundamental intuition behind 
centripetal acceleration in \cite[Figure 1]{peng2020training}. 
First, they considered uniform circular motion, where $\nabla V_t$ 
represents the instantaneous velocity at time $t$. The centripetal 
acceleration $\lim_{\delta t \rightarrow 
	0} (\nabla V_{t+\delta t}-\nabla V_{t} ) / \delta t$ 
	points 
towards the origin. 
And they argued that the cyclic behavior around a Nash 
equilibrium may resemble circular motion around the origin. 
Then, they proposed that the centripetal acceleration provides a 
direction along which the iterates can approach the target more 
quickly. 
Finally, they utilized $(\nabla V_{t+\delta t}-\nabla V_{t})$, 
referred to as the approximate centripetal acceleration term, to 
approximate $\lim_{\delta t \rightarrow 0}(\nabla V_{t+\delta 
	t}-\nabla V_{t}) / \delta t$, and this approximated term is 
applied to modify GDA resulting in Grad-SCA.

A natural idea of this work and \cite{Li2023training} is to explore 
how to understand the approximate centripetal acceleration method 
geometrically and how to construct a direction that directly points 
to the origin to alleviate limit cycling behavior, instead of 
employing the approximated centripetal acceleration term?

Motivated by the above question and inspired by OGDA, Grad-SCA, MPM 
and lookahead methods \cite{chavdarova2021taming,zhang2019lookahead}, 
we also considered the uniform circular motion, as shown in Figure 
\ref{figPPCA} below. Let $\nabla V_t$ denote the instantaneous 
velocity at time $t$. Firstly, we performed a prediction step at 
$t+1/2$, and obtain the instantaneous velocity $\nabla V_{t+1/2}$ at 
that time. Then, we computed the approximated centripetal 
acceleration term $ (\nabla V_{t+1/2}-\nabla V_{t} )$ at time $t$.  
Finally, we projected the approximated centripetal acceleration term 
$ (\nabla V_{t+1/2}-\nabla V_{t} )$ at time $t$ onto $\nabla V_t$, 
the instantaneous velocity at time $t$. By doing so, we obtained 
the projection centripetal acceleration term ${\prod}_{\big(\nabla 
	V (\theta_{t}, \phi_{t} ) \big)}{\big(\nabla 
	V (\theta_{t+1 / 2}, \phi_{t+1 / 2} ) - \nabla  
	V (\theta_{t}, \phi_{t} ) \big) }$, which precisely points to the 
	origin. For more details, see Figure \ref{figPPCA}. 
We also argued that the
cyclic behavior around a Nash equilibrium might be similar to the 
circular motion around the origin. Therefore, the projection 
centripetal acceleration term 
provides a direction, along which the
iterates can approach the target directly.  Then the projection 
centripetal acceleration term is also applied to modify GDA.

\begin{figure}[h!]  
	\subfloat[]{
			\centering\includegraphics[width=215 pt, height =200 
			pt]{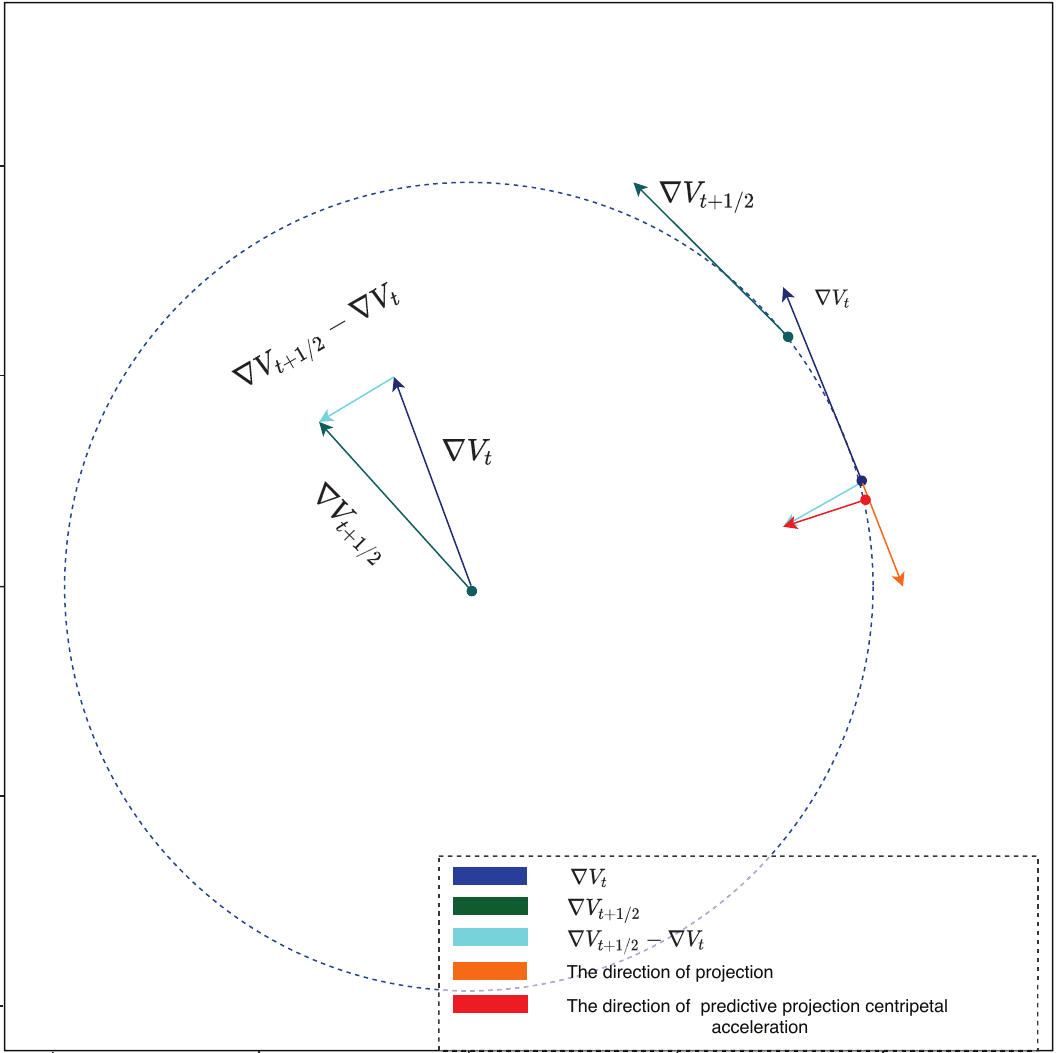}
			\label{subfig-ppca1}} \qquad 
	\subfloat[]{
			\centering\includegraphics[width=205 pt, height =200 
			pt]{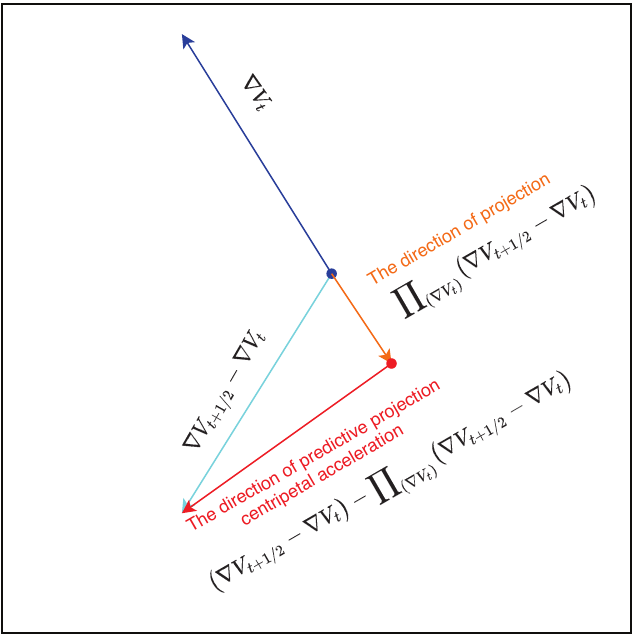}
			\label{subfig-ppca2} }
	\caption{(a): The basic intuition of predictive projection 
		centripetal acceleration methods in this work and 
		\cite{Li2023training}. 
		(b): It provides a magnified view and description of the 
		projection term $\prod_{ (\nabla V_{t} 
			 )}{ ( \nabla 
			V_{t+1/2}-\nabla V_{t}  )}$,  which represents the 
			projection of 
			$\nabla 
		V_{t+1/2}-\nabla V_{t}$ onto $\nabla V_{t}$. Additionally, it 
		shows  the projection 
		centripetal acceleration term $ ( \nabla V_{t+1/2}-\nabla 
		V_{t}  ) - \prod_{ (\nabla V_{t}  )}{ ( 
		\nabla 
			V_{t+1/2}-\nabla V_{t}  )} $ in subfigure (a).}
	\label{figPPCA}
\end{figure}  
Finally, we obtained the predictive projection centripetal 
acceleration algorithm (PPCA) in \cite{Li2023training} with the 
following form :
\begin{equation}\label{eqppca}
	\begin{aligned}
		&{\mathrm{Predictive~step:}}~
		\theta_{t+1 / 2} =\theta_{t}-\gamma \nabla_{\theta} 
		V (\theta_{t}, \phi_{t} ), 
		\phi_{t+1 / 2} =\phi_{t}+\gamma \nabla_{\phi} 
		V (\theta_{t}, \phi_{t} ); \\
		&{\mathrm{gradient~step:}}~
		\bigg(
		\begin{array}{c}
			\theta_{t+1}  \\
			\phi_{t+1}
		\end{array}
		\bigg)
		=\bigg(
		\begin{array}{c}
			\theta_{t}  \\
			\phi_{t}
		\end{array}
		\bigg) +\alpha \nabla\bar{V} (\theta_{t}, 
		\phi_{t} )+  \beta \bigg(\big(\nabla 
		\bar{V} (\theta_{t+1 / 2}, \phi_{t+1 / 2} ) -  
		\nabla \bar{V} (\theta_t, \phi_t ) \big) \\ 
		&{\qquad\qquad\qquad\qquad\qquad\qquad\qquad\qquad }-
		{\prod}_{\big(\nabla\bar{V} (\theta_{t}, 
			\phi_{t} )\big)}{\big(\nabla 
			\bar{V} (\theta_{t+1 
				/ 2}, \phi_{t+1 / 2} ) - \nabla  
			\bar{V} (\theta_{t}, \phi_{t} ) \big) }\bigg),
	\end{aligned}  
\end{equation}
where $\nabla\bar{V} (\theta_{t}, \phi_{t} ) := \bigg(
\begin{array}{c}
	-\nabla{V}_{\theta} (\theta_{t}, \phi_{t} )  \\
	\nabla{V}_{\phi} (\theta_{t}, \phi_{t} )
\end{array} \bigg)$, and ${\prod}_{\big( \nabla 
	\bar{V} (\theta_{t}, \phi_{t} ) \big)}{\big(\nabla 
	\bar{V} (\theta_{t+1 / 2}, \phi_{t+1 / 2} ) - \nabla  
	\bar{V} (\theta_{t}, \phi_{t} ) \big) } $ denotes 
the  projection of vector ${ \nabla \bar{V} (\theta_{t+1 / 
		2}, \phi_{t+1 / 2} ) - \nabla  \bar{V} (\theta_{t}, 
	\phi_{t} )  }$ onto vector ${ \nabla 
	\bar{V} (\theta_{t}, \phi_{t} ) }$.

\subsection{Predictive centripetal acceleration algorithm}

The study on the bilinear game  can reveal the fundamental issues of 
gradients in games \cite{balduzzi2018mechanics}. Moreover, the 
bilinear game is often used as a simple model for theoretical 
discussions on the reasons behind problems in training GANs 
\cite{azizian2020tight}, and they serve as important illustrative 
examples for analyzing and understanding new algorithms and 
techniques \cite{daskalakis2018training, gidel2020multi, gidel2019a, 
gidel2019negative,liang2019interaction, zhang2020convergence}. In 
fact, the bilinear game can reveal the challenges that GDA encounters 
in training GANs, and it can represent some simple forms of GANs 
\cite{pinetz2018optimized, daskalakis2018training, gidel2019a, 
mescheder2018training}.  
Moreover, since Goodfellow \cite{goodfellow2016nips} analyzed the 
circular trajectories of GDA on the two-dimensional bilinear game in 
2016, almost all research on proposing new algorithms based on GDA 
has discussed the bilinear game to demonstrate the effectiveness of 
the proposed algorithms. 
In addition to the references mentioned above, recent research papers 
include \cite{gidel2017frank,  daskalakis2018limit,  gidel2019a, 
zhang2020convergence, lorraine2021complex, anagnostides2020solving}, 
among others. 
Especially, following the research approach of the aforementioned 
papers, Li et al. \cite{Li2023training} discussed the properties of 
PPCA under the condition that the matrix is a full-rank square 
matrix. However, for a more general scenario where the matrix is not 
necessarily square, this aspect has not been explored yet. Therefore, 
the main focus of this paper is to investigate the properties of PPCA 
in the context of the bilinear game with a matrix that is not 
necessarily square.

In this section, we first discuss a property of algorithm 
\eqref{eqppca} on the 
following bilinear game. This property is helpful for understanding 
the performance of GDA in game and simplifying the iterative format 
of algorithm \eqref{eqppca}.
\begin{equation}\label{eq3.1}
	\min _{\theta \in \mathbb{R}^{m}} \max _{\phi \in \mathbb{R}^{n}} 
	\theta^{\mathrm{T}} A \phi+\theta^{\mathrm{T}} b+c^{\mathrm{T}} 
	\phi, \quad A \in \mathbb{R}^{m \times n}, \quad b \in 
	\mathbb{R}^{m}, \, c \in 
	\mathbb{R}^{n},
\end{equation}
where $ \theta \in \mathbb{R}^{m}, \, \phi \in \mathbb{R}^{n}$ 
implies that $\theta$-player and $\phi$-player  have different 
decision spaces. 
By the first order conditions 
of the game defined by \eqref{eq3.1}, we have 
\begin{equation*} 
	\begin{array}{rc}
		A \phi^{'}+b &=0, \\
		A^{\mathrm{T}} \theta^{'}+c &=0,
	\end{array}
\end{equation*}
where $ ( \theta^{'}, \phi^{'} )$ is a critical point of  
game \eqref{eq3.1}. If $b$ (resp. $c$) does not belong to the column 
space of $A$ (resp. $A^{\mathrm{T}}$), then game \eqref{eq3.1} 
has no Nash  equilibrium.  
In the following, drawing inspiration from the approach outlined in 
\cite{mishchenko2020revisiting, gidel2019negative, 
	zhang2020convergence}, we also make the assumption that a Nash 
equilibrium 
exists for this game. As a result,  there exist $\bar{\phi} $ and 
$\bar{\theta} $
such that $b =- A\bar{\phi} $ and $c =-A^{\mathrm{T}} \bar{\theta} $.
Then, we 
transform $ (\theta, \phi )$ to $ (\theta - \bar{\theta}, 
\phi - \bar{\phi} )$, thus game \eqref{eq3.1} is reformulated as:
\begin{equation}\label{eqbilinear}
	\min _{\theta \in \mathbb{R}^{m}} \max _{\phi \in \mathbb{R}^{n}} 
	V (\theta, \phi  ) = \theta^{\mathrm{T}} A \phi, \quad A 
	\in \mathbb{R}^{m\times n}.
\end{equation} 

Then, we present the following proposition, which is crucial for 
deriving our main algorithm and proving its convergence on 
zero-sum game \eqref{eqbilinear}.

\begin{proposition}\label{proposition1}
	If $A$ is a non-zero matrix in bilinear game \eqref{eqbilinear}, 
	and $\nabla\bar{V} (\theta_{t}, \phi_{t} ) \neq 0 $, then 
	the projection term of the iterative format for PPCA in 
	\eqref{eqppca} is zero. 
\end{proposition}

The proof is similar to Lemma 3.1 in 
\cite{Li2023training}, so it is omitted here.

Proposition \ref{proposition1} states that for game 
\eqref{eqbilinear} with 
non-zero matrix $A$,  the
approximate centripetal acceleration term $\nabla 
\bar{V} (\theta_{t+1/2}, \phi_{t+1/2} ) - \nabla 
\bar{V} (\theta_{t}, \phi_{t} )$ is orthogonal to $\nabla 
\bar{V} (\theta_{t}, \phi_{t} )$. Therefore, the approximate 
centripetal acceleration term $\nabla \bar{V} (\theta_{t+1/2}, 
\phi_{t+1/2} ) - \nabla \bar{V} (\theta_{t}, 
\phi_{t} )$ directly points towards the center's direction 
at time $t$ with 
any non-zero predictive step size $\gamma$. Therefore, combining 
Proposition 1 in \cite{daskalakis2018training}, we have the 
following remark.

\begin{remark}\label{remark3.1}
	\cite[Proposition 1]{daskalakis2018training} states that when 
	using GDA to solve bilinear game $\min_{\theta \in 
		\mathbb{R}^{n}} \max_{\phi \in \mathbb{R}^{n}}$ $ 
	\theta^{\mathrm{T}} \phi$, for any initial point $(\theta_0, 
	\phi_0)$ satisfying $\theta_0, \phi_0 \neq 0$, the iteration 
	process diverges. Motivated by \cite[Proposition 
	1]{daskalakis2018training}, this paper raises 
	the following questions: ``How to geometrically understand the
	Proposition 1? And for game \eqref{eqbilinear} with a non-zero 
	matrix 
	$A \in \mathbb{R}^{m \times n}$, does GDA  also diverge 
	in terms of last-iterate sense?" 
	
	In fact, Proposition \ref{proposition1} answers the above two 
	questions. 
	According to Proposition \ref{proposition1}, $\nabla
	\bar{V} (\theta_{t},  \phi_{t} )$, $\nabla
	\bar{V} (\theta_{t+1 / 2}, \phi_{t+1 / 2} )$, and 
	$\nabla 
	\bar{V} (\theta_{t+1 / 2}, \phi_{t+1 / 2} ) -
	\nabla \bar{V} (\theta_{t}, \phi_{t} )$ form a 
	right-angled 
	triangle with $\nabla
	\bar{V} (\theta_{t}, \phi_{t} )$ and  $\nabla
	\bar{V} (\theta_{t+1 / 2}, \phi_{t+1 / 2} )- \nabla
	\bar{V} (\theta_{t}, \phi_{t} )$   as the perpendicular 
	sides, 
	and $\nabla
	\bar{V} (\theta_{t+1 / 2}, \phi_{t+1 / 2} )$ as the 
	hypotenuse. This means that when using GDA to solve the game 
	mentioned above, after each iteration, the magnitude of gradient 
	$\nabla \bar{V} (\theta_{t+1}, \phi_{t+1} )$ at the next 
	iteration point continuously increases relative to gradient 
	$\nabla 
	\bar{V} (\theta_{t}, \phi_{t} )$ at the current step.
	Therefore, GDA diverges on bilinear game \eqref{eqbilinear}. 
	Theorem \ref{proposition1} geometrically explains Proposition 1 
	in \cite{daskalakis2018training}. Simultaneously, 
	Theorem \ref{proposition1} 
	provides a geometric interpretation of the fact that when using 
	GDA 
	to solve game $\min_{\theta \in \mathbb{R}^{m}} \max_{\phi \in 
		\mathbb{R}^{n}} 
	\theta^{\mathrm{T}} A \phi$, with any initial point $(\theta_0, 
	\phi_0)$ satisfying $\theta_0, \phi_0 \neq 0$, it diverges, where 
	$A$ 
	is a non-zero matrix.
\end{remark}

Moreover, from Proposition \ref{proposition1}, we see that the 
projection 
term of in algorithm \eqref{eqppca}  is zero. Thus, algorithm 
\eqref{eqppca} has the following simple form on bilinear game
\eqref{eqbilinear}.
\begin{equation}\label{eqpcaa}
	\begin{split}
		{\rm Predictive ~ step:}  \quad
		\theta_{t+1 / 2} &=\theta_{t}-\gamma \nabla_{\theta} 
		V (\theta_{t}, \phi_{t} ), \\
		\phi_{t+1 / 2} &=\phi_{t}+\gamma \nabla_{\phi} 
		V (\theta_{t}, \phi_{t} ); \\
		{\rm gradient ~ step:}   \quad
		\theta_{t+1} &= \theta_{t} - \alpha \nabla_{\theta} V 
		 (\theta_{t}, \phi_{t} ) - \beta \big(\nabla  
		V_{\theta}   (\theta_{t+1 / 2}, \phi_{t+1 / 2}  ) - 
		\nabla V_{\theta}  (\theta_{t}, \phi_{t} ) 
		\big),  \\
		\phi_{t+1} &= \phi_{t} + \alpha \nabla V_{\phi} 
		 (\theta_{t}, \phi_{t} ) + \beta \big(\nabla 
		V_{\phi} (\theta_{t+1 / 2}, \phi_{t+1 / 2} ) -  
		\nabla V_{\phi} (\theta_t, \phi_t  )\big).
	\end{split}
\end{equation}

In this paper, we will refer to this algorithm as the predictive 
centripetal acceleration algorithm (PCAA). Given its simplicity and 
effectiveness in the test examples, we have chosen to study it as the 
main algorithm for our research.  Additionally, it is easy to observe 
that by adjusting GDA to different extents using the approximate 
centripetal acceleration term, we can obtain algorithms named OGDA, 
GOGDA in \cite{mokhtari2020unified}, and Grad-SCA, respectively. 
Therefore, combining 
Proposition \ref{proposition1}, the following remark can be obtained, 
which can also be found in \cite{Li2023training}.

\begin{remark}[\cite{Li2023training}] \label{remarkPCAA}
	Although the idea of PCAA originates from OGDA, GOGDA, and 
	Grad-SCA, while PCAA still differs from them, primarily in terms 
	of how they adjust GDA. OGDA, GOGDA, and Grad-SCA use the 
	approximate centripetal acceleration term at time $t$, $\nabla 
	\bar{V} (\theta_{t+1}, \phi_{t+1} ) - \nabla 
	\bar{V} (\theta_t, \phi_t )$, to adjust the algorithm's 
	performance at time $t+1$. In contrast, PCAA first obtains the 
	gradient information for the predictive step and then uses this 
	information to adjust the algorithm's performance at time $t$ in 
	the direction pointing towards the center. Specifically, 
	according to Theorem \ref{proposition1}, this difference is 
	particularly pronounced in case of bilinear game 
	\eqref{eqbilinear}.
\end{remark}

Next, we provide the following relationships regarding the parameters 
in iterative formats between PCAA and GDA, EG and MPM, which can 
also be found in \cite{Li2023training}.

\begin{remark}[\cite{Li2023training}] \label{formatPCAA}
	It is evident that if $\gamma = 0, \alpha \neq 0$ or $\beta = 0, 
	\alpha \neq 0$ in PCAA iterative format \eqref{eqpcaa}, then
	PCAA  is equivalent to GDA. 
	Additionally,   if $\beta = \alpha \neq 0$ in \eqref{eqpcaa}, 
	then PCAA will be reduced to MPM.
	Furthermore, if $\beta = \alpha = \gamma$ in \eqref{eqpcaa}, then 
	PCAA is equivalent to EG.
\end{remark}

\begin{figure}[h!] 
	\subfloat[]{ \centering
		\includegraphics[width=225 pt, height =200 
		pt]{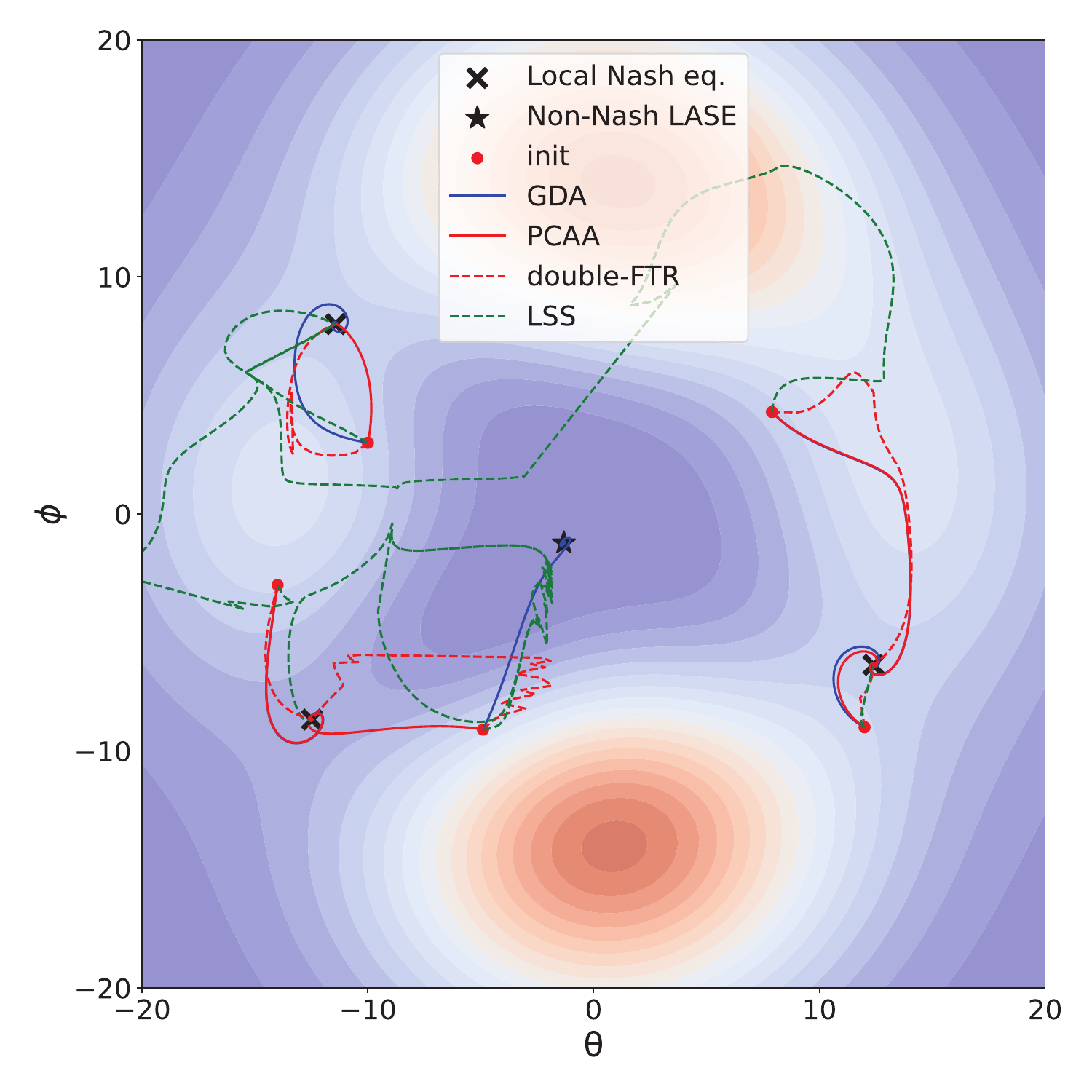}}
	\subfloat[]{\centering
		\includegraphics[width=105 pt, height =198 
		pt]{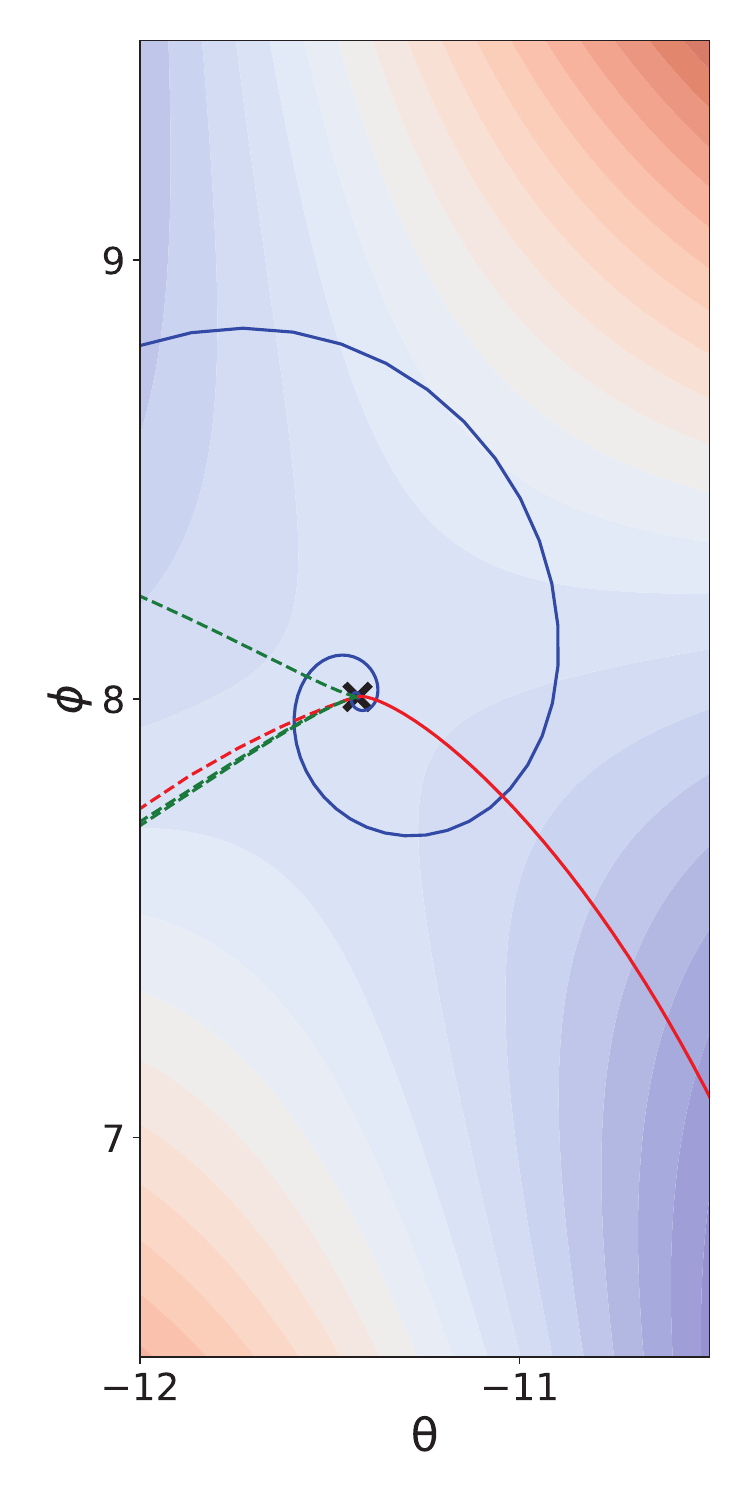}}
	\subfloat[]{\centering
		\includegraphics[width=125 pt, height =198 
		pt]{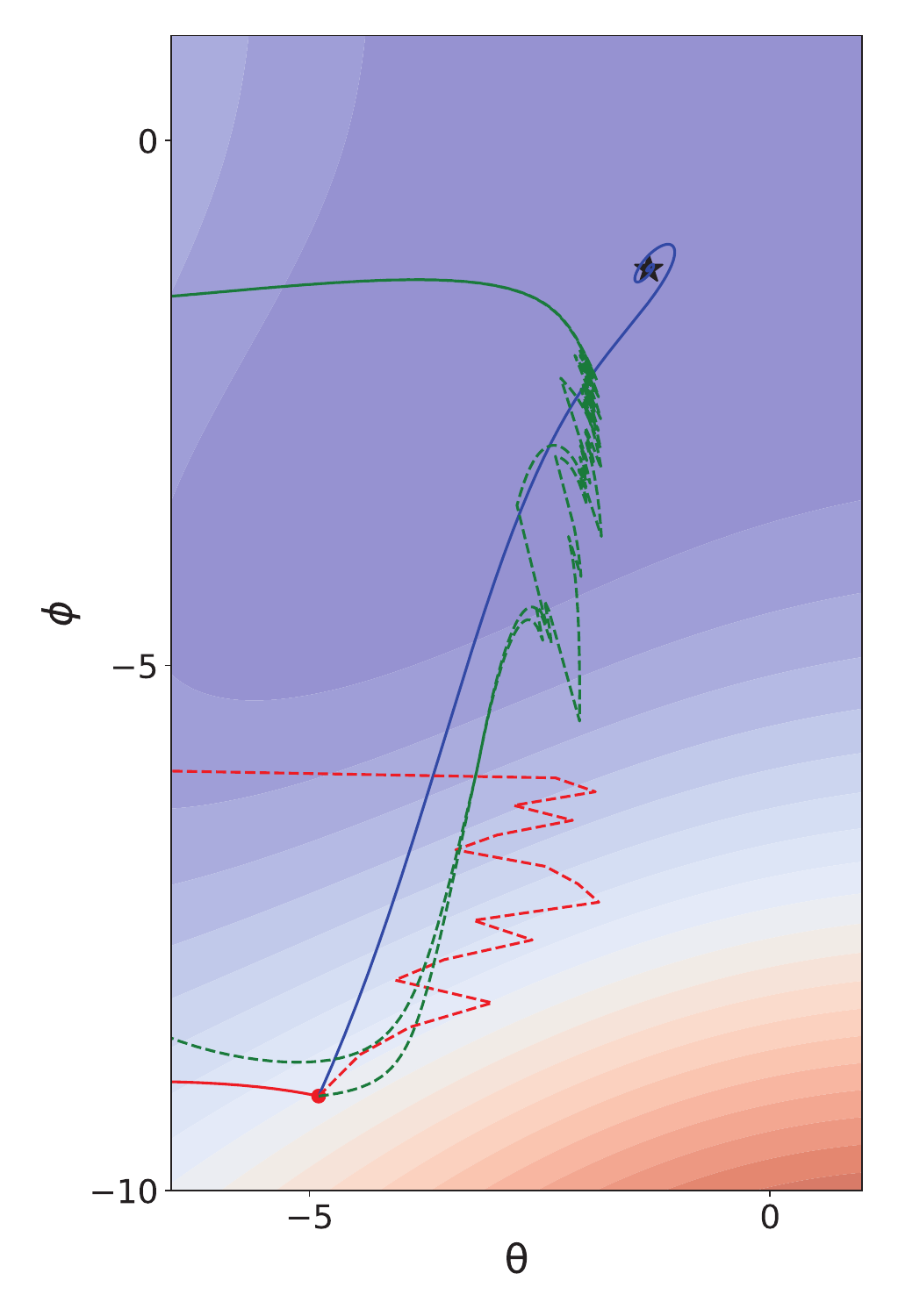} }
	\caption{(a): Evolution of GDA, double-FTR and PCAA in the 
		2-D toy example from multiple initial points. (b): 
		Zoom-in near a local Nash equilibrium point. (c): 
		Zoom-in 
		near a non-Nash LASE for the GDA algorithm. The definitions 
		of 
		LASE and LSS can be found in \cite{bao2022finding}.}
	\label{exPPCA2}
\end{figure}

Finally, we conducted comparative experiments on the example in 
\cite{bao2022finding} to illustrate the effectiveness of 
PCAA in terms of avoiding spurious critical points and oscillatory 
behavior near local Nash equilibria. For specific experimental 
details, please refer to Figure \ref{exPPCA2}.

As shown in Figure \ref{exPPCA2}, while GDA may converge to critical 
points that are not local Nash equilibria, PCAA can avoid such 
spurious critical points. Additionally, GDA exhibits oscillatory 
behavior near local Nash equilibria, whereas PCAA does not have 
oscillatory behavior near local Nash equilibria. For reference, we 
also show the trajectories of Local Symplectic Surgery (LSS) 
\cite{mazumdar2019finding}, as mentioned in \cite{bao2022finding}.

\section{Two illustrative examples}

\subsection{An illustrative example: Learning the mean of a 
multivariate Gaussian distribution.}

The third part of \cite{daskalakis2018training} 
considers a simple Wasserstein GAN example, where data is generated 
from a 
multivariate normal distribution, i.e. $Q \triangleq  {N}(v, I)$ for 
some $v \in \mathbb{R}^{n}$. The purpose is to enable the generator 
to learn the unknown parameter $v$. The specific details are as 
follows.

Consider a Wasserstein GAN, where the discriminator is a linear 
function and the generator is an additive shift of the input noise 
$z$, where the noise $z$ is sampled from the distribution $F 
\triangleq {N}(0, I)$, i.e.
\begin{equation*}
	\begin{aligned}
		D_{\phi}(x)  = \langle \phi, x \rangle,  \quad 
		G_{\theta}(z)  = z + \theta.
	\end{aligned}
\end{equation*}
The goal of the generator is to approximate the true distribution, 
i.e., for $\theta$ to converge to $v$. The loss function of the 
Wasserstein GAN takes the following form:
\begin{equation*} 
	L (\theta, \phi)=\mathbb{E}_{x \sim  {N}(v, I)}[\langle \phi, 
	x\rangle]-\mathbb{E}_{z \sim  {N}(0, I)}[\langle \phi, 
	z+\theta\rangle].
\end{equation*}
Consider a scenario where we optimize the true expectation mentioned 
above instead of assuming only samples of $x$ and $z$. Given the 
linearity property of expectations, the expected zero-sum game takes 
the following form:
\begin{equation}\label{Gm2}
	\inf _{\theta} \sup _{\phi}\langle \phi, v-\theta\rangle.
\end{equation}
For more detailed derivations regarding the loss function, please 
refer to \cite{daskalakis2018training}.
Note that the equilibrium point of above game \eqref{Gm2} is 
unique, namely the generator parameters are $v = \theta$ and the 
discriminator parameters are $\phi = 0$. For this simple zero-sum 
game, we have $\nabla_{\mathbf{\phi}} L = v - \theta$ and 
$\nabla_{\mathbf{\theta}} L = -\phi$. Therefore, the iterative format 
for GDA is as follows:

\begin{equation*} \label{Gm3}
	\begin{aligned}
		\phi_{t+1}  =\phi_{t} + \eta    (v-\theta_{t} ), 
		\quad
		\theta_{t+1}  =\theta_{t} + \eta  \phi_{t}.
	\end{aligned}  
\end{equation*}
The iterative format for  OGDA is as follows:
\begin{equation*} \label{Gm4}
	\begin{aligned}
		\phi_{t+1}  = \phi_{t}+2 \eta  
		 (v-\theta_{t} )-\eta   (v-\theta_{t-1} ), 
		\quad
		\theta_{t+1}  = \theta_{t}+2 \eta  \phi_{t}-\eta  \phi_{t-1}.
	\end{aligned}
\end{equation*}
The iterative format for MPM is as follows:
\begin{equation*} \label{Gm5}
	\begin{aligned}
		&\phi_{t+1/2}  = \phi_{t} + \gamma   
		 (v-\theta_{t} ),  \qquad
		\theta_{t+1/2} = \theta_{t} + \gamma   \phi_{t}; \\
		&\phi_{t+1} = \phi_{t} +  \beta  
		 (v-\theta_{t+1/2} ),  \quad
		\theta_{t+1} = \theta_{t} +  \beta   \phi_{t+1/2}.
	\end{aligned}
\end{equation*}
While the iterative format for  PCAA in this paper is as follows:  
\begin{equation*} \label{Gm6}
	\begin{aligned}
		&\phi_{t+1/2} = \phi_{t} + \gamma   
		 (v-\theta_{t} ),   \quad 
		\quad\quad\quad\quad\quad\quad~~
		\theta_{t+1/2} = \theta_{t} + \gamma   \phi_{t}; \\
		&\phi_{t+1} = \phi_{t} +  \alpha  (v-\theta_{t} ) - 
		\beta  (\theta_{t+1/2} - \theta_{t} ),  \quad
		\theta_{t+1}  = \theta_{t} +  \alpha   \phi_{t} + \beta   
		(\phi_{t+1/2} - \phi_{t}  ).
	\end{aligned}
\end{equation*}
The effectiveness of OGDA in alleviating the issue of limit cycling 
behavior is demonstrated in Section 3 of 
\cite{daskalakis2018training}, using Wasserstein GAN as an example. 
Inspired by the work of \cite{daskalakis2018training}, this paper 
compares the numerical performance of several algorithms in 
mitigating limit cycling behavior using the same example, 
highlighting 
that PCAA is more effective than other methods in addressing the 
issue of limit cycling behavior.

\begin{figure}[h!]  
	\subfloat[GDA dynamics]{\centering
		\includegraphics[width=220 pt, height =90 
		pt]{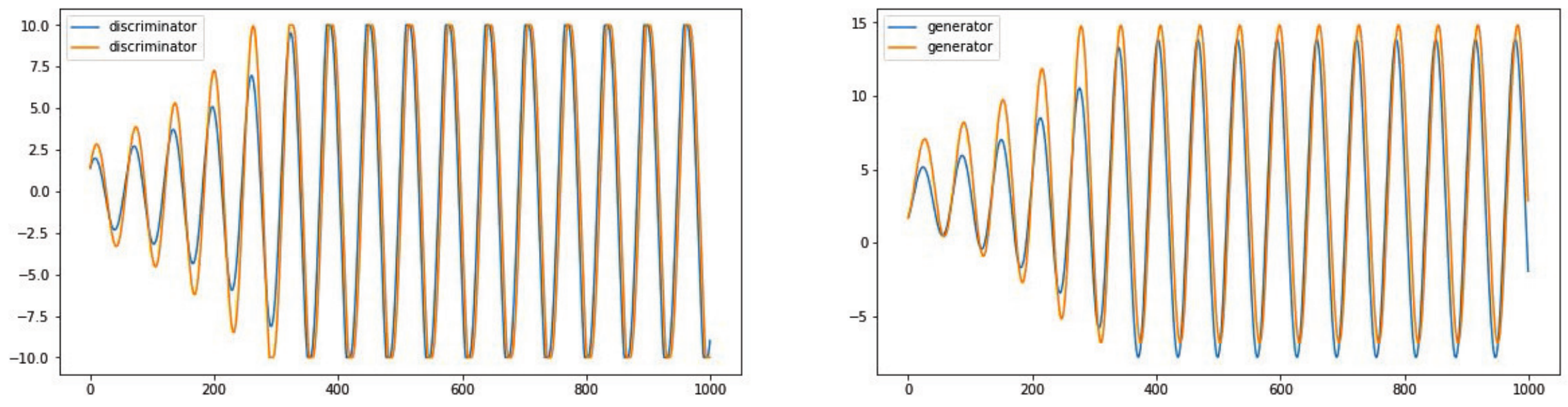}}  \quad
	\subfloat[OGDA dynamics]{
		\centering
		\includegraphics[width=220 pt, height =90 
		pt]{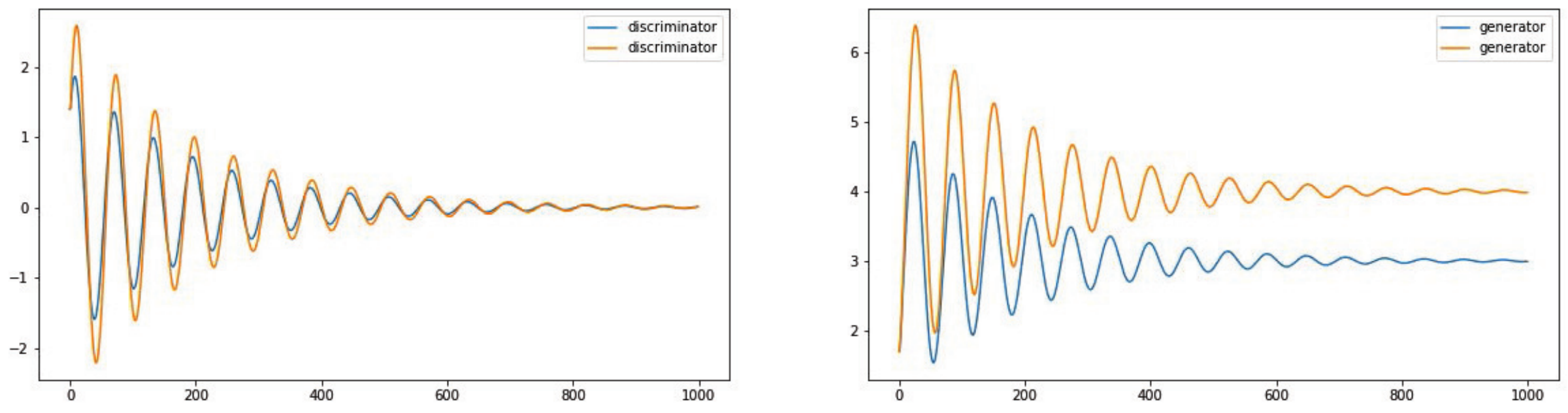} } 
	\caption{A reproduction of the results from Figure 1 in 
		\cite{daskalakis2018training}. (a):  Training this 
		Wasserstein GAN using 
		GDA converges to a limit cycle around the equilibrium point. 
		To prevent the GDA iteration process from cycling, the weight 
		clipping technique is applied to the discriminator parameters 
		during training.		 
		(b): Training Wasserstein GAN using OGDA converges to an 
		equilibrium point in the last-iterate convergence.}
	\label{Gaussmean1}
\end{figure}

\begin{figure}[h!]  
     \subfloat[MPM dynamics]{
		\centering\includegraphics[width=220 pt, height =90 
		pt]{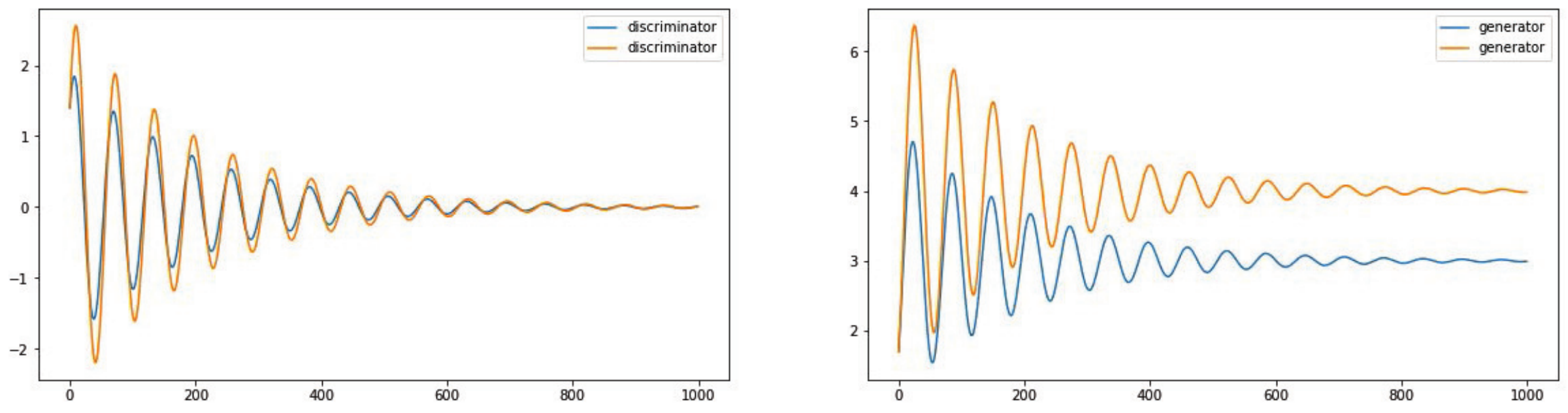}} \quad 
	\subfloat[PCAA dynamics]{
		\centering\includegraphics[width=220 pt, height =90 
		pt]{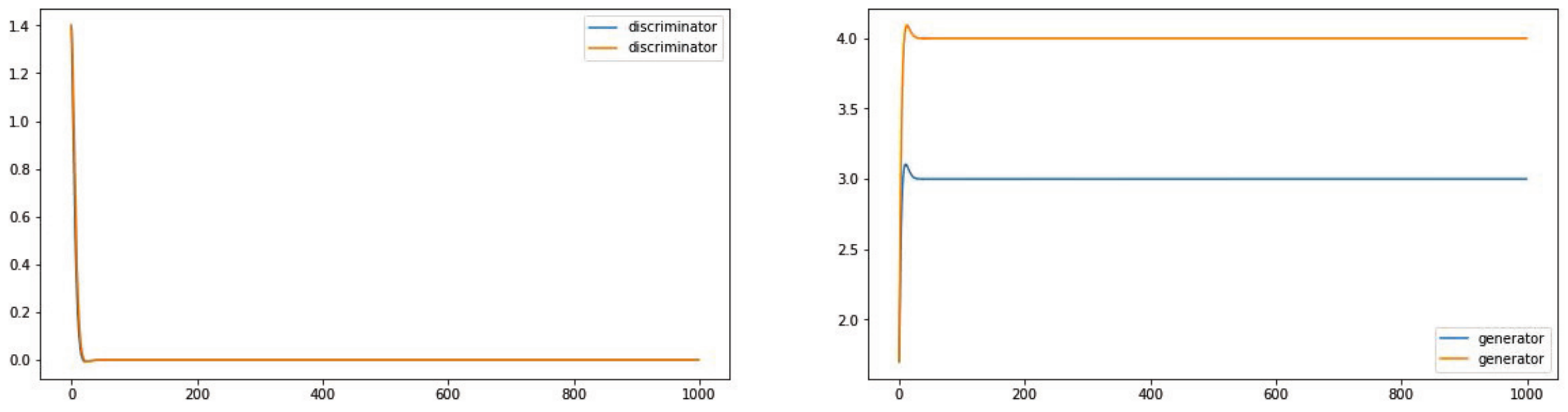}}
	\caption{(a):Training the Wasserstein GAN using MPM leads to 
		convergence in the sense of the last-iterate convergence, but 
		the 
		algorithm exhibits slight oscillations. (b): Training the 
		Wasserstein GAN using PCAA results in convergence in terms of 
		the 
		last-iterate convergence, and the iteration process shows 
		almost 
		no oscillations or limit cycling behavior.}
	\label{Gaussmean2}
\end{figure}

We proceed with simultaneous training on zero-sum game 
\eqref{Gm2} using GDA, OGDA, MPM and PCAA. It is noted in 
\cite{daskalakis2018training} that the GDA iteration process 
exhibits limit cycling behavior regardless of the step size or other 
modifications. Figure \ref{Gaussmean1} to Figure \ref{Gaussmean2} 
display 
the dynamic behavior of GDA, OGDA, MPM, and PCAA on this game when $v 
= (3, 4)$.
The horizontal axis in Figure \ref{Gaussmean1} and  
Figure \ref{Gaussmean2} represents the number of iterations, while 
the 
vertical axis represents the corresponding values for the generator 
or 
discriminator. From the results of the four experiments mentioned 
above, 
it can be observed that the GDA dynamic leads to limit cycling 
behavior (although it does converge to the true vector $v$ in the 
average sense) in this game. On the other hand, OGDA, MPM, and 
PCAA dynamics can converge to the true vector $v$ in terms of 
last-iterate sense. 
However,  OGDA and MPM dynamics exhibit slight 
oscillations, while the PCAA dynamic shows almost no 
oscillation. Furthermore,  Daskalakis et al. 
\cite{daskalakis2018training} 
pointed out that improving the stability of the GDA dynamic 
can be achieved by adding a gradient penalty term to the GDA 
iteration scheme, introducing Nesterov momentum, or updating the 
generator once 
and the discriminator multiple times. However, unlike OGDA, these 
methods only narrow down the range of limit cycles, and the limit 
cycling behavior still persists. In fact, comparing the variation of 
oscillation 
amplitude with the number of iterations in the four experiments 
mentioned above, it can be observed that when OGDA, MPM, and PCAA 
have 
the same corresponding parameters, PCAA exhibits better numerical 
performance in alleviating the limit cycling behavior compared to 
OGDA and MPM.

\subsection{Another  example: Learning a co-variance matrix.}

\begin{figure}[h!] 
	\subfloat[GDA dynamics with $\alpha = 0.1, T = 500, \lambda 
	= 0.4$. ]{
		\begin{minipage}[t]{0.33\textwidth}
			\centering
			\centering\includegraphics[width=149 pt, height =130 
			pt]{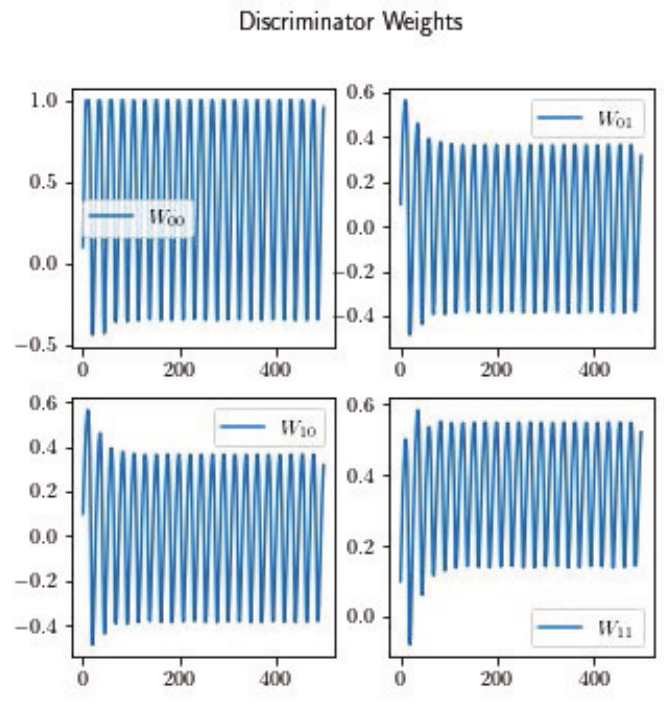}
		\end{minipage} 
		\begin{minipage}[t]{0.33\textwidth}
			\centering
			\centering\includegraphics[width=149 pt, height =130 
			pt]{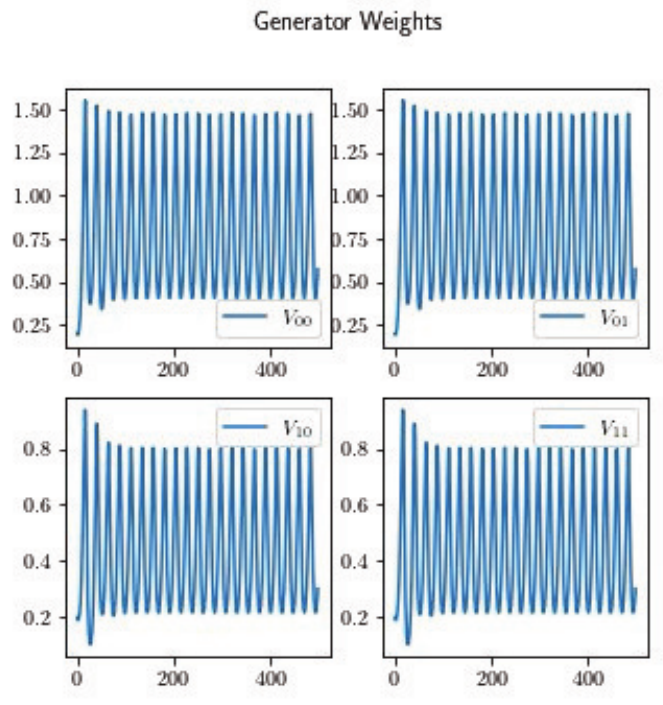}
		\end{minipage}
		\begin{minipage}[t]{0.33\textwidth}
			\centering
			\centering\includegraphics[width=149 pt, height =130 
			pt]{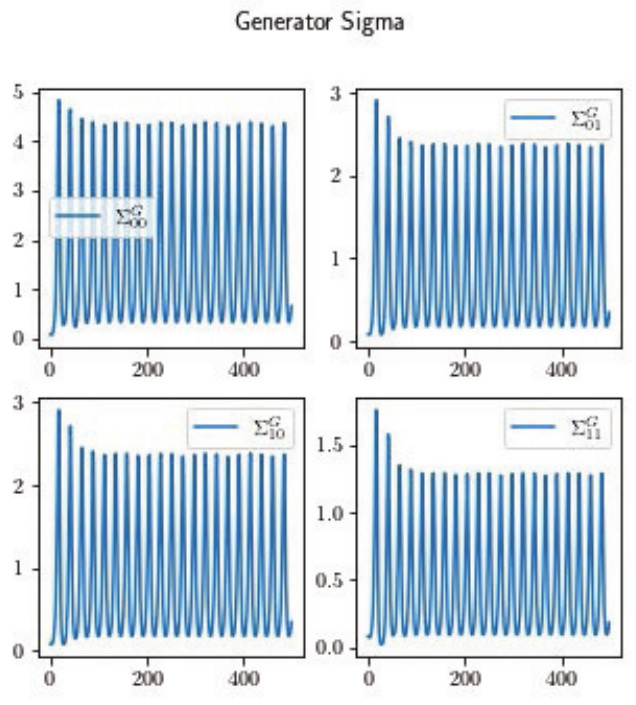}
	\end{minipage}} \\
	\subfloat[OGDA dynamics with $\alpha = 0.1, T = 500, \lambda = 
	0.4$.]{
		\begin{minipage}[t]{0.32\textwidth}
			\centering
			\centering\includegraphics[width=151 pt, height =130 
			pt]{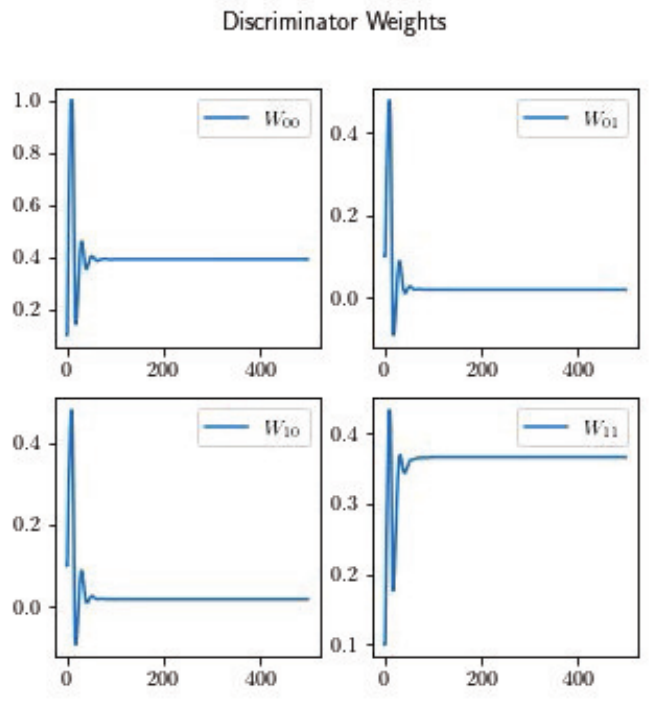}
		\end{minipage}~
		\begin{minipage}[t]{0.32\textwidth}
			\centering
			\centering\includegraphics[width=151 pt, height =130 
			pt]{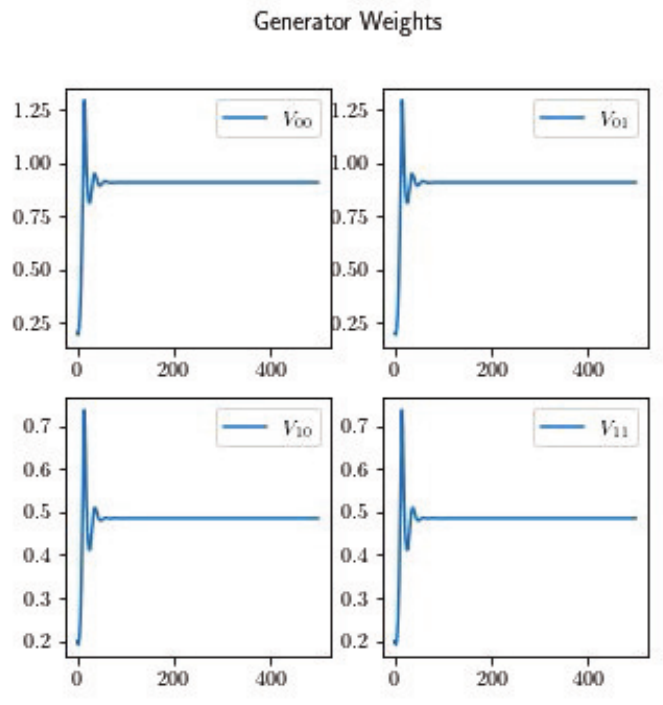}
		\end{minipage}~
		\begin{minipage}[t]{0.32\textwidth}
			\centering
			\centering\includegraphics[width=151 pt, height =130 
			pt]{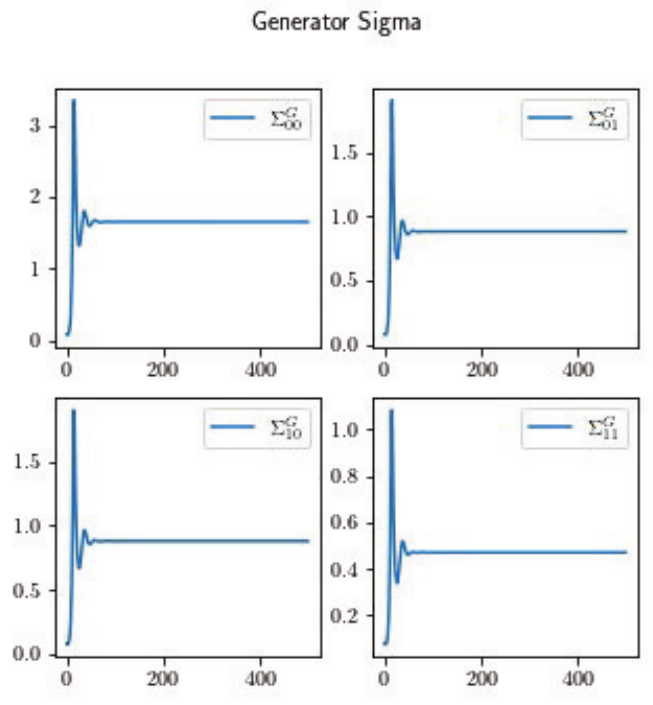}
	\end{minipage}} \\
	\subfloat[MPM dynamics with $\gamma=\alpha = 0.1, T = 500, 
	\lambda = 
	0.4$.]{
		\begin{minipage}[t]{0.32\textwidth}
			\centering
			\centering\includegraphics[width=151 pt, height =130 
			pt]{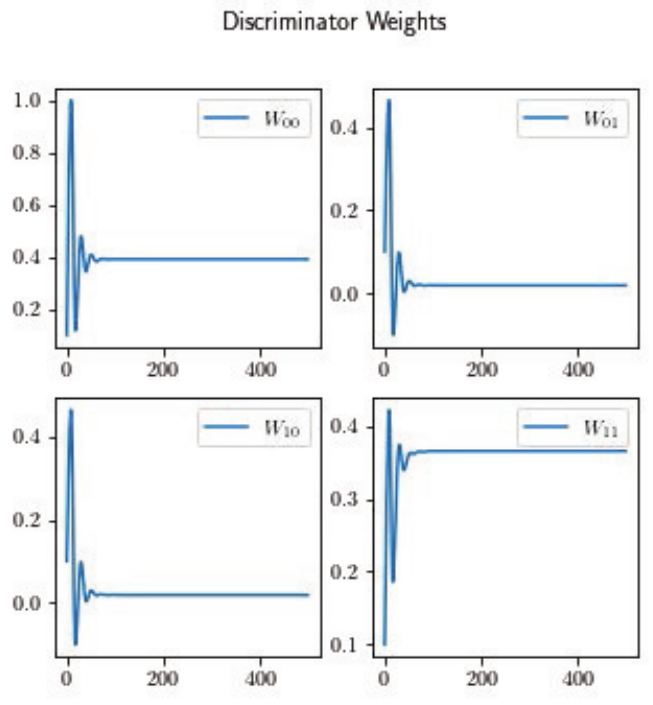}
		\end{minipage}~  
		\begin{minipage}[t]{0.32\textwidth}
			\centering
			\centering\includegraphics[width=151 pt, height =130 
			pt]{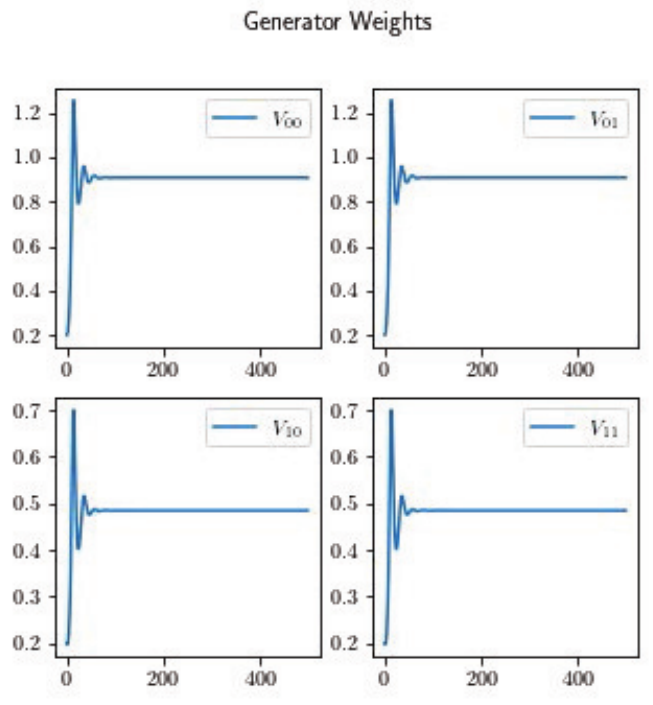}
		\end{minipage}~ 
		\begin{minipage}[t]{0.32\textwidth}
			\centering
			\centering\includegraphics[width=151 pt, height =130 
			pt]{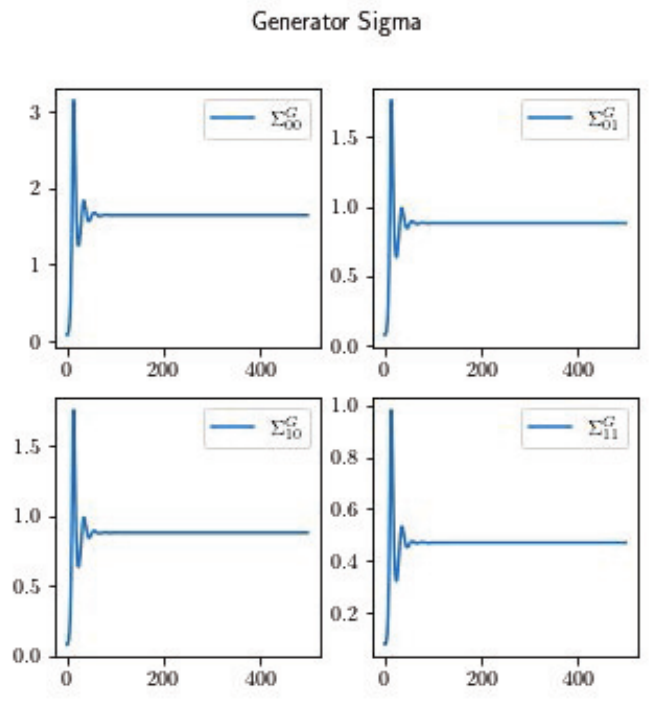}
	\end{minipage}} \\
	\subfloat[PCAA dynamics with $\gamma=\alpha = 0.1, \beta 
	=0.07, T = 500, \lambda = 0.4$.]{
		\begin{minipage}[t]{0.32\textwidth}
			\centering
			\centering\includegraphics[width=151 pt, height =130 
			pt]{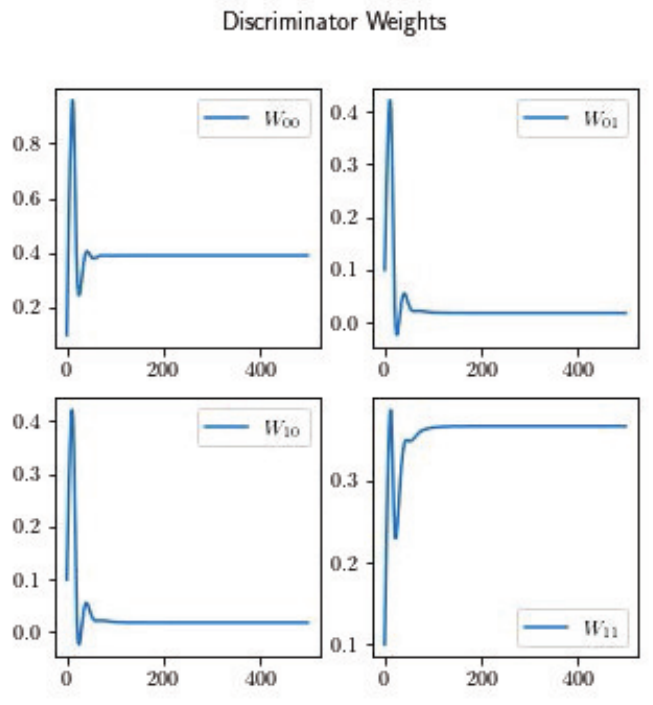}
		\end{minipage}~  
		\begin{minipage}[t]{0.32\textwidth}
			\centering
			\centering\includegraphics[width=151 pt, height =130 
			pt]{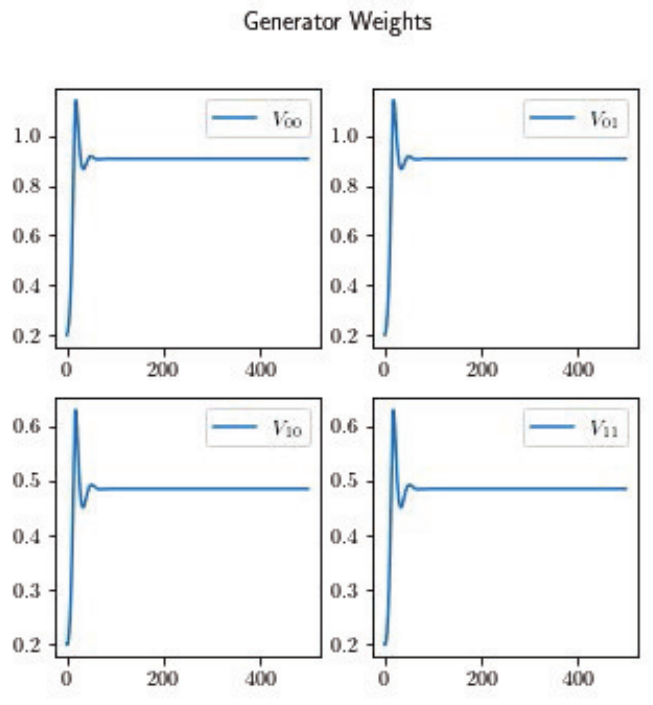}
		\end{minipage}~ 
		\begin{minipage}[t]{0.32\textwidth}
			\centering
			\centering\includegraphics[width=151 pt, height =130 
			pt]{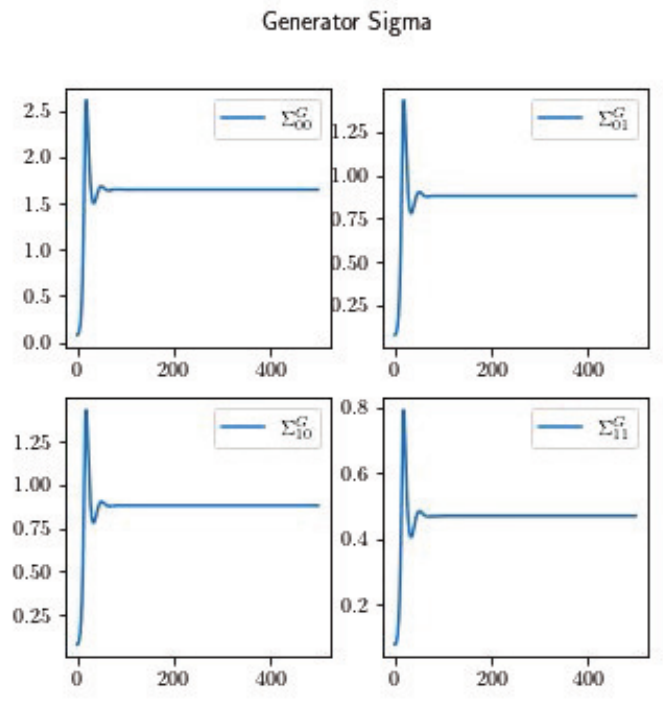}
			\label{gsc4-1}	
	\end{minipage}}
	\caption{Subfigure (a) to (d) present comparative experiments on 
		the stability of GDA, OGDA, MPM, and PCA in covariance matrix 
		learning for a two-dimensional Gaussian distribution ($d = 
		2$). 
		Weight clipping in [-1,1] was applied in four dynamics. 
	}		\label{gaussc1}	
\end{figure}

\begin{figure}[h!]  
	\subfloat[GDA dynamics with $\alpha = 0.1, T = 500, \lambda 
	= 0.4$. ]{
		\begin{minipage}[t]{0.32\textwidth}
			\centering
			\centering\includegraphics[width=154 pt, height =130 
			pt]{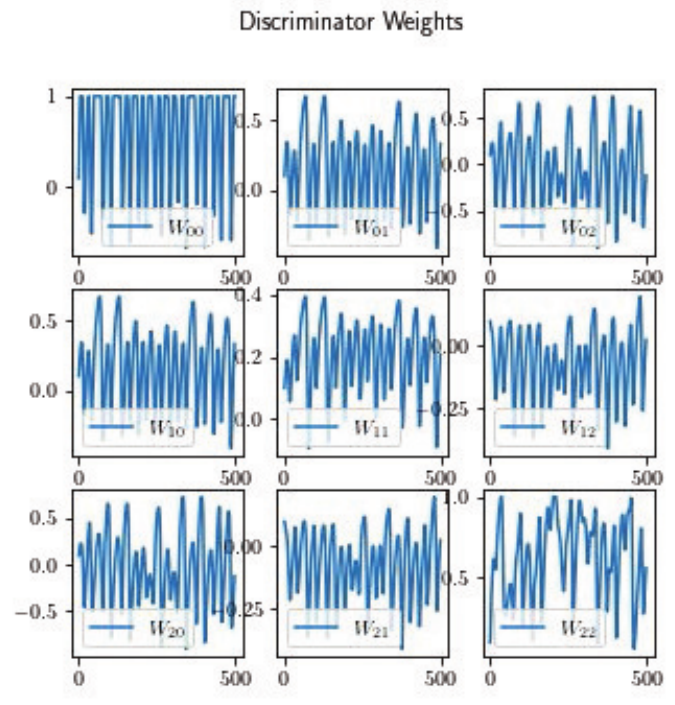}
		\end{minipage}~  
		\begin{minipage}[t]{0.32\textwidth}
			\centering
			\centering\includegraphics[width=154 pt, height =130 
			pt]{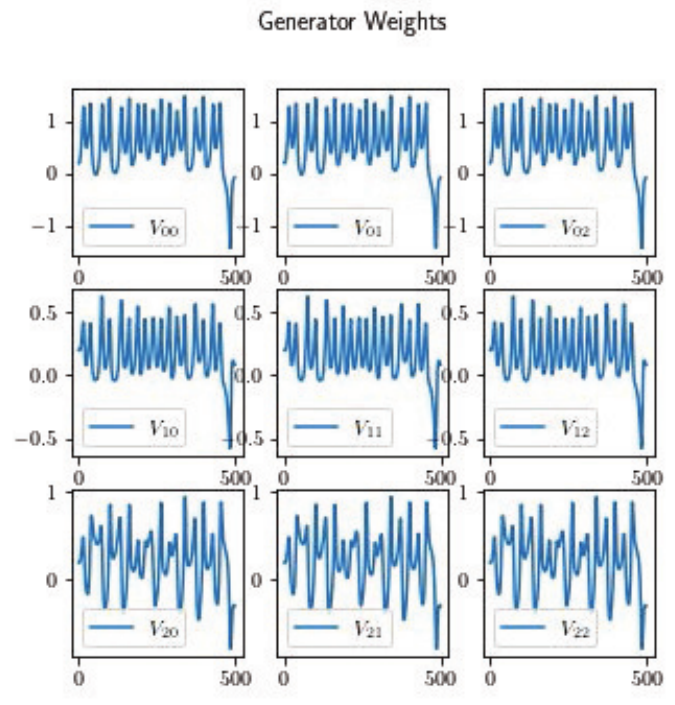}
		\end{minipage}~ 
		\begin{minipage}[t]{0.32\textwidth}
			\centering
			\centering\includegraphics[width=154 pt, height =130 
			pt]{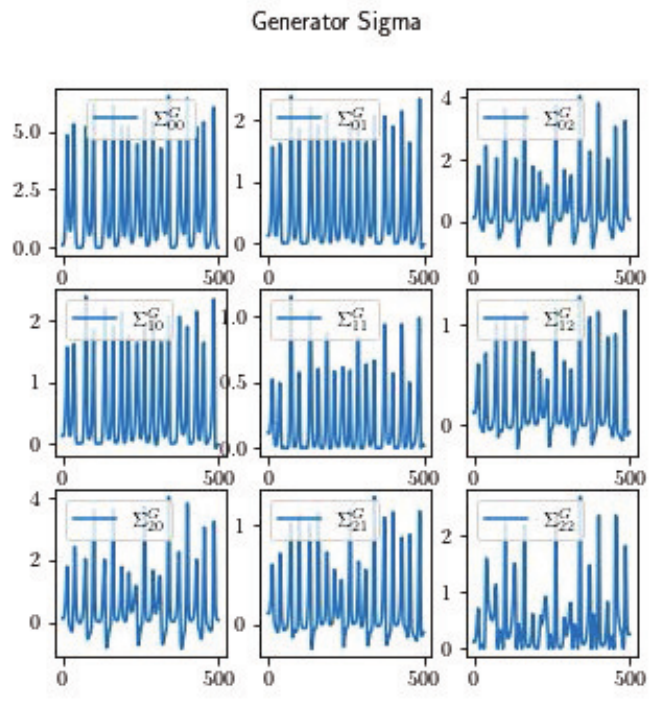}
	\end{minipage} } \\
	\subfloat[OGDA dynamics with $\alpha = 0.1, T = 500, 
	\lambda = 
	0.4$.]{
		\begin{minipage}[t]{0.32\textwidth}
			\centering
			\centering\includegraphics[width=154 pt, height =130 
			pt]{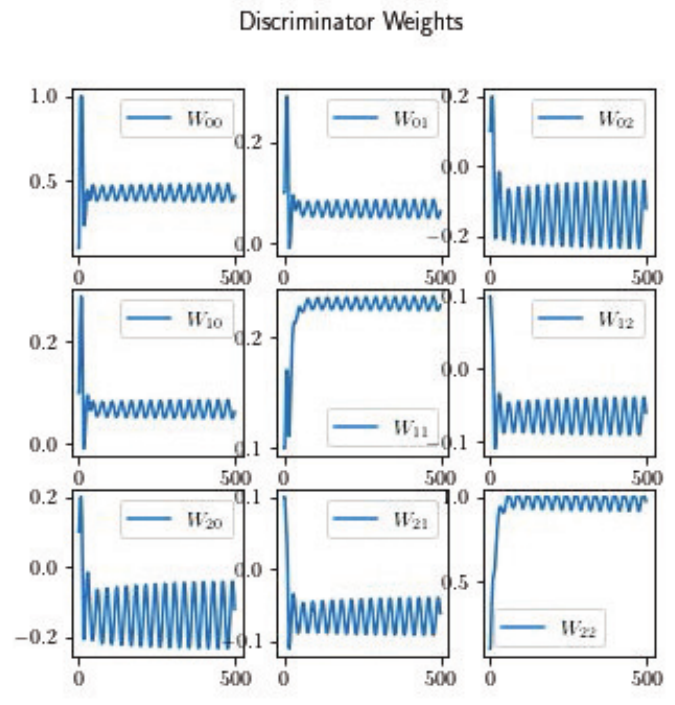}
		\end{minipage}~  
		\begin{minipage}[t]{0.32\textwidth}
			\centering
			\centering\includegraphics[width=154 pt, height =130 
			pt]{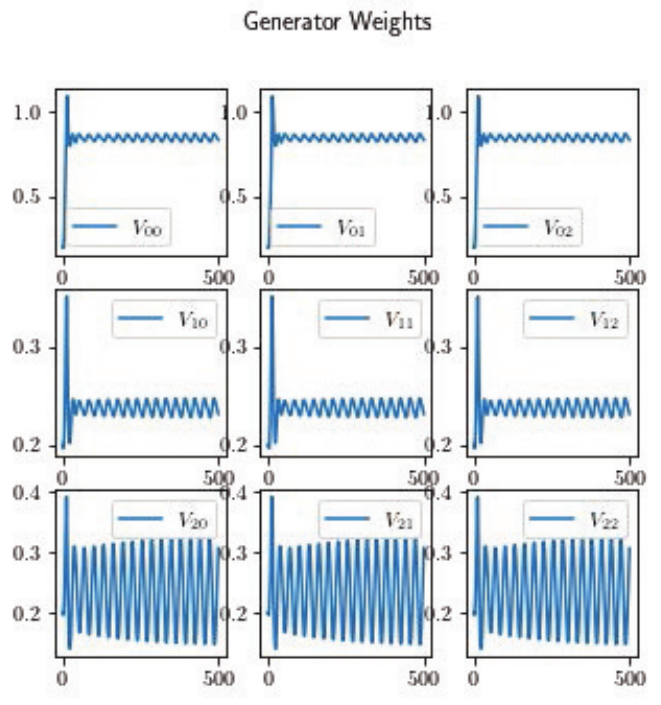}
		\end{minipage}~ 
		\begin{minipage}[t]{0.32\textwidth}
			\centering
			\centering\includegraphics[width=154 pt, height =130 
			pt]{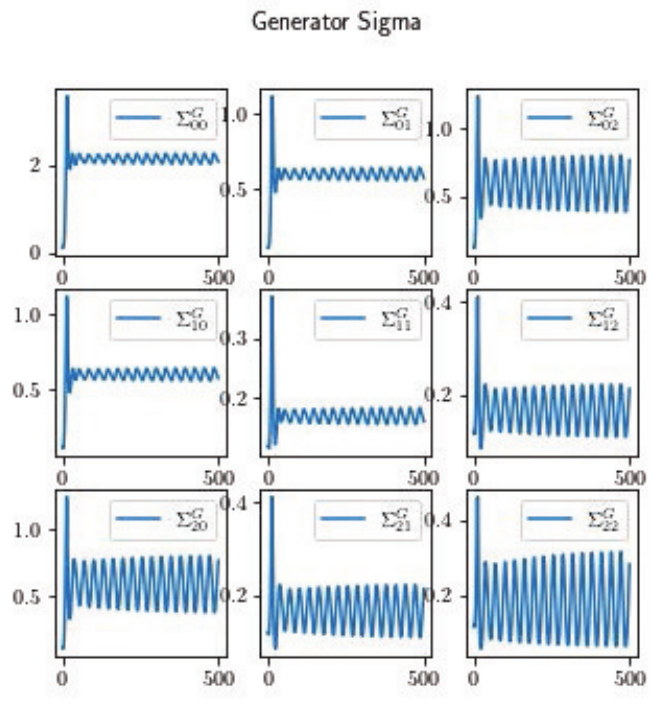}
	\end{minipage}} \\
	\subfloat[MPM dynamics with $\gamma=\alpha = 0.1, T = 500, 
	\lambda = 
	0.4$. ]{
		\begin{minipage}[t]{0.32\textwidth}
			\centering
			\centering\includegraphics[width=154 pt, height =130 
			pt]{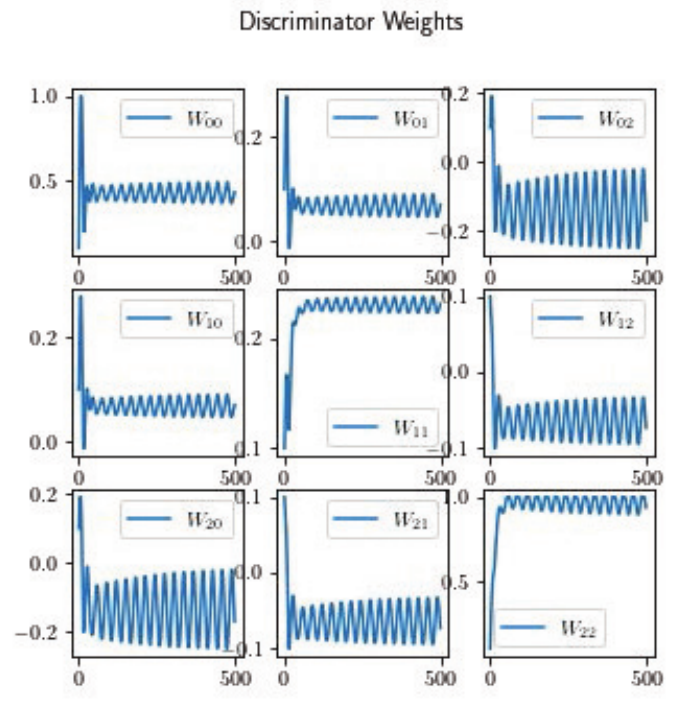}
		\end{minipage}~  
		\begin{minipage}[t]{0.32\textwidth}
			\centering
			\centering\includegraphics[width=154 pt, height =130 
			pt]{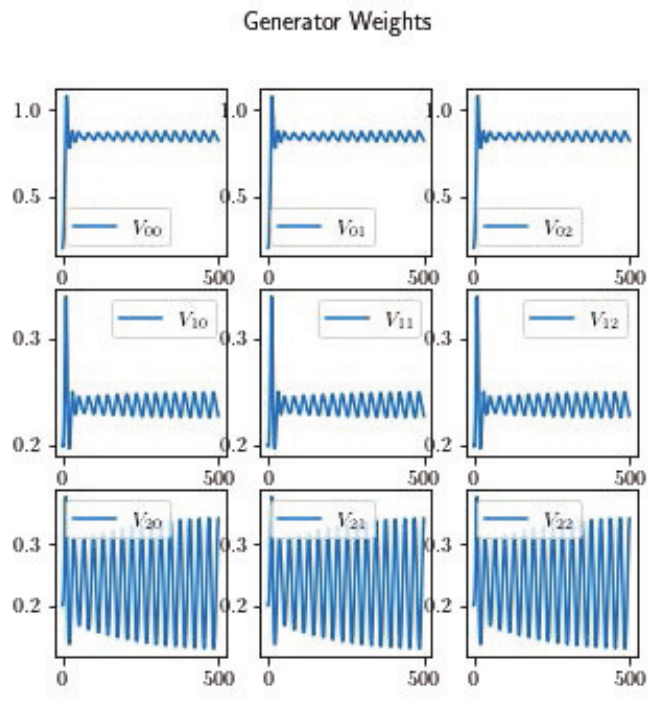}
		\end{minipage}~ 
		\begin{minipage}[t]{0.32\textwidth}
			\centering
			\centering\includegraphics[width=154 pt, height =130 
			pt]{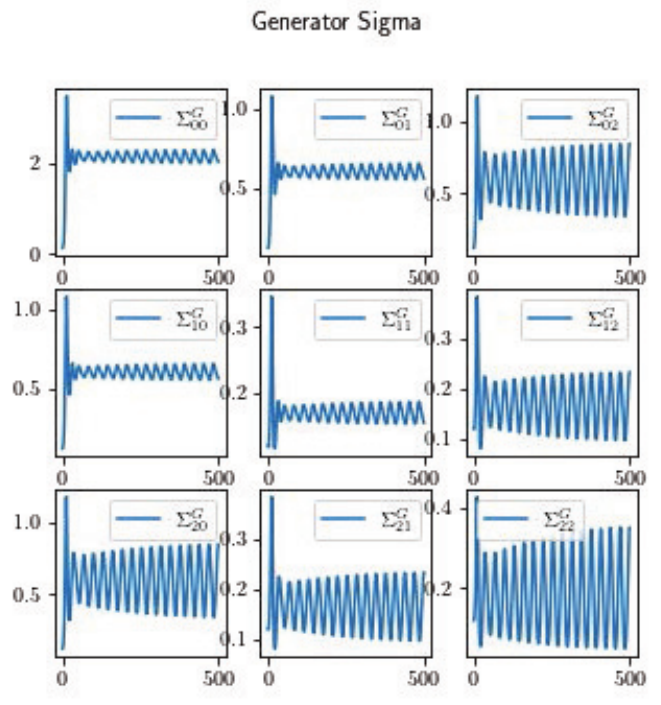}
	\end{minipage}} \\
	\subfloat[PCAA dynamics with $\gamma=\alpha = 0.1, \beta =0.04, T 
	= 500, \lambda = 0.4$.]{
		\begin{minipage}[t]{0.32\textwidth}
			\centering
			\centering\includegraphics[width=154 pt, height =130 
			pt]{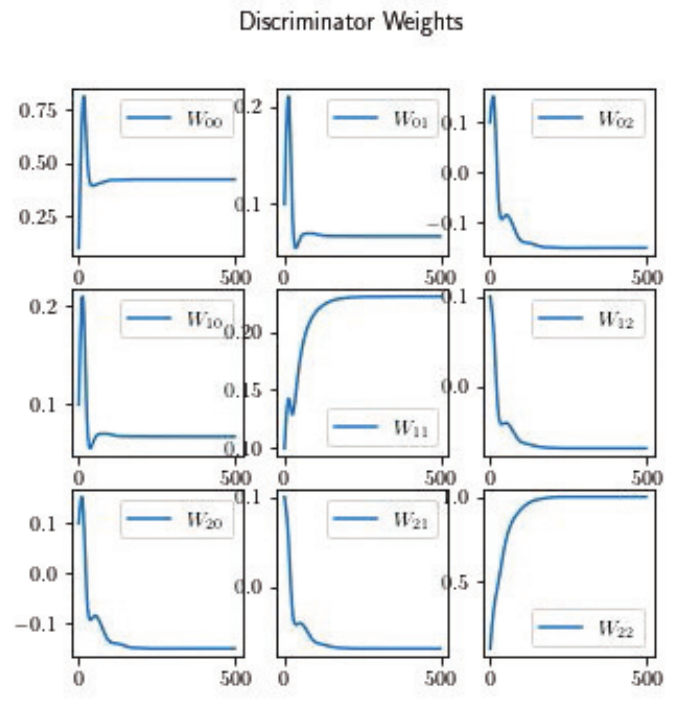}
		\end{minipage}~  
		\begin{minipage}[t]{0.32\textwidth}
			\centering
			\centering\includegraphics[width=154 pt, height =130 
			pt]{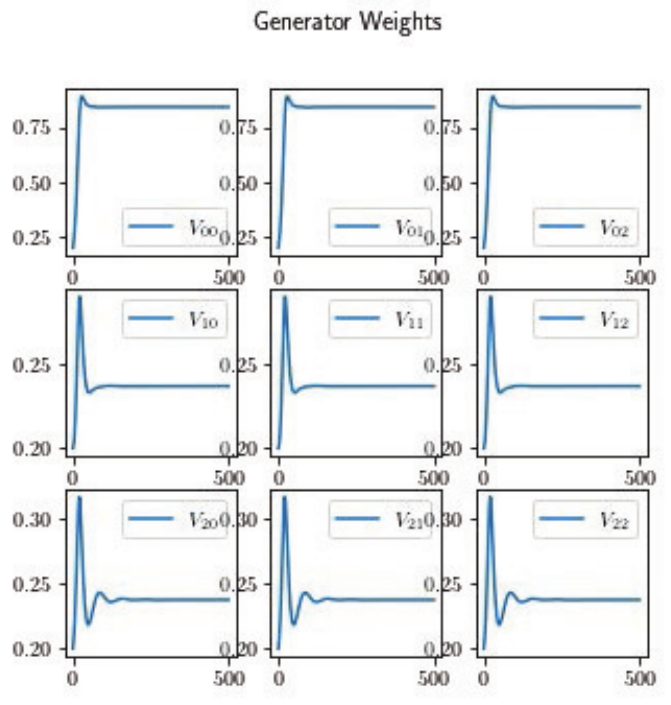}
		\end{minipage}~ 
		\begin{minipage}[t]{0.32\textwidth}
			\centering
			\centering\includegraphics[width=154 pt, height =130 
			pt]{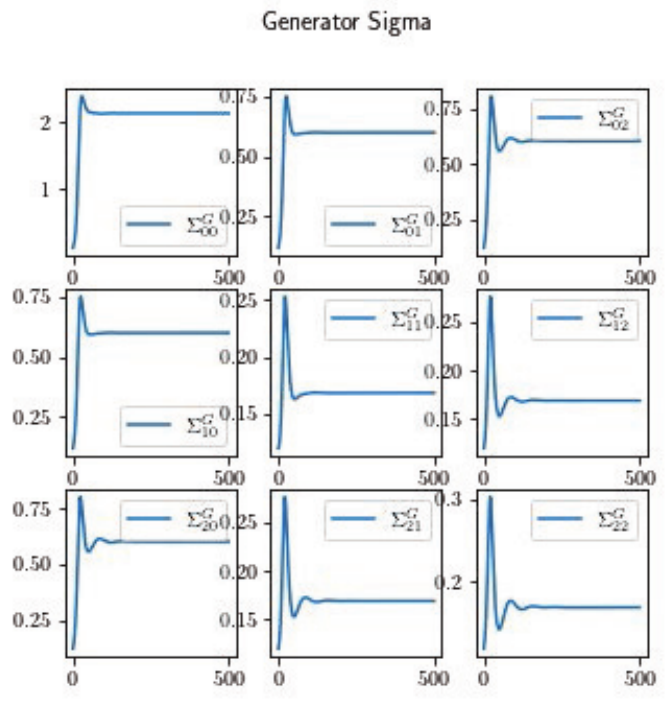}
	\end{minipage}}
	\caption{Subfigure (a) to (d) present comparative experiments on 
		the stability of GDA, OGDA, MPM, and PCA in covariance matrix 
		learning for a three-dimensional Gaussian distribution ($d = 
		3$). 
		Weight clipping in [-1,1] was applied in four dynamics. 
		Weight 
		clipping in [-1,1] was applied in four dynamics. }
	\label{gaussc2}
\end{figure}

This part demonstrates the effectiveness of PCAA in alleviating the 
limit cycling behaviors through an example of learning the covariance 
matrix of a multivariate Gaussian distribution. The problem 
description in this section is similar to that in  
\cite{daskalakis2018training}. It assumes that the data distribution 
follows a multivariate normal distribution with zero mean and an 
unknown covariance matrix, i.e., $x \sim N(0, \Sigma)$. Now, let's 
consider the case where the discriminator is a quadratic function, 
i.e.,
\begin{equation*}\label{Gc1}
	D_{W}(x)=\sum_{i j} W_{i j} x_{i} x_{j}=x^{\mathrm{T}} W x,
\end{equation*}
The generator is a linear function of the input random noise, denoted 
as $z \sim N(0, I)$, and is defined as follows:
\begin{equation*}\label{Gc2}
	G_{V}(z)=V z, 
\end{equation*}
where $W, V \in \mathbb{R}^{n \times n}$.
The Wasserstein GAN loss function with the aforementioned generator 
and discriminator is given by:
\begin{equation}\label{Gc3}
	L(V, W)=\mathbb{E}_{x \sim N(0, \Sigma)}\left[x^{\mathrm{T}} W 
	x\right]-\mathbb{E}_{z \sim N(0, I)}\left[z^{\mathrm{T}} 
	V^{\mathrm{T}} W V z\right].
\end{equation}
Expanding the loss function \eqref{Gc3} and assuming that the 
covariance matrix is a positive definite matrix, which can be 
obtained through Cholesky decomposition as $\Sigma = 
UU^{\mathrm{T}}$, then the loss function \eqref{Gc3} can be 
simplified to:
\begin{equation*}\label{Gc4}
	L(V, W)=\sum_{i j} W_{i j}\bigg(\Sigma_{i j}-\sum_{k} V_{i k} 
	V_{j k}\bigg)=\sum_{i j k} W_{i j}\bigg(U_{i k} U_{j k}-V_{i k} 
	V_{j k}\bigg).
\end{equation*}
For more detailed derivations regarding the loss function, please 
refer to \cite{daskalakis2018training}. Consider the 
iteration process of the algorithm without sampling noise. Then, The 
GDA update rule for this example is as follows:
\begin{equation*} 
	\begin{aligned}
		W_{t}  = W_{t-1} +\alpha \big(\Sigma-V_{t-1} 
		V_{t-1}^{\mathrm{T}}\big),  \quad 
		V_{t}  = V_{t-1} + 
		\alpha\big(W_{t-1}+W_{t-1}^{\mathrm{T}}\big) V_{t-1}.
	\end{aligned}
\end{equation*}
Similarly, the OGDA update rule can be written as follows:
\begin{equation*} 
	\begin{aligned}
		W_{t} &=W_{t-1} + 2 \alpha \big(\Sigma-V_{t-1} 
		V_{t-1}^{\mathrm{T}}\big)- \alpha \big(\Sigma-V_{t-2} 
		V_{t-2}^{\mathrm{T}}\big), \\
		V_{t} &=V_{t-1}+2 
		\alpha\big(W_{t-1}+W_{t-1}^{\mathrm{T}}\big) 
		V_{t-1}-\alpha \big(W_{t-2}+W_{t-2}^{\mathrm{T}}\big) 
		V_{t-2}.
	\end{aligned}
\end{equation*}
The MPM update rule can be written as follows:
\begin{equation*}
	\begin{aligned}
		&W_{t-1/2} = W_{t-1} + \gamma \big(\Sigma-V_{t-1} 
		V_{t-1}^{\mathrm{T}}\big),  
		V_{t-1/2}  = V_{t-1} + \gamma \big(W_{t-1} + 
		W_{t-1}^{\mathrm{T}}\big) V_{t-1};\\
		&W_{t} = W_{t-1} + \alpha \big(\Sigma-V_{t-1/2} 
		V_{t-1/2}^{\mathrm{T}}\big), 
		V_{t}  = V_{t-1} + \alpha\big(W_{t-1/2} + 
		W_{t-1/2}^{\mathrm{T}}\big) V_{t-1/2}.
	\end{aligned}
\end{equation*}
The PCAA update rule can be written as follows:
\begin{equation*}
	\begin{aligned}
		&W_{t-1/2} = W_{t-1} + \gamma \big(\Sigma-V_{t-1} 
		V_{t-1}^{\mathrm{T}}\big),  \quad
		V_{t-1/2}  = V_{t-1} + \gamma \big(W_{t-1} + 
		W_{t-1}^{\mathrm{T}}\big) V_{t-1};\\
		&W_{t} = W_{t-1} + \alpha \big(\Sigma-V_{t-1} 
		V_{t-1}^{\mathrm{T}}\big)  + \beta \bigg( 
		\big(\Sigma-V_{t-1/2} V_{t-1/2}^{\mathrm{T}}\big) - 
		\big(\Sigma-V_{t-1} 
		V_{t-1}^{\mathrm{T}}\big) \bigg), \\
		&V_{t} = V_{t-1} + \alpha \big(W_{t-1} + 
		W_{t-1}^{\mathrm{T}}\big) V_{t-1} + \beta 
		\bigg(\big(W_{t-1/2} + W_{t-1/2}^{\mathrm{T}}\big) 
		V_{t-1/2} - \big(W_{t-1} + 
		W_{t-1}^{\mathrm{T}}\big) V_{t-1} \bigg).
	\end{aligned}
\end{equation*}
The two experiments below provided the weights of the discriminator 
and generator distributions, as well as the hidden covariance matrix 
$\Sigma = VV^{\mathrm{T}}$, at each iteration step, respectively, 
on two-dimensional and three-dimensional Gaussian distributions. 
Comparing the oscillation amplitudes with the number of iterations in 
the experimental results, it was found that PCAA performed better in 
terms of stabilizing the training process compared to the other three 
algorithms. 
What needs to be clarified is that in Figure \ref{gaussc1} and 
Figure \ref{gaussc2}, the x-axis represents the number of iterations, 
while 
the y-axis represents the numerical values corresponding to each 
component of the generator, discriminator, and covariance matrix.
By comparing the oscillation amplitudes with the number of 
iterations in Figure \ref{gaussc1}, it can be observed that in 
the 
covariance 
matrix learning problem of the two-dimensional Gaussian 
distribution, OGDA, MPM, and PCAA all effectively alleviate the 
issue of limit cycles.

Although all three methods were effective in alleviating 
limit cycling behavior in the two-dimensional experiment, in the 
example of learning the covariance matrix of a three-dimensional 
Gaussian distribution, it can be observed by comparing the changes in 
oscillation magnitude with iteration count for the four algorithms 
that PCAA is more effective than OGDA and MPM in mitigating the 
occurrence of limit cycling behavior during the GDA iteration process.

\section{Last-iterate linear convergence  of PCAA}

Below, we first present the characterization of the existence of Nash 
equilibria for game \eqref{eqbilinear} as described by Anagnostides 
and Penna in \cite{anagnostides2020solving}, and then we will address 
the second question we raised in this work.

\begin{proposition}[\cite{anagnostides2020solving}]
	\label{proposition5.1}
	A pair of strategies $(\theta^*, \phi^*) \in \mathbb{R}^m \times  
	\mathbb{R}^n$ constitutes a Nash equilibrium for
	bilinear game \eqref{eqbilinear} if and only if $A\phi^*=0$, and 
	$A^\mathrm{T}\theta^* = 0$, that is $\phi^* \in \mathcal{N}(A)$ 
	and $\theta^* \in \mathcal{N}(A^\mathrm{T})$, where $ 
	\mathcal{N}(A )$ represents the null space of matrix $A$.
\end{proposition}

\begin{proposition}[] 
	\label{proposition3.1}
	For bilinear game \eqref{eqbilinear}, we assume that matrix $A 
	\in 
	\mathbb{R}^{m \times n}$ and $ rank (A) = n$. If PCAA 
	converges, Then the limit points 
	are Nash equilibria.
\end{proposition}

\begin{proof}
	In bilinear game  \eqref{eqbilinear},  the 
	iterations of predictive step in \eqref{eqpcaa} are simplified to
	\begin{equation*}  
		\begin{aligned} 
			\bigg( 
			\begin{array}{c}  
				\theta_{t+1 / 2} \\
				\phi_{t+1 / 2}   
			\end{array} \bigg) 
			=\bigg( 
			\begin{array}{c}  
				\theta_{t} \\
				\phi_{t}
			\end{array} \bigg)  +     \bigg( 
			\begin{array}{c}  
				- \gamma A \phi_{t} \\
				\gamma A^{\mathrm{T}} \theta_{t} 
			\end{array} \bigg) 
			=\bigg( 
			\begin{array}{c}  
				\theta_{t} \\
				\phi_{t}
			\end{array} \bigg)  + \bigg(
			\begin{matrix}
				0 & -\gamma A \\
				\gamma A^{\mathrm{T}} &0
			\end{matrix} \bigg) 
			\bigg(
			\begin{array}{c}
				\theta_{t}\\ 
				\phi_{t}
			\end{array} \bigg). 
		\end{aligned}	
	\end{equation*}		
	Then, the gradient step dynamics in \eqref{eqpcaa} can be written 
	as 
	\begin{equation}\label{eqaa2}
		\begin{aligned}
			\bigg( 
			\begin{array}{c}  
				\theta_{t+1} \\
				\phi_{t+1}   
			\end{array} \bigg)
			&= \bigg( 
			\begin{array}{c}  
				\theta_{t} \\
				\phi_{t}   
			\end{array} \bigg) +    \alpha \bigg( 
			\begin{array}{c}  
				-A \phi_{t} \\
				A^{\mathrm{T}} \theta_{t}   
			\end{array} \bigg) +\beta \bigg( 
			\begin{array}{c}  
				- A  \phi_{t+1 / 2}  -    (-A \phi_t  )  \\
				A^{\mathrm{T}}  \theta_{t+1 / 2} -  A^{\mathrm{T}} 
				\theta_t   
			\end{array} \bigg)  \\
			&= \bigg( 
			\begin{array}{c}  
				\theta_{t} \\
				\phi_{t}
			\end{array} \bigg)  + \bigg(
			\begin{matrix}
				0 & -\alpha A \\
				\alpha A^{\mathrm{T}} &0
			\end{matrix} \bigg) 
			\bigg(
			\begin{array}{c}
				\theta_{t}\\ 
				\phi_{t}
			\end{array} \bigg) +\beta \bigg( 
			\begin{array}{c}  
				- A  ( \phi_{t+1 / 2}  -  \phi_t  )  \\
				A^{\mathrm{T}}  (\theta_{t+1 / 2} -  \theta_t  ) 
			\end{array} \bigg) \\
			&= \bigg( 
			\begin{array}{c}  
				\theta_{t} \\
				\phi_{t}
			\end{array} \bigg)  + \bigg(
			\begin{matrix}
				0 & -\alpha A \\
				\alpha A^{\mathrm{T}} &0
			\end{matrix} \bigg) 
			\bigg(
			\begin{array}{c}
				\theta_{t}\\ 
				\phi_{t}
			\end{array} \bigg) +\beta \bigg(
			\begin{matrix}
				0 & -A \\
				A^{\mathrm{T}} &0
			\end{matrix} \bigg)  \bigg(
			\begin{matrix}
				0 & -\gamma A \\
				\gamma A^{\mathrm{T}} &0
			\end{matrix} \bigg) 
			\bigg(
			\begin{array}{c}
				\theta_{t}\\ 
				\phi_{t}
			\end{array} \bigg) \\
			&= \bigg( 
			\begin{array}{c}  
				\theta_{t} \\
				\phi_{t}
			\end{array} \bigg)  + \bigg(
			\begin{matrix}
				0 & -\alpha A \\
				\alpha A^{\mathrm{T}} &0
			\end{matrix} \bigg) 
			\bigg(
			\begin{array}{c}
				\theta_{t}\\ 
				\phi_{t}
			\end{array} \bigg) +  \bigg(
			\begin{matrix}
				-\gamma \beta A A^{\mathrm{T}} & 0 \\
				0 & -\gamma \beta A^{\mathrm{T}} A
			\end{matrix} \bigg) 
			\bigg(
			\begin{array}{c}
				\theta_{t}\\ 
				\phi_{t}
			\end{array} \bigg) \\
			&= \bigg( 
			\begin{array}{c}  
				\theta_{t} \\
				\phi_{t}
			\end{array} \bigg)  + \bigg(
			\begin{matrix}
				-\gamma \beta A A^{\mathrm{T}} & -\alpha A \\
				\alpha A^{\mathrm{T}} & -\gamma \beta A^{\mathrm{T}} A
			\end{matrix} \bigg) 
			\bigg(
			\begin{array}{c}
				\theta_{t}\\ 
				\phi_{t}
			\end{array} \bigg) . 
		\end{aligned}
	\end{equation}
Given the assumption that PCAA converges, we set $\theta^* 
= \lim_{t 
	\to \infty} \theta_t$, and $\phi^* = \lim_{t \to \infty} \phi_t$. 
	It 
follows from \eqref{eqaa2} that  
\begin{equation}\label{eqaa3}
	\lim _{t 
		\rightarrow+\infty} (\theta_{t+1}-\theta_t )= - 
	\gamma \beta AA^{\mathrm{T}}\theta^* -\alpha A \phi^* =0,
\end{equation}
\begin{equation}\label{eqaa4}
	\lim _{t 
		\rightarrow+\infty} (\phi_{t+1}-\phi_t )=  
	\alpha A^{\mathrm{T}} \theta^*  -\gamma \beta A^{\mathrm{T}}A 
	\phi^* =0.
\end{equation}
Assuming that the parameters $\alpha, \gamma, \beta$ are all 
non-zero, we can deduce from \eqref{eqaa3} that
$A \phi^* = - \frac{\gamma \beta}{\alpha} AA^{\mathrm{T}}\theta^*.
$
By substituting it into \eqref{eqaa4}, we obtain
\begin{equation}\label{eqaa6}
	\alpha A^{\mathrm{T}} \theta^*  +  
	\frac{\gamma^2 \beta^2}{\alpha} A^{\mathrm{T}} 
	AA^{\mathrm{T}}\theta^* 
	=  \bigg(\alpha I + \frac{\gamma^2 \beta^2}{\alpha} 
	A^{\mathrm{T}} 
	A \bigg) A^{\mathrm{T}}\theta^* =0.
\end{equation}
By left-multiplying both sides of \eqref{eqaa6} with 
$A^{\mathrm{T}}\theta^*$, it can be inferred that
\begin{equation*}
	(A^{\mathrm{T}}\theta^*)^{\mathrm{T}} 
	\bigg(\alpha I + \frac{\gamma^2 \beta^2}{\alpha} A^{\mathrm{T}} 
	A\bigg) A^{\mathrm{T}}\theta^* =0.
\end{equation*}
Given the assumption that matrix $A$ has full column rank, we know 
that $A^{\mathrm{T}} 
A$ is a positive definite matrix. Combining this with the previous 
\eqref{eqaa6}, we can conclude that $A^{\mathrm{T}}\theta^* =0$. By 
substituting it into \eqref{eqaa3} and combining with 
Proposition \ref{proposition5.1}, 
we can conclude that $A \phi^* =0$.
Thus,
strategies $(\theta^*, \phi^*) $ constitutes a 
Nash equilibrium for bilinear game \eqref{eqbilinear}.
\end{proof}

The above proof was established under the assumption that the 
parameters $\gamma$, $\beta$, and $\alpha$ are non-zero. However, in 
the iterative algorithm, we do not set all the step size-related 
parameters to zero. Therefore, we need to consider the correctness of 
the conclusion in the following two scenarios:
(i) When $\alpha\neq 0$, if either $\gamma=0$ or $\beta=0$, it is 
evident from \eqref{eqaa3} and \eqref{eqaa4} that the conclusion 
holds.
(ii) When $\alpha=0$ and both $\gamma$ and $\beta$ are non-zero, from 
\eqref{eqaa3} and \eqref{eqaa4}, we have $A^{\mathrm{T}}A\phi^* =0$ 
and $(A^{\mathrm{T}}A)A^{\mathrm{T}}\theta^* = 0$. By considering the 
positive definiteness of $(A^{\mathrm{T}}A)$ and following the 
reasoning of Proposition \ref{proposition3.1}, we can conclude that 
the conclusion also holds.

\begin{corollary}[] 
	\label{corollary3.1}
	For bilinear game \eqref{eqbilinear}, we assume that matrix $A 
	\in 
	\mathbb{R}^{m \times n}$ and $ rank (A) = m$. If PCAA 
	converges, Then the limit points 
	are Nash equilibria.
\end{corollary}

\begin{proof}
Combining the proof process of Proposition \ref{proposition3.1}  and 
assuming that $\alpha, \gamma, \beta$ are not equal to zero, it can 
be inferred from \eqref{eqaa4} that $A^{\mathrm{T}} \theta^* =  
\frac{\gamma \beta}{\alpha} A^{\mathrm{T}}A\phi^*$.
And substituting it into \eqref{eqaa3}, we have 
\begin{equation}\label{eqaa8}
	-\frac{\gamma^2 \beta^2}{\alpha} AA^{\mathrm{T}}A \phi^*  +  
	 \alpha  A \phi^* 
	= - \bigg( \frac{\gamma^2 \beta^2}{\alpha} 
	AA^{\mathrm{T}} + \alpha I \bigg) A \phi^* =0.
\end{equation}
By \eqref{eqaa8}, it can be known that
\begin{equation}\label{eqaa9}
  (A \phi^*)^{\mathrm{T}} \bigg( \frac{\gamma^2 \beta^2}{\alpha} 
	AA^{\mathrm{T}} + \alpha I \bigg) A \phi^* =0.
\end{equation}
It can be inferred from the fact that matrix A has full row rank that 
$AA^{\mathrm{T}}$ is positive definite. This, combined with 
\eqref{eqaa9}, leads to $A \phi^* =0$. Substituting it into 
\eqref{eqaa4}, 
it can be inferred that $A^{\mathrm{T}} \theta^* =0$. This completes 
the proof.
\end{proof}

Moreover, Goodfellow \cite{goodfellow2016nips} argued that there was 
neither a theoretical argument that GANs should converge with GDA 
when updating the parameters of deep neural networks, nor a 
theoretical argument that the games should not converge. It was 
interesting to note that GANs could be hard to train, and in 
practice, it was often observed that gradient descent-based GANs 
optimization did not lead to convergence \cite{mescheder2018training}.
Indeed, Proposition 1 in the recent work by Daskalakis et al. 
\cite{daskalakis2018training} proved that GDA  applied to  problem 
$\min_{\theta} \max_{\phi} , \theta^{\mathrm{T}} \phi$ 
diverged starting from any initialization $\theta_{0}, \phi_{0}$ such 
that $\theta_{0}, \phi_{0} \neq 0$. However, in this article, PCAA 
was introduced, which showed convergence for bilinear game 
\eqref{eqbilinear} with an appropriate step size.

Then, we prove PCAA enjoy the last-iterate linear convergence 
on bilinear game \eqref{eq3.1}, as follows.

\begin{theorem}[]\label{theorem3.2}
	We assume that matrix $A \in \mathbb{R}^{m \times n}$ in bilinear 
	game \eqref{eq3.1} is a full column rank 
	matrix and $(\theta^*, \phi^*)$ is Nash equilibrium point. And  
	apply Algorithm \eqref{eqpcaa} to it.  
	For any given $\gamma > 0$ chose $\beta >0$, such that $2\gamma 
	\beta \geq 
	\alpha^2$. 
	Then,  the 
	rate is 
	\begin{equation}
		\big\|\theta_t-\theta^* \big\|^2 + 
		\big\|\phi^t-\phi^*\big\|^2 
		\leq \rho^{ 
		t}\Big(\big\|\theta_0-\theta^*\big\|^2+\big\|\phi_0-
	    \phi^*\big\|^2\Big),
	\end{equation}	
    where $ \rho  :=  \max 
   \Big\{ \big(1-\gamma \beta \sigma^2_{\max}(A)\big)^2+\alpha^2 
   \sigma^2_{\max }(A),     \big(1-\gamma \beta \sigma^2_{\min }(A 
   )\big)^2+\alpha^2 \sigma^2_{\min}(A) \Big\} $.
\end{theorem}

\begin{proof}
   For bilinear game  \eqref{eq3.1},  the 
   iterations of predictive step in \eqref{eqpcaa} are simplified to  
	\begin{equation*}
	\begin{aligned} 
		\bigg( 
		\begin{array}{c}  
			\theta_{t+1 / 2 }- \theta^* \\
			\phi_{t+1 / 2 } -\phi*  
		\end{array} \bigg) 
		=\bigg( 
		\begin{array}{c}  
			\theta_{t}- \theta^* \\
			\phi_{t}- \theta^*
		\end{array} \bigg)  +     \bigg( 
		\begin{array}{c}  
			- \gamma A            (\phi_{t} - \theta^*)\\
			\gamma A^{\mathrm{T}} (\theta_{t} - \phi^*)
		\end{array} \bigg) 
		=\bigg( 
		\begin{array}{c}  
			\theta_{t} - \theta^*\\
			\phi_{t}  - \phi^*
		\end{array} \bigg)  + \bigg(
		\begin{matrix}
			0 & -\gamma A \\
			\gamma A^{\mathrm{T}} &0
		\end{matrix} \bigg) 
		\bigg(
		\begin{array}{c}
			\theta_{t} - \theta^*\\ 
			\phi_{t} - \phi^*
		\end{array} \bigg). 
	\end{aligned}	
\end{equation*}		
	Then, the gradient step dynamics in \eqref{eqpcaa} can be 
wirtten as 
\begin{equation*}
	\begin{aligned}
		&\bigg( 
		\begin{array}{c}  
			\theta_{t+1} - \theta^*\\
			\phi_{t+1} - \phi^*  
		\end{array} \bigg) \\
		=& \bigg( 
		\begin{array}{c}  
			\theta_{t} - \theta^*\\
			\phi_{t}  - \phi^* 
		\end{array} \bigg) +    \alpha \bigg( 
		\begin{array}{c}  
			-A (\phi_{t} - \phi^*)\\
			A^{\mathrm{T}} (\theta_{t} - \theta^*)   
		\end{array} \bigg) +\beta \bigg( 
		\begin{array}{c}  
			- A (\phi_{t+1/2} - \phi^* ) -  \big(  -A (\phi_t - 
			\phi^* )  \big)\\
			A^{\mathrm{T}}  (\theta_{t+1 / 2}- \theta^*) -  
			A^{\mathrm{T}} 
			(\theta_t  - \theta^* )
		\end{array} \bigg)  \\
		=& \bigg( 
		\begin{array}{c}  
			\theta_{t} - \theta^*\\
			\phi_{t} - \phi^*
		\end{array} \bigg)  + \bigg(
		\begin{matrix}
			0 & -\alpha A \\
			\alpha A^{\mathrm{T}} &0
		\end{matrix} \bigg) 
		\bigg(
		\begin{array}{c}
			\theta_{t} - \theta^*\\ 
			\phi_{t} - \phi^*
		\end{array} \bigg) +\beta \bigg( 
		\begin{array}{c}  
			- A  \big( (\phi_{t+1 / 2}- \phi^*)  -  (\phi_t - 
			\phi^*)  \big)  \\
			A^{\mathrm{T}}  \big((\theta_{t+1 / 2} - \theta^*)-  
			(\theta_t - \theta^*) \big) 
		\end{array} \bigg) \\
		=& \bigg( 
		\begin{array}{c}  
			\theta_{t} - \theta^*\\
			\phi_{t} - \phi^*
		\end{array} \bigg)  + \bigg(
		\begin{matrix}
			0 & -\alpha A \\
			\alpha A^{\mathrm{T}} &0
		\end{matrix} \bigg) 
		\bigg(
		\begin{array}{c}
			\theta_{t} - \theta^*\\ 
			\phi_{t}- \phi^*
		\end{array} \bigg) +\beta \bigg(
		\begin{matrix}
			0 & -A \\
			A^{\mathrm{T}} &0
		\end{matrix} \bigg)  \bigg(
		\begin{matrix}
			0 & -\gamma A \\
			\gamma A^{\mathrm{T}} &0
		\end{matrix} \bigg) 
		\bigg(
		\begin{array}{c}
			\theta_{t}- \theta^*\\ 
			\phi_{t}- \phi^*
		\end{array} \bigg) \\
		=& \bigg( 
		\begin{array}{c}  
			\theta_{t} - \theta^* \\
			\phi_{t} - \phi^*
		\end{array} \bigg)  + \bigg(
		\begin{matrix}
			0 & -\alpha A \\
			\alpha A^{\mathrm{T}} &0
		\end{matrix} \bigg) 
		\bigg(
		\begin{array}{c}
			\theta_{t} - \theta^*\\ 
			\phi_{t}- \phi^*
		\end{array} \bigg) +  \bigg(
		\begin{matrix}
			-\gamma \beta A A^{\mathrm{T}} & 0 \\
			0 & -\gamma \beta A^{\mathrm{T}} A
		\end{matrix} \bigg) 
		\bigg(
		\begin{array}{c}
			\theta_{t} - \theta^*\\ 
			\phi_{t} - \phi^*
		\end{array} \bigg) \\
		=& \bigg( 
		\begin{array}{c}  
			\theta_{t} - \theta^*\\
			\phi_{t} - \phi^*
		\end{array} \bigg)  + \bigg(
		\begin{matrix}
			-\gamma \beta A A^{\mathrm{T}} & -\alpha A \\
			\alpha A^{\mathrm{T}} & -\gamma \beta A^{\mathrm{T}} A
		\end{matrix} \bigg) 
		\bigg(
		\begin{array}{c}
			\theta_{t} - \theta^*\\ 
			\phi_{t} - \phi^*
		\end{array} \bigg) \\
	=& \bigg(
	\begin{matrix}
		I-\gamma \beta A A^{\mathrm{T}} & -\alpha A \\
		\alpha A^{\mathrm{T}} & I-\gamma \beta A^{\mathrm{T}} A
	\end{matrix} \bigg) 
		\bigg(
    \begin{array}{c}
	\theta_{t} - \theta^*\\ 
	\phi_{t} - \phi^*
    \end{array} \bigg). 
	\end{aligned}
\end{equation*}
For any given matrix $A$, apply SVD decomposition to $A: A = U \Sigma 
V^{\mathrm{T}}$ , where 
$U$ and $V$ are orthogonal and $\Sigma = \mathrm{diag}(\sigma_1, 
\cdots, 
\sigma_n )$. Then

\begin{equation*}
		\Bigg\|\bigg( 
		\begin{array}{c}  
			\theta_{t+1} - \theta^*\\
			\phi_{t+1} - \phi^*  
		\end{array} \bigg) \Bigg\|_2 \leq  \Bigg\|  \bigg(
		\begin{matrix}
			I-\gamma \beta A A^{\mathrm{T}} & -\alpha A \\
			\alpha A^{\mathrm{T}} & I-\gamma \beta A^{\mathrm{T}} A
		\end{matrix} \bigg) \Bigg\|_2
		\Bigg\| \bigg(
		\begin{array}{c}
			\theta_{t} - \theta^*\\ 
			\phi_{t} - \phi^*
		\end{array} \bigg)\Bigg\|_2 .
\end{equation*}
Since $U$ and $V$ are orthogonal, we have
\begin{equation*} 
		 A A^{\mathrm{T}}= U  \Sigma^2 U^{\mathrm{T}}, \;
		A^{\mathrm{T}}A = V  \Sigma^2 V^{\mathrm{T}},
\end{equation*}
and 
\begin{equation}\label{eqaaa5}
	\begin{aligned}
		\Bigg\|\left(\begin{array}{cc}
			 I -\gamma \beta A A^{\mathrm{T}} & -\alpha A \\
			-\alpha A^{\mathrm{T}}  & I -\gamma \beta  A^{\mathrm{T}}A
		\end{array}\right)\Bigg\|_2 & =\Bigg\|\Bigg(\begin{array}{cc}
			U  & 0 \\
			 0 & V
		\end{array}\Bigg) \Bigg(\begin{array}{cc}
			I-\gamma \beta  \Sigma^2 & -\alpha  \Sigma  \\
			 \alpha  \Sigma & I-\gamma \beta  \Sigma^2
		\end{array}\Bigg)\Bigg(\begin{array}{cc}
			U^{\mathrm{T}} & 0 \\
			0 & V^{\mathrm{T}}
		\end{array}\Bigg)\Bigg\|_2 \\
		& =\Bigg\|\Bigg(\begin{array}{cc}
			 I-\gamma \beta  \Sigma^2 & -\alpha  \Sigma \\
			\alpha  \Sigma &  I -\gamma \beta  \Sigma^2
		\end{array}\Bigg)\Bigg\|_2 \\
		& =\max _i \sqrt{\big(1-\gamma \beta 
		\sigma_i^2 \big)^2+\alpha^2 \sigma_i^2} .
	\end{aligned}
\end{equation}
Assume without loss of generality that $\sigma_1, \cdots, \sigma_n $.
For any given $\gamma >0$, chose  $\beta >0$ satisfies $2\gamma \beta 
\geq \alpha^2$, Note that function 
$f(x^2) = 
(1-\gamma 
\beta x^2)^2 + \alpha^2 x^2$
is monotonically
decreasing on $(0,c)$ and monotonically increasing on $(c,
+ \infty)$. 
Consequently, it holds
\begin{equation*}
	\max _i\Big\{ (1-\gamma \beta \sigma_i^2 )^2+\alpha^2 
	\sigma_i^2\Big\}=\max  \Big\{ (1-\gamma \beta \sigma_1^2 
	)^2+\alpha^2 \sigma_1^2, (1-\gamma \beta \sigma_n^2 )^2+\alpha^2 
	\sigma_n^2 \Big \}
\end{equation*}

\end{proof}

The conditions for $\gamma$ and $\beta$ in Theorem \ref{theorem3.2} 
are necessary, but not sufficient. To guarantee convergence, one 
needs to have $\rho < 1$, and below, we provide an example that 
satisfies this condition. 
Under the same assumption as in Theorem \ref{theorem3.2}, if $\gamma 
\beta = 1/ \sigma_{\max}^2(A)$ and $\alpha =0$, and assuming that the 
columns of matrix $A$ are full rank, we have: 
\begin{equation} \label{eqaaa7}
		  \big\|\theta_t - 
		\theta^*\big\|_2^2+\big\|\phi_t-\phi^*\big\|_2^2 \\
		   \leq\Bigg(1-\frac{\sigma^2_{\min }(A)}{ \sigma^2_{\max 
		}(A)}\Bigg)^{2 
		t}\bigg(\big\|\theta_0-\theta^*\big\|_2^2
	    +\big\|\phi_0-\phi^*\big\|_2^2\bigg).
\end{equation}

\begin{remark}\label{reark5.3}	
Notably, when the parameter $\beta$ in PCAA is set to zero, the 
update mechanism of the algorithm becomes equivalent to GDA. As 
mentioned above, to ensure the convergence of the algorithm, we need 
to ensure that $\rho < 1$. According to the proof of Theorem 
\ref{theorem3.2} and the definition of $\rho$, it can be observed 
that for any $\gamma > 0$, if $\beta = 0$, then from  
$\eqref{eqaaa5}$, we have $\rho = 1 + \alpha^2 \sigma^2_{\max(A)}$. 
In this case, regardless of the value of $\alpha$, it is not possible 
to have $\rho < 1$, meaning that GDA will not converge no matter how 
the step size $\alpha$ is chosen. This further provides a theoretical 
answer to Q1 raised in the introduction, and at the same time, it 
also demonstrates the rationality of the design of the PCAA iteration 
format.
\end{remark}

\begin{remark}\label{reark5.4}	
	It is worth noting that literature \cite{mokhtari2020unified} has 
	proven, under the conditions that matrix $A$ is a square matrix 
	of size $m=n$ and has full rank, that when  EG with  step size 
	$\gamma= \beta $ satisfies the condition $\gamma = \beta = 
	1/(2\sqrt{2 
	\lambda(A^{\mathrm{T}}A)})$, it leads to a convergence result of 
	\begin{equation} \label{eqaaa8}
		\big\|\theta_t - 
		\theta^*\big\|_2^2+\big\|\phi_t-\phi^*\big\|_2^2 \\
		\leq\Bigg(1-\frac{1}{20}\frac{\lambda_{\min} 
		(A^{\mathrm{T}}A)} 
		{\lambda_{\max} (A^{\mathrm{T}}A))}\Bigg)^{ 
			t}\bigg(\big\|\theta_0-\theta^*\big\|_2^2
		+\big\|\phi_0-\phi^*\big\|_2^2\bigg).
	\end{equation}
Subsequently, literature \cite{mishchenko2020revisiting} improved the 
convergence result of equation (\ref{eqaaa8}) under the conditions 
that matrix $A$ is a square matrix and has full rank. Specifically, 
when EG step size $\gamma = \beta$ is set to $\gamma = \beta = 
1/(\sqrt{2} \sigma_{\max}(A))$, the iterative sequence satisfies 
	\begin{equation} \label{eqaaa9}
	\big\|\theta_t - 
	\theta^*\big\|_2^2+\big\|\phi_t-\phi^*\big\|_2^2 \\
	\leq\Bigg(1-\frac{1}{6}\frac{\sigma^2_{\min} 
		(A)} 
	{\sigma^2_{\max} (A)}\Bigg)^{ 2
		t}\bigg(\big\|\theta_0-\theta^*\big\|_2^2
	+\big\|\phi_0-\phi^*\big\|_2^2\bigg).
\end{equation}
Furthermore, literature \cite{mishchenko2020revisiting} states that 
for bilinear game when full-rank matrix $A$ satisfies 
$\sigma_{\min}(A) > 0$ and MPM uses a step size of $\gamma = 
 \kappa / (\sqrt{2} \sigma^2_{\max}(A)), \beta = 1 / (\sqrt{2} \kappa 
 \sigma^2_{\max}(A)) $ with $\kappa := 
 \frac{\sigma^2_{\min}(A)}{\sigma^2_{\max}(A)}$, the convergence rate 
 is given 
 by:
\begin{equation} \label{eqaaa10}
	\big\|\theta_t - 
	\theta^*\big\|_2^2+\big\|\phi_t-\phi^*\big\|_2^2 \\
	\leq\Bigg(1-\frac{1}{4}\frac{\sigma^2_{\min} 
		(A)} 
	{\sigma^2_{\max} (A)}\Bigg)^{ 2
		t}\bigg(\big\|\theta_0-\theta^*\big\|_2^2
	+\big\|\phi_0-\phi^*\big\|_2^2\bigg).
\end{equation}
By comparing equations 
\eqref{eqaaa7}, \eqref{eqaaa8}, \eqref{eqaaa9} and \eqref{eqaaa10}, 
it can 
be observed that PCAA achieves better linear convergence performance 
than EG and MPM in bilinear game. This also demonstrates the 
effectiveness of our algorithm framework in incorporating the current 
gradient information during the update step. 
\end{remark} 

\begin{remark}\label{reark5.5}	
It is worth noting that in the special case where matrix $A$ is a 
full-rank square matrix, Theorem \ref{theorem3.2} states that the 
complexity of PCAA  is $\mathcal{O}\big(\kappa 
\log(\frac{1}{\epsilon})\big)$. This result aligns with the optimal 
complexity results for EG and MPM with full-rank square matrix $A$ in 
\cite{mishchenko2020revisiting} and 
\cite{mokhtari2020unified}. 
However, the assumptions of our Theorem 
\ref{theorem3.2} are weaker compared to the assumptions in   
\cite[Theorem 6]{mokhtari2020unified} and \cite[Theorem 
4]{mokhtari2020unified} regarding the matrix $A$. 
Specifically, we do 
not require matrix $A$ to be a square matrix and satisfy the 
full-rank condition.  
One additional point we would like to add is 
that even in the case of a bilinear game where matrix $A$ 
satisfies the condition of being a full-rank square matrix, the 
complexity result in Theorem \ref{theorem3.2} has improved compared 
to the complexity results of MPM in \cite[Theorem 
4]{liang2019interaction} and PCAA in \cite[Theorem 
4.1]{Li2023training} under the same condition, which yielded  a 
complexity of $\mathcal{O}\big(\kappa^{-2} 
\log(\frac{1}{\epsilon})\big)$.	
\end{remark}

\begin{remark} \label{remark5.6}
When the matrix $A \in \mathbb{R}^{m \times n}$ and 
$\gamma=\frac{1}{2\sigma_{\max}(A)}$, Gidel et al. obtained the 
linear convergence result for solving bilinear game \eqref{eq3.1} 
using EG, as stated in \cite[ Corollary 1]{gidel2019a}.
	\begin{equation} \label{eqaaa11}
	\big\|\theta_t - 
	\theta^*\big\|_2^2+\big\|\phi_t-\phi^*\big\|_2^2 \\
	\leq\Bigg(1-\frac{1}{8}\frac{\sigma^2_{\min} 
		(A)} 
	{\sigma^2_{\max} (A)}\Bigg)^{ 
		t}\bigg(\big\|\theta_0-\theta^*\big\|_2^2
	+\big\|\phi_0-\phi^*\big\|_2^2\bigg).
\end{equation}
Comparing the linear convergence results between \eqref{eqaaa11} and 
\eqref{eqaaa7}, it is evident that our linear convergence is superior 
to their results. This further demonstrates the effectiveness of our 
algorithm's iterative scheme.
\end{remark}

Li et al. \cite{Li2023training} did not discuss the lower 
complexity bound of the algorithm. Inspired by recent literature 
\cite{zhang2022on, Ouyang2021lower, liang2019interaction}, we provide 
a lower complexity bound for PCAA on  
bilinear game \eqref{eqbilinear}. 

\begin{proposition}[Simple lower bound for PCAA] \label{proposition2}
	Consider bilinear game \eqref{eqbilinear} with matrix $A \in 
	\mathbb{R}^{m \times n}$,  and $rank (A) =m$. $(\theta^*, 
	\phi^*)$ is Nash equilibrium point, let $\delta 
	=\big\|(\theta_0 - \theta^{*} )^{\mathrm{T}},
	( \phi_0 -
	\phi^{*}  )^{\mathrm{T}} \big\|_2$. Then,  
	PCAA dynamic \eqref{eqpcaa} for any fixed $\gamma >0$ with $	
	0<\beta <\frac{1}  {2\gamma \lambda_{\max}^2  ( A^{\mathrm{T}} A 
		)}$  satisfies 
	\begin{equation*}  \label{eqcorollary1}
		\bigg\|{\bigg(
			\begin{array}{c}
				\theta_{t+1}-\theta^*\\ 
				\phi_{t+1}-\phi^*
			\end{array} \bigg)}\bigg\|_2^2  \geq 
		\bigg( 1- 2\gamma\beta \lambda_{\max} ( 
			A^{\mathrm{T}} A
			) \bigg)
		\bigg\|{ \bigg(
			\begin{array}{c}
				\theta_{t}-\theta^*\\ 
				\phi_{t}-\phi^*
			\end{array} \bigg)}\textbf{}\bigg\|_2^2. 
	\end{equation*}
\end{proposition}

\begin{proof}
	For bilinear game  \eqref{eqbilinear},  the 
	iterations of predictive step in \eqref{eqpcaa} are simplified to
	\begin{equation*} 
		\begin{aligned} 
			\bigg( 
			\begin{array}{c}  
				\theta_{t+1 / 2} \\
				\phi_{t+1 / 2}   
			\end{array} \bigg) 
        		=\bigg( 
        \begin{array}{c}  
        	\theta_{t} \\
        	\phi_{t}
        \end{array} \bigg)  + \bigg(
        \begin{matrix}
        	0 & -\gamma A \\
        	\gamma A^{\mathrm{T}} &0
        \end{matrix} \bigg) 
        \bigg(
        \begin{array}{c}
        	\theta_{t}\\ 
        	\phi_{t}
        \end{array} \bigg). 	
    \end{aligned}	
\end{equation*}		
Then, the gradient step dynamics in \eqref{eqpcaa} can be written 
as 	
	\begin{equation}\label{eqcorollary3}
	\begin{aligned}
		\bigg( 
		\begin{array}{c}  
			\theta_{t+1} \\
			\phi_{t+1}   
		\end{array} \bigg)
		&= \bigg( 
		\begin{array}{c}  
			\theta_{t} \\
			\phi_{t}   
		\end{array} \bigg) +    \alpha \bigg( 
		\begin{array}{c}  
			-A \phi_{t} \\
			A^{\mathrm{T}} \theta_{t}   
		\end{array} \bigg) +\beta \bigg( 
		\begin{array}{c}  
			- A  \phi_{t+1 / 2}  -    (-A \phi_t  )  \\
			A^{\mathrm{T}}  \theta_{t+1 / 2} -  A^{\mathrm{T}} 
			\theta_t   
		\end{array} \bigg)  \\
		&= \bigg( 
		\begin{array}{c}  
			\theta_{t} \\
			\phi_{t}
		\end{array} \bigg)  + \bigg(
		\begin{matrix}
			-\gamma \beta A A^{\mathrm{T}} & -\alpha A \\
			\alpha A^{\mathrm{T}} & -\gamma \beta A^{\mathrm{T}} A
		\end{matrix} \bigg) 
		\bigg(
		\begin{array}{c}
			\theta_{t}\\ 
			\phi_{t}
		\end{array} \bigg) \\
		&= \bigg(
		\begin{matrix}
			I -\gamma \beta A A^{\mathrm{T}} & -\alpha A \\
			\alpha A^{\mathrm{T}} & I -\gamma \beta 
			A^{\mathrm{T}} A
		\end{matrix} \bigg) 
		\bigg(
		\begin{array}{c}
			\theta_{t}\\ 
			\phi_{t}
		\end{array} \bigg). 
	\end{aligned}
\end{equation}

Let's analyze the singular values of the operator 
$
K :=\Bigg( I-\bigg(
\begin{matrix}
	\gamma \beta A A^{\mathrm{T}} & \alpha A \\
	-\alpha A^{\mathrm{T}} & \gamma \beta A^{\mathrm{T}} A
\end{matrix} \bigg) \Bigg)
$ in \eqref{eqcorollary3}, 
or equivalently, the eigenvalues of $ KK^{\mathrm{T}}$,
\begin{equation}\label{eqcorollary4}
	\begin{aligned}
		KK^{\mathrm{T}}
				= \Bigg(
		\begin{matrix}
			 (I -\gamma \beta A A^{\mathrm{T}}  )^2 + 
			\alpha^2 A A^{\mathrm{T}}  & 0
			\\
			0 & (I -\gamma \beta  A^{\mathrm{T}} A  )^2 
			+ \alpha^2 A^{\mathrm{T}} A
		\end{matrix} \Bigg). 
	\end{aligned}
\end{equation}  
For any given predictive rate $\gamma$, when the learning rate 
$\alpha$ and adjustment rate $\beta$ are appropriately chosen, 
consider the smallest eigenvalue of the matrix $  (I -\gamma \beta A 
A^{\mathrm{T}}  )^2 + \alpha^2 A 
A^{\mathrm{T}} $ in \eqref{eqcorollary3}. 
By performing the singular value decomposition of matrix $A$ (i.e., 
$A = UDV$), we obtain
\begin{equation}\label{eqcorollary5}
	\begin{aligned}
		&  (I -\gamma \beta A A^{\mathrm{T}} )^2 + 
		\alpha^2 A A^{\mathrm{T}}
		= U\big(  (I -\gamma \beta D^2 )^2 + \alpha^2 
		D^2 \big) 
		U^{\mathrm{T}} \\
		\succeq & \Big[1- 2\gamma  \lambda_{\max} (A 
		A^{\mathrm{T}}  ) \beta  + 
		\gamma^2\lambda_{\min}^2 (A A^{\mathrm{T}}  ) 
		\beta^2+ \alpha^2 \lambda_{\min} (A A^{\mathrm{T}} 
		 ) \Big] I, 
	\end{aligned}
\end{equation}
Next, we consider the minimum eigenvalue of matrix $(I -\gamma \beta  
A^{\mathrm{T}} A  )^2 + \alpha^2 A^{\mathrm{T}} A$ in 
\eqref{eqcorollary5}. Using the same 
singular value decomposition as above, we have
\begin{equation}\label{eqcorollary6}
	\begin{aligned}
		  (I -\gamma \beta A^{\mathrm{T}}A  )^2 + 
		\alpha^2  A^{\mathrm{T}}A
		&= V^{\mathrm{T}} \big(  (I -\gamma \beta D^2 )^2 + \alpha^2 
		D^2 \big) V 
		 \\
		&\succeq  \Big[1- 2\gamma \beta  \lambda_{\max} ( 
	    A^{\mathrm{T}} A  )  \Big] I, 
	\end{aligned}
\end{equation}
with 
\begin{equation*} 
	0<\beta <\frac{1}  {2\gamma \lambda_{\max}^2  ( A^{\mathrm{T}} A 
	)}.
\end{equation*}
The last equation in \eqref{eqcorollary6} is derived from the 
positive semi definiteness 
of the matrix $A^{\mathrm{T}}A$ and the fact that the smallest 
eigenvalue $\lambda_{\min} ( A^{\mathrm{T}}A  ) = 0$. Comparing the 
inequalities in \eqref{eqcorollary5} and \eqref{eqcorollary6}, it can 
be concluded that due to the positive definiteness of matrix 
$AA^{\mathrm{T}}  $, the right-hand side of inequality 
\eqref{eqcorollary6} is smaller. 
Hence, by the Rayleigh quotient, \eqref{eqcorollary3} and 
\eqref{eqcorollary4}, we have
\begin{equation*} \label{eqcorollary7}  
	\bigg\|{\bigg(
		\begin{array}{c}
			\theta_{t+1} -\theta^*\\ 
			\phi_{t+1}-\phi^*
		\end{array} \bigg) 
	}\bigg\|_2^2 = {\bigg(
		\begin{array}{c}
			\theta_{t}-\theta^*\\ 
			\phi_{t}-\phi^*
		\end{array} \bigg)^{\mathrm{T}} }  K^{\mathrm{T}} K \bigg(
	\begin{array}{c}
		\theta_{t}-\theta^*\\ 
		\phi_{t}-\phi^*
	\end{array} \bigg) \geq 
	\lambda_{\min} ( K^{\mathrm{T}} K)
	\bigg \|{\bigg(
		\begin{array}{c}
			\theta_{t}-\theta^*\\ 
			\phi_{t}-\phi^*
		\end{array} \bigg) } \bigg \|_2^2, 
\end{equation*}
which together with \eqref{eqcorollary4} and \eqref{eqcorollary5}, we 
deduce 
that	
\begin{equation}  \label{eqcorollary10}
	\bigg\|{\bigg(
		\begin{array}{c}
			\theta_{t+1}-\theta^*\\ 
			\phi_{t+1}-\phi^*
		\end{array} \bigg)}\bigg\|_2  \geq 
	\sqrt{1- 2\gamma\beta \lambda_{\max} ( 
			A^{\mathrm{T}} A
			 )}
	\bigg\|{\bigg(
		\begin{array}{c}
			\theta_{t}-\theta^*\\ 
			\phi_{t}-\phi^*
		\end{array}  \bigg)}\bigg\|_2. 
\end{equation}
Using recurrence \eqref{eqcorollary10}, we obtain
\begin{equation*}  
	\bigg\|{\bigg(
		\begin{array}{c}
			\theta_{t+1}-\theta^*\\ 
			\phi_{t+1}-\phi^*
		\end{array} \bigg)}\bigg\|_2  \geq 
	\Bigg(\sqrt{1- 2\gamma\beta \lambda_{\max} ( 
		A^{\mathrm{T}} A
		)} \Bigg)^t
	\bigg\|{\bigg(
		\begin{array}{c}
			\theta_{0} -\theta^*\\ 
			\phi_{0}-\phi^*
		\end{array} \bigg)}\bigg\|_2. 
\end{equation*}
Based on the assumption that  $\big\|(\theta_0 - \theta^{*} 
)^{\mathrm{T}},
( \phi_0 -
\phi^{*}  )^{\mathrm{T}} \big\|_2 = \delta$, and the fact that 
$\forall  ( x, \xi ) \in 
\mathbb{R} \times \mathbb{R}^+, (1-x)^\xi \leq e^{-\xi x }$, by 
setting the right-hand side of the equation 
equal to $\epsilon$, we can obtain a lower bound for game 
\eqref{eqbilinear}  
solved 
by PCAA as follows:
\begin{equation*} 
	\Omega\bigg(\frac{1}{\gamma \beta 
	\lambda_{\max}^2(A^{\mathrm{T}}A)} 
	\log(\frac{\delta}{\epsilon})
	\bigg).
\end{equation*}
This completes the proof.		 
\end{proof}

\begin{corollary} \label{corollary2}
	Consider bilinear game \eqref{eqbilinear} with matrix $A \in 
	\mathbb{R}^{n \times n}$,  and $rank (A) =n$. $(\theta^*, 
	\phi^*)$ is Nash equilibrium point, let 
	$\delta 
	=\big\|(\theta_0 - \theta^{*} )^{\mathrm{T}},
	( \phi_0 -
	\phi^{*}  )^{\mathrm{T}} \big\|_2$. Then,  
	PCAA dynamics \eqref{eqpcaa} for any fixed $\gamma$ with $\beta = 
	\frac{\lambda_{\max}(AA^{\mathrm{T}})}{\gamma 
		\lambda_{\min}^2(AA^{\mathrm{T}})} $ and 
	$ 
	\frac{\lambda_{\max}^2(AA^{\mathrm{T}})}{
		\lambda_{\min}^3(AA^{\mathrm{T}})}  > \alpha^2 \geq 
	\frac{\lambda_{\max}^2(AA^{\mathrm{T}})-\lambda_{\min}^2
		(AA^{\mathrm{T}})}{		 
		\lambda_{\min}^3(AA^{\mathrm{T}})}$ satisfies 
	\begin{equation*}  \label{eqcoro1}
		\bigg\|{\bigg(
			\begin{array}{c}
				\theta_{t+1}-\theta^*\\ 
				\phi_{t+1}-\phi^*
			\end{array} \bigg)}\bigg\|_2^2  \geq 
		\Bigg( {1- \bigg(\frac{\lambda_{\min}^2 (A 
				A^{\mathrm{T}} ) -\alpha^2 
				\lambda_{\max}^3 (A 
				A^{\mathrm{T}}  ) }{\lambda_{\max}^2 (A 
				A^{\mathrm{T}} 
				)} \bigg)} \Bigg)
		\bigg\|{ \bigg(
			\begin{array}{c}
				\theta_{t}-\theta^*\\ 
				\phi_{t}-\phi^*
			\end{array} \bigg)}\textbf{}\bigg\|_2^2. 
	\end{equation*}
\end{corollary}

\begin{proof}
	Building on the analysis discussed in Proposition 
	\ref{proposition2}, let's analyze the singular values of the 
	operator 
	$
	K :=\Bigg( I-\bigg(
	\begin{matrix}
		\gamma \beta A A^{\mathrm{T}} & \alpha A \\
		-\alpha A^{\mathrm{T}} & \gamma \beta A^{\mathrm{T}} A
	\end{matrix} \bigg) \Bigg)
	$, 
	or equivalently, the eigenvalues of $ KK^{\mathrm{T}}$,
	\begin{equation}\label{eqcoro4}
		\begin{aligned}
			KK^{\mathrm{T}}
			= \Bigg(
			\begin{matrix}
				(I -\gamma \beta A A^{\mathrm{T}}  )^2 + 
				\alpha^2 A A^{\mathrm{T}}  & 0
				\\
				0 & (I -\gamma \beta  A^{\mathrm{T}} A  )^2 
				+ \alpha^2 A^{\mathrm{T}} A
			\end{matrix} \Bigg). 
		\end{aligned}
	\end{equation}  
	For any given predictive rate $\gamma$, when the learning rate 
	$\alpha$ and adjustment rate $\beta$ are appropriately chosen, 
	consider the smallest eigenvalue of the matrix $  (I -\gamma 
	\beta A 
	A^{\mathrm{T}}  )^2 + \alpha^2 A 
	A^{\mathrm{T}} $. 
	By performing the singular value decomposition of matrix $A$ 
	(i.e., 
	$A = UDV$), we obtain
	\begin{equation}\label{eqcoro5}
		\begin{aligned}
			&  (I -\gamma \beta A A^{\mathrm{T}} )^2 + 
			\alpha^2 A A^{\mathrm{T}}
			= U\big(  (I -\gamma \beta D^2 )^2 + \alpha^2 
			D^2 \big) 
			U^{\mathrm{T}} \\
			\succeq & \Big[1- 2\gamma  \lambda_{\max} (A 
			A^{\mathrm{T}}  ) \beta  + 
			\gamma^2\lambda_{\min}^2 (A A^{\mathrm{T}}  ) 
			\beta^2+ \alpha^2 \lambda_{\min} (A A^{\mathrm{T}} 
			) \Big] I \\
			=& \Bigg[1- \bigg(\frac{\lambda_{\max}^2 (A 
				A^{\mathrm{T}} ) -\alpha^2 
				\lambda_{\min}^3 (A 
				A^{\mathrm{T}}  ) }{\lambda_{\min}^2 (A 
				A^{\mathrm{T}}  )} \bigg) \Bigg] I,
		\end{aligned}
	\end{equation}
	with 
	\begin{equation*}    
		\frac{\lambda_{\max}^2(AA^{\mathrm{T}})}{
			\lambda_{\min}^3(AA^{\mathrm{T}})}  > \alpha^2 \geq 
		\frac{\lambda_{\max}^2(AA^{\mathrm{T}})-\lambda_{\min}^2
			(AA^{\mathrm{T}})}{		 
			\lambda_{\min}^3(AA^{\mathrm{T}})},
	\end{equation*} and 
	\begin{equation*}
		\beta = \frac{ \lambda_{\max} (A A^{\mathrm{T}} )} 
		{\gamma \lambda_{\min}^2  (A A^{\mathrm{T}} )}.
	\end{equation*}
	By the Rayleigh quotient, \eqref{eqcoro4} and 
	\eqref{eqcoro5}, we have
	\begin{equation*} \label{eqcoro8}  
		\bigg\|{\bigg(
			\begin{array}{c}
				\theta_{t+1} -\theta^*\\ 
				\phi_{t+1}-\phi^*
			\end{array} \bigg) 
		}\bigg\|_2^2 = {\bigg(
			\begin{array}{c}
				\theta_{t}-\theta^*\\ 
				\phi_{t}-\phi^*
			\end{array} \bigg)^{\mathrm{T}} } K K^{\mathrm{T}} \bigg(
		\begin{array}{c}
			\theta_{t}-\theta^*\\ 
			\phi_{t}-\phi^*
		\end{array} \bigg) \geq 
		\lambda_{\min} (K K^{\mathrm{T}} )
		\bigg \|{\bigg(
			\begin{array}{c}
				\theta_{t}-\theta^*\\ 
				\phi_{t}-\phi^*
			\end{array} \bigg) } \bigg \|_2^2, 
	\end{equation*}
	which together with \eqref{eqcoro4} and 
	\eqref{eqcoro5}, we 
	deduce 
	that	
	\begin{equation}  \label{eqcoro9}
		\bigg\|{\bigg(
			\begin{array}{c}
				\theta_{t+1}-\theta^*\\ 
				\phi_{t+1}-\phi^*
			\end{array} \bigg)}\bigg\|_2  \geq 
		\sqrt{1- \bigg(\frac{\lambda_{\max}^2 (A 
				A^{\mathrm{T}} ) -\alpha^2 
				\lambda_{\min}^3 (A 
				A^{\mathrm{T}}  ) }{\lambda_{\min}^2 (A 
				A^{\mathrm{T}} 
				)} \bigg)}
		\bigg\|{\bigg(
			\begin{array}{c}
				\theta_{t}-\theta^*\\ 
				\phi_{t}-\phi^*
			\end{array}  \bigg)}\bigg\|_2. 
	\end{equation}
	Using recurrence \eqref{eqcoro9}, we obtain
	\begin{equation*}  
		\bigg\|{\bigg(
			\begin{array}{c}
				\theta_{t+1}-\theta^*\\ 
				\phi_{t+1}-\phi^*
			\end{array} \bigg)}\bigg\|_2  \geq 
		\Bigg(\sqrt{1- \bigg(\frac{\lambda_{\max}^2 (A 
				A^{\mathrm{T}} ) -\alpha^2 
				\lambda_{\min}^3(A 
				A^{\mathrm{T}} ) }{\lambda_{\min}^2 (A 
				A^{\mathrm{T}} 
				)} \bigg)} \Bigg)^t
		\bigg\|{\bigg(
			\begin{array}{c}
				\theta_{0} -\theta^*\\ 
				\phi_{0}-\phi^*
			\end{array} \bigg)}\bigg\|_2. 
	\end{equation*}
	Based on the assumption that  $\big \| (\theta_0 - \theta^{*} 
	)^{\mathrm{T}},
		( \phi_0 -
		\phi^{*}  )^{\mathrm{T}} \big \|_2 
	= \delta$, and the fact that $\forall  ( x, \xi ) \in 
	\mathbb{R} \times \mathbb{R}^+, (1-x)^\xi \leq e^{-\xi x }$, by 
	setting the right-hand side of the equation 
	equal to $\epsilon$, we can obtain a lower bound for game 
	\eqref{eqbilinear}  
	solved 
	by PCAA as follows:
	\begin{equation*} 
		\Omega\bigg(\frac{		
			\lambda_{\min}^2(AA^{\mathrm{T}})}{\lambda_{\max}^2(AA^{\mathrm{T}})-
			\alpha^2 \lambda_{\min}^3(AA^{\mathrm{T}})} 
		\log(\frac{1}{\epsilon})
		\bigg).
	\end{equation*}
	This completes the proof.		 
\end{proof} 

The above Proposition \ref{proposition2} shows that for any $\gamma$, 
to 
obtain 
$\epsilon$-solution, the number of PCAA iteration is at least 
$\Omega\Big(\frac{		
	\lambda_{\min}^2(AA^{\mathrm{T}})}{\lambda_{\max}^2(AA^{\mathrm{T}})-
	\alpha^2 \lambda_{\min}^3(AA^{\mathrm{T}})} 
	\log(\frac{1}{\epsilon}) 
\Big)$. 

\section{Applications of predictive centripetal acceleration 
	algorithm}

\subsection{PCAA with Anderson mixing} \label{sectionPCAA-AM}

In 2022, He et al. \cite{he2022solve} utilized the dynamic 
information of GDA iterations and proposed a novel minimax optimizer 
called GDA-AM by cleverly combining past iteration points. This 
optimizer addresses the shortcomings of GDA divergence on certain 
minimax optimization problems and also accelerates the convergence of 
AGDA.
Noting that the GDA iterative format is a specific case of the PCAA 
iterative format, in light of this, we raise the following questions:
{\it Q1: How does PCAA combine with Anderson mixing?
	Q2: How does the new algorithm PCAA-AM perform in terms of 
	iteration 
	number and running time when PCAA is combined with Anderson 
	mixing?}

Next, we answer the first question. Obviously, one way to combine 
PCAA with Anderson mixing is to replace the update step of GDA with 
the overall prediction step and update step of PCAA. The specific 
form can be found in Algorithm \ref{PCAA-AM}. This approach has 
two advantages: firstly, it has a relatively simple form, and 
secondly, it preserves the property that PCAA can reduce to  
GDA, meaning that PCAA-AM can also reduce to GDA-AM. Specifically, 
when the parameter $\gamma$ of the PCAA-AM algorithm is set to $0$, 
the iterative format of the algorithm is equivalent to GDA-AM. In 
this case, we can obtain the same numerical performance as the GDA-AM 
algorithm. Theoretically, GDA-AM provides a lower bound for the 
numerical performance of the PCAA-AM algorithm.
For the second question, we will answer it by conducting comparative 
experiments in the numerical experiments section as described in 
subsection \ref{Numerical-PCAA-AM}.

\begin{algorithm}[h!] 
	\footnotesize
	\caption{PCAA-AM: Predictive centripetal acceleration algorithm 
	with Anderson mixing}
	\label{PCAA-AM}
	{\bf Require:}  Initial parameters $\theta_0, \phi_0 $, 
	predictive rate $\gamma$, step-size parameters $\alpha$, adaptive 
	rate $\beta$, and Anderson table size $p$;\\
	{\bf Set:} $w_0=\left[\theta_0, 
	\phi_0\right]$, 
	${\mathrm {sx}}= {\mathrm {length}} (\theta_0 )$; 
	\begin{algorithmic}[1]
		\WHILE {$t = 0, 1, 2, \cdots, $  } 
		\STATE  $\mathbf{\theta}_t, 
		\mathbf{\phi}_t = w_t[0: \mathrm{sx}-1], 
		w_t[\mathrm{sx}: \mathrm{end}]$; \
		\STATE $\theta_{t+1/2} =\theta_t-\gamma 
		\nabla_{\theta} V  (\theta_t, \phi_t )$; \
		\STATE $\phi_{t+1/2}=\phi_t -\gamma \nabla_{\phi} 
		V  (\theta_t, \phi_t )$; \
		\STATE $\theta_{t+1} 
		= \theta_{t} - \alpha \nabla_{\theta} V  (\theta_{t}, 
		\phi_{t} ) - \beta \big( \nabla_{\theta}   V 
		 (\theta_{t+1 / 2}, \phi_{t+1 / 2}  ) -  \nabla 
		_{\theta}	V  (\theta_{t}, \phi_{t} )  \big)$; \
		\STATE $\phi_{t+1} = \phi_{t} + \alpha \nabla_{\phi} V 
	    (\theta_{t}, \phi_{t} ) + \beta 
		\big(\nabla_{\phi} V  (\theta_{t+1 / 2}, \phi_{t+1 / 
			2} ) -  \nabla_{\phi} V  (\theta_t, \phi_t 
		 ) \big)$; \
		\STATE $w_{t+1} = \bigg[ \begin{array}{l} 
			\theta_{t+1}  \\
			\phi_{t+1} 
		\end{array} \bigg] $;\ 
		\STATE { Use Anderson mixing with table size $p$ to 
			extrapolate $w_{t+1}$}; (The algorithm for Aderson mixing 
		can 
		be found in \cite[algorithm 1]{he2022solve})\
		\ENDWHILE
		\STATE $\theta_t, \phi_t=w_{t+1}[0: \mathrm{sx}-1], 
		w_{t+1}[\mathrm{sx}: \mathrm{end}]$;\
		\RETURN $\theta_t, \phi_t$. \
	\end{algorithmic}
\end{algorithm}

\subsection{Adam with predictive centripetal acceleration} 
\label{PCAA-Adamp}

It is noteworthy that recent literature, such as 
\cite{mertikopoulos2019optimistic, daskalakis2018training, 
mertikopoulos2018cycles, chavdarova2021taming, mescheder2018training} 
and, 
shares a common idea. The idea is to make improvements to the 
original algorithms under more restrictive assumptions as a 
principled approach to training GANs. Subsequently, these 
improvements are tested to see if they can be extended to  more 
general learning environments. This approach proves to be 
beneficial.
The purpose of this subsection is similar to the references 
\cite{gidel2019a, ryu2019ode, gidel2020multi, 
	mertikopoulos2019optimistic, daskalakis2018training, 
	he2022solve}. Our goal is not to achieve the best IS or FID 
values by improving the GANs network architecture or modifying 
the GANs loss function. Instead, we aim to optimize the training 
of standard GANs using the optimization techniques, which are 
supported by theoretical foundations in the bilinear case, 
introduced earlier for better optimization. These 
optimization techniques are separate from the improvements in GANs 
network architecture and the setting of loss functions, thus they 
can be applied to train any type of GANs. 

Due to the success of 
using Adam for training Wasserstein GAN  compared to SGD, Daskalakis 
et al. \cite{daskalakis2018training} proposed an optimistic version 
of Adam called Optimistic-Adam by combining their algorithm OGDA with 
Adam. They recommended using 
Optimistic-Adam instead of directly using OGDA to train Wasserstein 
GAN for generating images. 
Subsequently, Gidel et al. \cite{gidel2019a} proposed the 
``extrapolation from past" algorithm, which is equivalent to 
OGDA in the unconstrained setting and involves storing and 
reusing extrapolated gradients for extrapolation. Then, they 
discussed the convergence of the proposed algorithm by leveraging 
the monotonicity of variational inequalities. Furthermore, they 
combined the re-extra gradient algorithm with Adam and applied it 
to practical GANs training.
Recently, Mertikopoulos et al. further noted that the fundamental 
idea of mirror descent is to obtain a new state by performing a 
mirror step along a direction similar to the gradient from the 
initial state. Inspired by the concept of generating intermediate 
points using extra gradient techniques and returning to the 
current point for updates, they proposed the optimistic mirror 
descent algorithm for saddle point problems and discussed its 
convergence using variational inequality tools.Subsequently, they 
incorporated the extra-gradient techniques into Adam and proposed 
the Extra-Adam algorithm for practical GANs training. Through 
experiments, they 
discovered that incorporating extra gradient techniques 
effectively stabilizes the training process of GANs. 
Additionally, their experiments illustrate that the first and 
second gradient steps of their algorithm are more effective when 
using two sets of different moment estimates. 

Noticeably, PCAA we obtained on the general bilinear game can 
reduce to EG and GDA. In light of this, we present the 
following two questions: {\it
	Q1: How can we integrate the PCAA algorithm into the Adam 
	algorithm while ensuring the preservation of the degradable 
	relationship?
	Q2: How does the numerical performance of our newly proposed Adam 
	algorithm with integrated PCAA fare?}

Next, let's address the first question. By observing the 
relationship between PCAA and the iterative format of EG, and 
drawing inspiration from the construction ideas of optimistic-Adam 
and Extra-Adam, it is evident that one possible way to integrate 
PCAA into Adam is to incorporate extrapolation information into 
the update step of Extra-Adam, for more details, please refer to 
Algorithm \ref{PCAA-adam}. This approach is both 
straightforward and ensures that PCAA-Adam can reduce to 
Extra-Adam.  
Clearly, under the condition of other factors being unchanged, 
when the algorithm uses extrapolation from the current step and 
the parameters $\gamma \neq 0, \alpha=\beta \neq 0$, PCAA-Adam can 
reduce to Extra-Adam as described in \cite{mescheder2018training}. If 
$\gamma \neq 0$, the output of 
the PCAA-Adam algorithm is the same as Adam. With the existence of 
such relationship, from a set-theoretic perspective, theoretically 
PCAA-Adam can be applied to any deep learning task that utilizes 
Adam and Extra-Adam. Moreover, by only modifying the algorithm 
while keeping other conditions unchanged, both Adam and Extra-Adam 
provide lower bounds for the performance of PCAA-Adam.
The answer to the second question is provided in 
subsection \ref{WGAN-GP6.2}.

\begin{algorithm}[h!] 
	\footnotesize
	\caption{PCAA-Adam: proposed Adam with predictive centripetal 
	acceleration}
	\label{PCAA-adam}
	{\bf Require:}  Predictive rate $\gamma$, step size parameters 
	$\alpha$ and adaptive rate $\beta$, 
	exponential decay rates for moment estimates $\beta_1, \beta_2 
	\in 
	[0, 1) $, access to the stochastic gradients  $\nabla 
	\ell_{t}(\cdot)$, initial parameters $\omega_0, m_0, 
	v_0$;\  
	\begin{algorithmic}[1]
		\WHILE {$t = 1, \cdots, T$  } 
		\STATE  {\bf Option 1: Standard extrapolation;} \
		\STATE  \qquad Sample new mini-batch and compute 
		stochastic gradient: 
		$g_{t-1} \leftarrow 
		\nabla \big(\ell_{t-1}(\omega_{t-1}) \big)$;\
		\STATE  {\bf Option 2: Extrapolation from the past.} \
		\STATE \qquad Load previously saved stochastic 
		gradient: $g_{t-1} \leftarrow 
		\nabla \big(\ell_{t-3/2}(\omega_{t-3/2}) \big)$; \
		\STATE Update estimate of first moment for 
		extrapolation: 
		$m_{t-1/2} \leftarrow \beta_1 \cdot  m_{t-1} + 
		(1- \beta_1) \cdot g_{t-1}$;\
		\STATE Update estimate of second moment for 
		extrapolation: $v_{t-1/2} \leftarrow \beta_2\cdot v_{t-1} + 
		(1-\beta_2)\cdot g_{t-1}^2$;\
		\STATE Correct the bias for first moments:  
		$\widehat{m}_{t-1/2} \leftarrow m_{t-1/2} 
		/(1-\beta_{1}^{t} )$;\ 
		\STATE Correct the bias for second moments: $\widehat{v}_{t} 
		\leftarrow v_{t-1/2} 
		/(1-\beta_{2}^{t} )$;\
		\STATE Perform extrapolation step at time 
		$t-1$:  
		{  
			$\theta_{t-1/2} 
			\leftarrow 
			\theta_{t-1}-\gamma 
			\cdot \frac{\widehat{m}_{t-1/2} }
			{\sqrt{\widehat{v}_{t-1/2}} + \epsilon} $};\
		\STATE Sample new mini-batch and compute stochastic gradient: 
		$g_{t-1/2} \leftarrow 
		\nabla \big(\ell_{t-1/2}(\omega_{t-1/2}) \big)$;\
		\STATE Update estimate of first moment: $m_{t} \leftarrow 
		\beta_1 \cdot m_{t-1/2} + 
		(1- 
		\beta_1) \cdot g_{t-1/2}$;\
		\STATE Update estimate of second moment: $v_{t} \leftarrow 
		\beta_2 \cdot v_{t-1/2} + 
		(1-\beta_2) \cdot g_{t-1/2}^2$;\
		\STATE Correct the bias for first moments: $\widehat{m}_{t} 
		\leftarrow m_{t-1/2} 
		/(1-\beta_{1}^{t} )$;\
		\STATE Correct the bias for second moments: $\widehat{v}_{t} 
		\leftarrow v_{t-1/2} 
		/(1-\beta_{2}^{t} )$;\
		\STATE Perform update step at time $t-1$: {  
			$\theta_{t} \leftarrow 
			\theta_{t-1}-\alpha 
			\cdot \frac{\widehat{m}_{t-1/2} }
			{ \sqrt{\widehat{v}_{t-1/2}}+\epsilon } - \beta \cdot 
			\big( 
			\frac{\widehat{m}_{t} }
			{ \sqrt{\widehat{v}_{t}}+\epsilon } - 
			\frac{\widehat{m}_{t-1/2} }
			{ \sqrt{\widehat{v}_{t-1/2}}+\epsilon }  \big)$};\
		\ENDWHILE
		\RETURN $\theta_{T-1/2}$ or $\theta_T$. \
	\end{algorithmic}
\end{algorithm}

\section{Numerical simulation}

\subsection{PCAA-AM on bilinear problems} \label{Numerical-PCAA-AM}

We answer the second question in subsection \ref{sectionPCAA-AM} by 
comparing the numerical performance 
of PCAA and PCAA-AM with other algorithms on the examples provided in 
Subsection 5.1 of \cite{he2022solve}. These algorithms 
include GDA, AGDA, OGDA, EG, EG with negative momentum (EG-NM), EG 
with positive momentum (EG-PM), GDA-AM, and AGDA-AM. The experimental 
setup is the same as that in \cite{he2022solve}, where the 
values of A, b, c, and initial point are generated using normally 
random numbers. The maximum number of iterations is set to $5 \times 
10^4$, the stopping criterion is set to $1 \times 10^5$, and the 
convergence is characterized using the norm of distance between  
iteration point and optima. For more detailed information on the 
experimental setup, please refer to \cite{he2022solve}.

\begin{figure}[h!]  
	\centering
	\subfloat[$n=100$]{\includegraphics[width=151 pt, height =140 
		pt]{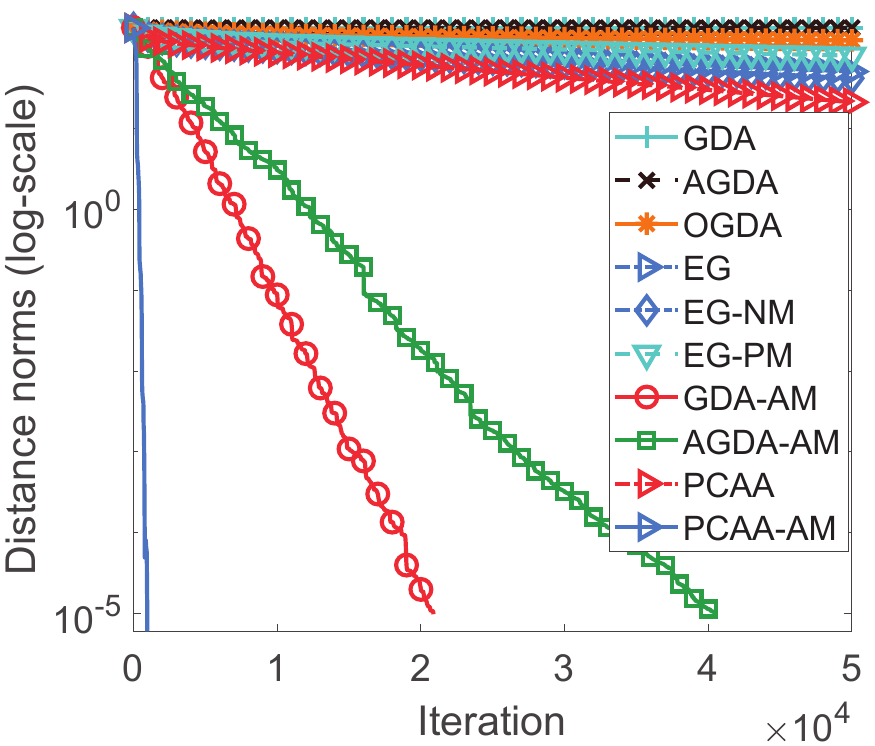}}
	\subfloat[$n=500$]{ 
	\includegraphics[width=151 pt, height =140 
	pt]{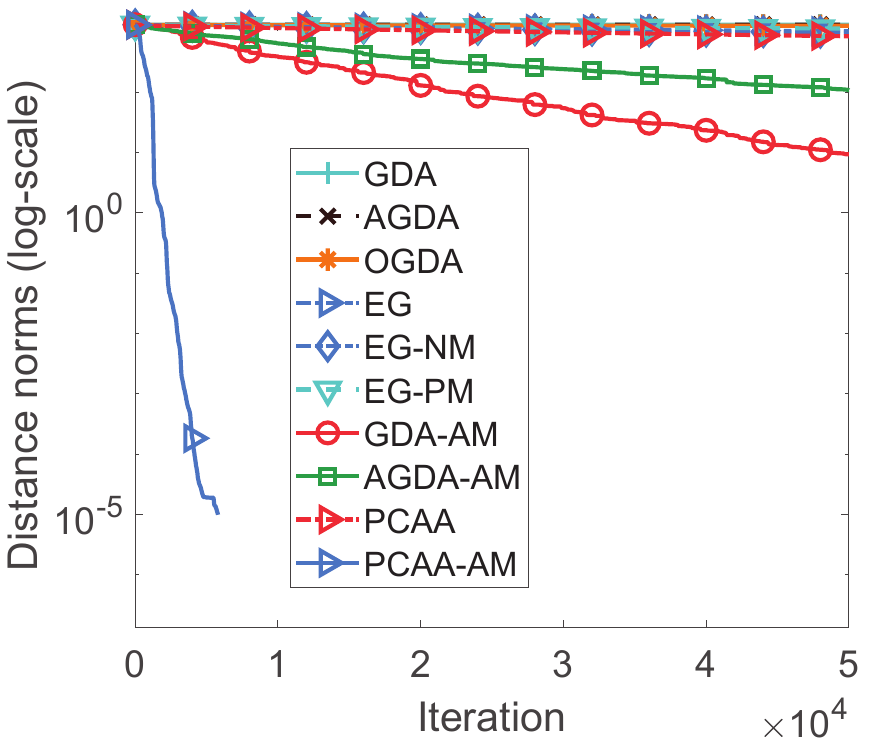}}
	\subfloat[$n=1000$]{\includegraphics[width=151 pt, height =140 
		pt]{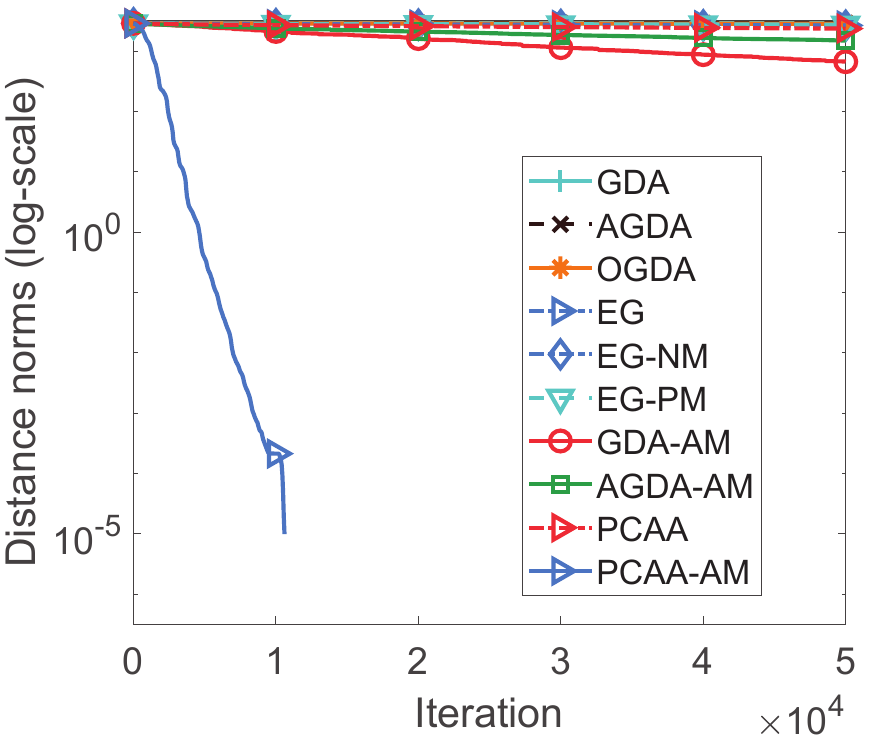}}
	\caption{Comparison in terms of iteration on games $\min 
		_{\theta} \max_{\phi} V(\theta,\phi)
		=	\theta^{\mathrm{T}} A \phi+\theta^{\mathrm{T}} 
		b+c^{\mathrm{T}} \phi$. Except for PCAA and PCAA-AM using 
		parameters $\gamma=1,\alpha=0, \beta=1$ in subfigures (a) and 
		(b), and 
		using parameters $\alpha=0, \gamma=1, \beta=2$ in subfigure 
		(c), the parameter settings for the other algorithms in these 
		three experiments are the same as in \cite[Figure 
		4]{he2022solve}. }		
	\label{pcaa-am1}	
\end{figure}

\begin{figure}[h!]  
	\centering
    \subfloat[$n=100$]{ 
		\centering\includegraphics[width=150 pt, height =140 
		pt]{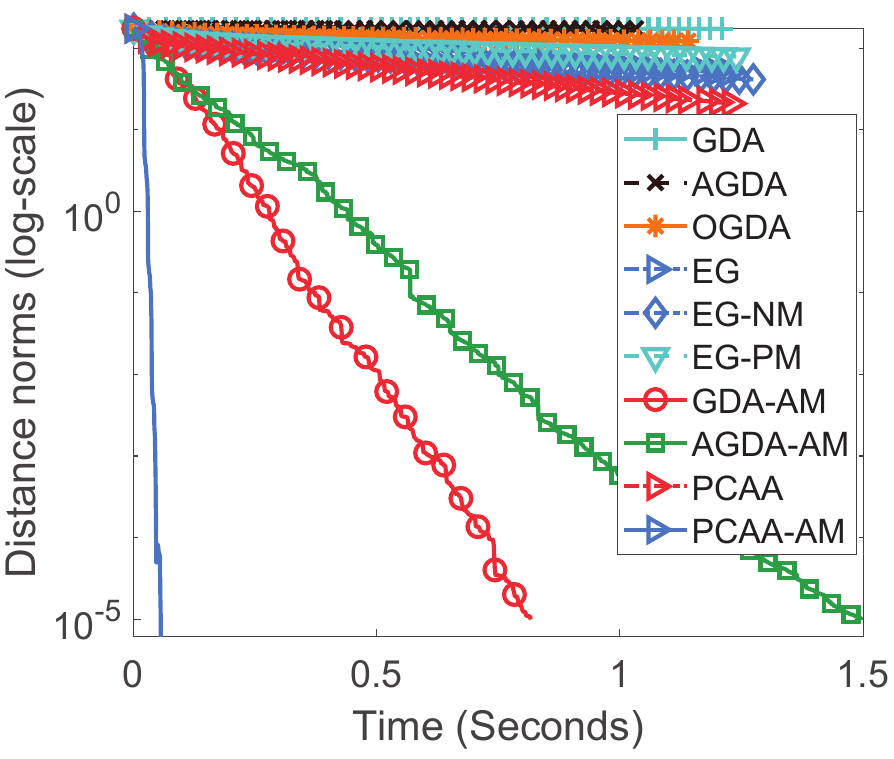}}
	\subfloat[$n=500$]{
	\includegraphics[width=150 pt, height =140 
		pt]{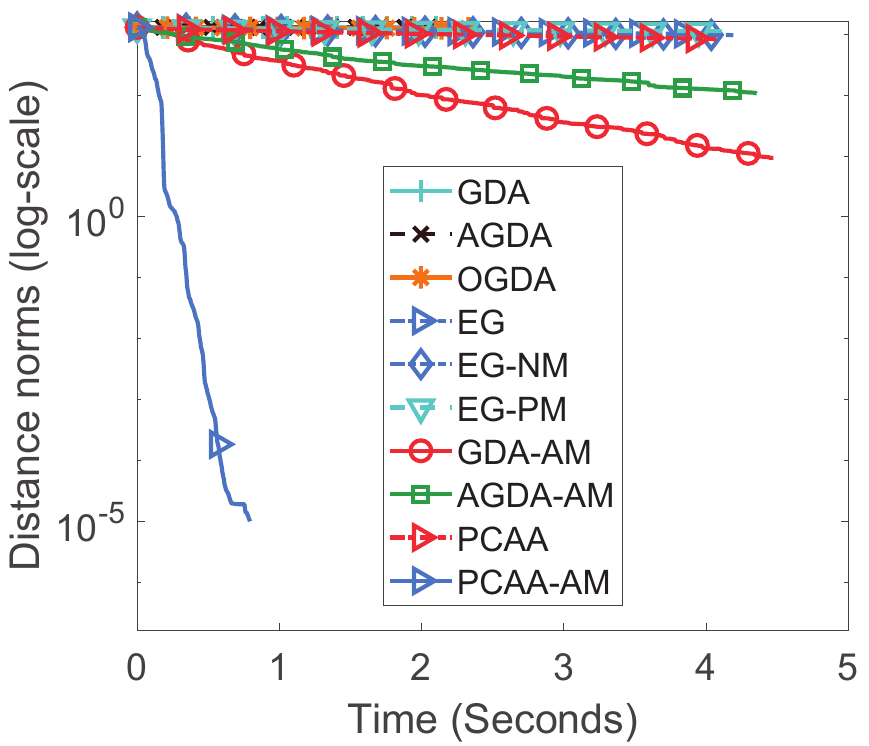}} 
	\subfloat[$n=1000$]{ \includegraphics[width=150 pt, height =140 
		pt]{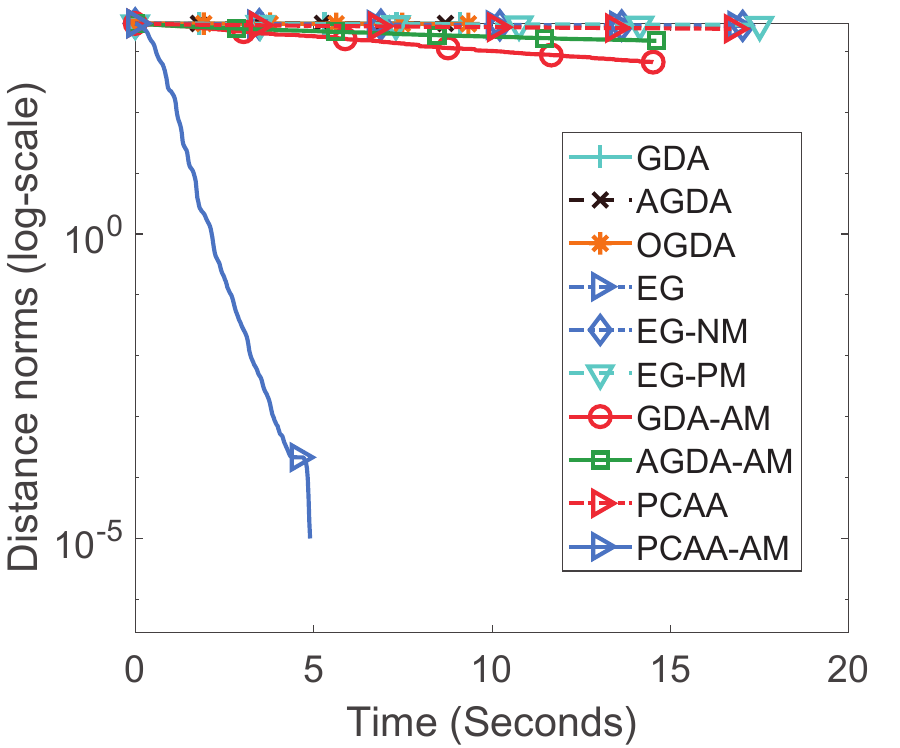}}
	\caption{Comparison between methods in terms of time. 
	}		\label{pcaa-am2}	
\end{figure}

Figure \ref{pcaa-am1} illustrates the convergence results of 
different 
algorithms in terms of the number of iterations for various problem 
sizes $(n=100, n=500, n=1000)$. 
From the figure, it is evident that PCAA-AM converges with fewer 
iterations compared to the other algorithms in the three different 
problem sizes. Moreover, OGDA, EG, EG-NM, EG-PM and PCAA also 
eventually converge, but they require more iterations compared to 
PCAA-AM. This further demonstrates the scalability of the PCAA 
framework and the effectiveness of Adeson Mixing. It also answers the 
question about the performance of PCAA-AM in terms of iteration 
number.  Figure \ref{pcaa-am2} depicts the performance of all methods 
in 
terms of runtime during convergence. Despite the slower speed of each 
iteration in PCAA-AM, its overall runtime is still faster than the 
other methods. This answers the question regarding the performance of 
PCAA-AM in terms of running time.

\begin{figure}[h!]  
	\centering
	\subfloat[$m=100, n=98$]{\includegraphics[width=151 pt, height 
	=140 
		pt]{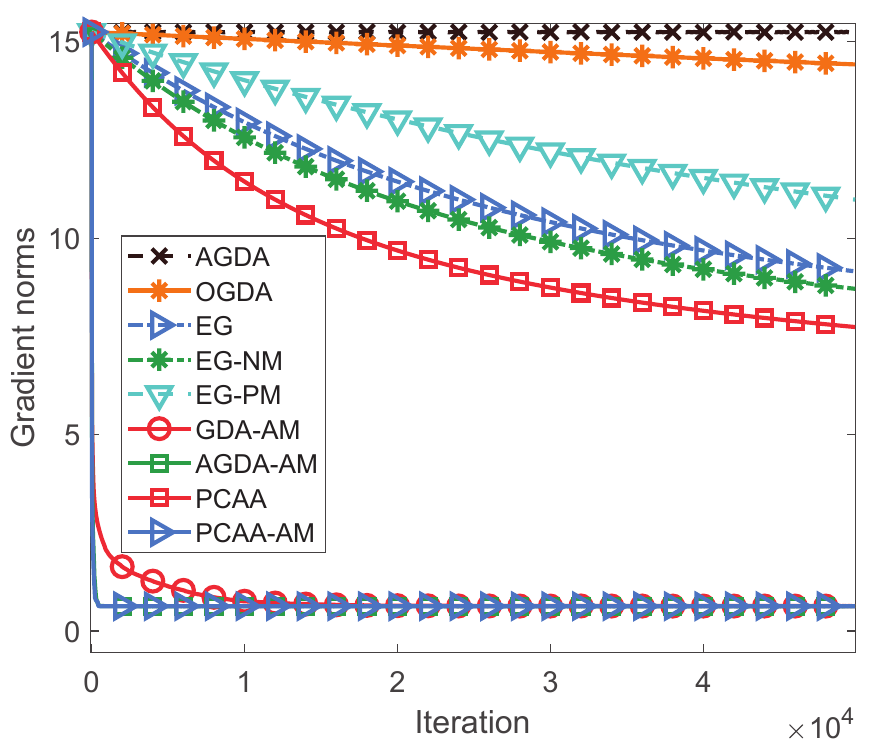}}
	\subfloat[$m=100, n=102$]{ 
		\includegraphics[width=151 pt, height =140 
		pt]{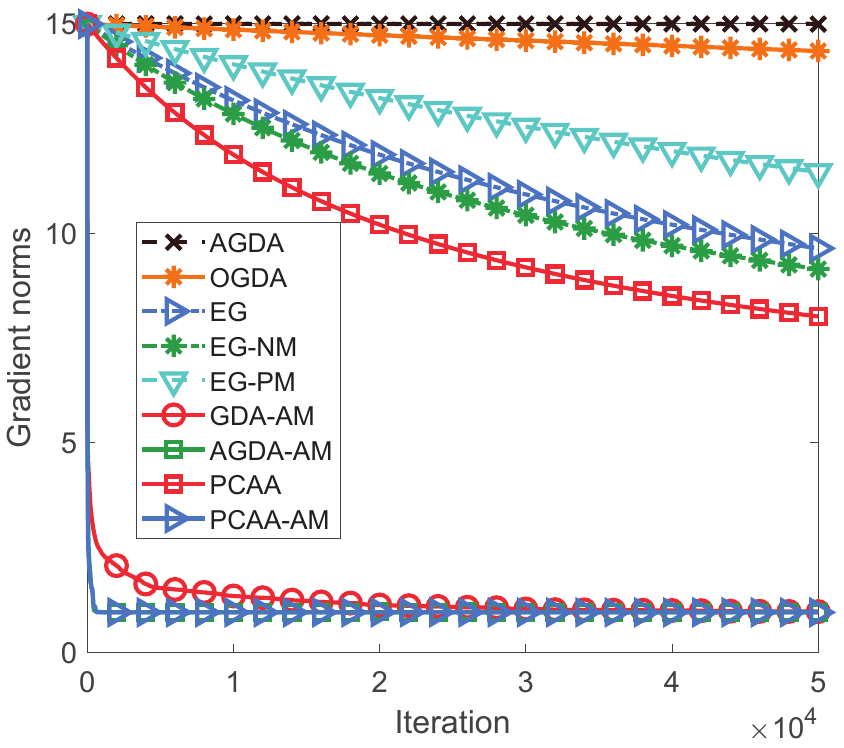}}
	\subfloat[$m=500, n=510$]{\includegraphics[width=151 pt, height 
	=140 
		pt]{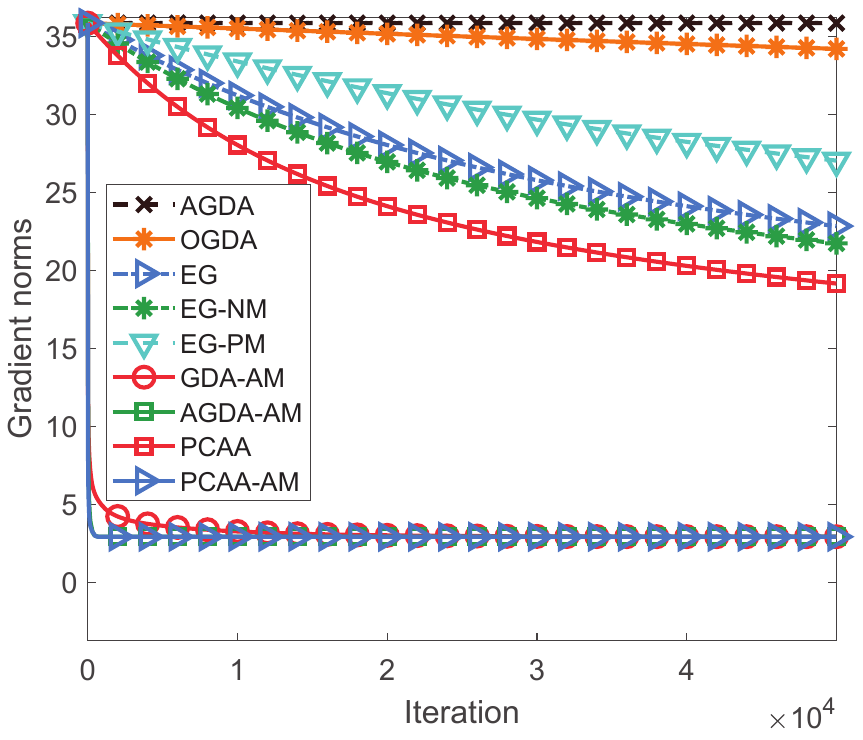}}
	\caption{Comparison in terms of iteration on games $\min 
		_{\theta} \max_{\phi} V(\theta,\phi)
		=	\theta^{\mathrm{T}} A \phi+\theta^{\mathrm{T}} 
		b+c^{\mathrm{T}} \phi, A \in \mathbb{R}^{m \times n}$. Except 
		for PCAA and PCAA-AM using 
		parameters $\gamma = 0.01, \alpha=0, \beta=0.01$ in 
		subfigures (a),(b) and (c), the parameter settings for the 
		other algorithms in these 
		three experiments are are all set to $0.01$. }		
	\label{pcaa-am3}	
\end{figure}

\begin{figure}[h!]  
	\centering
	\subfloat[$m=100, n=98$]{ 
		\centering\includegraphics[width=150 pt, height =140 
		pt]{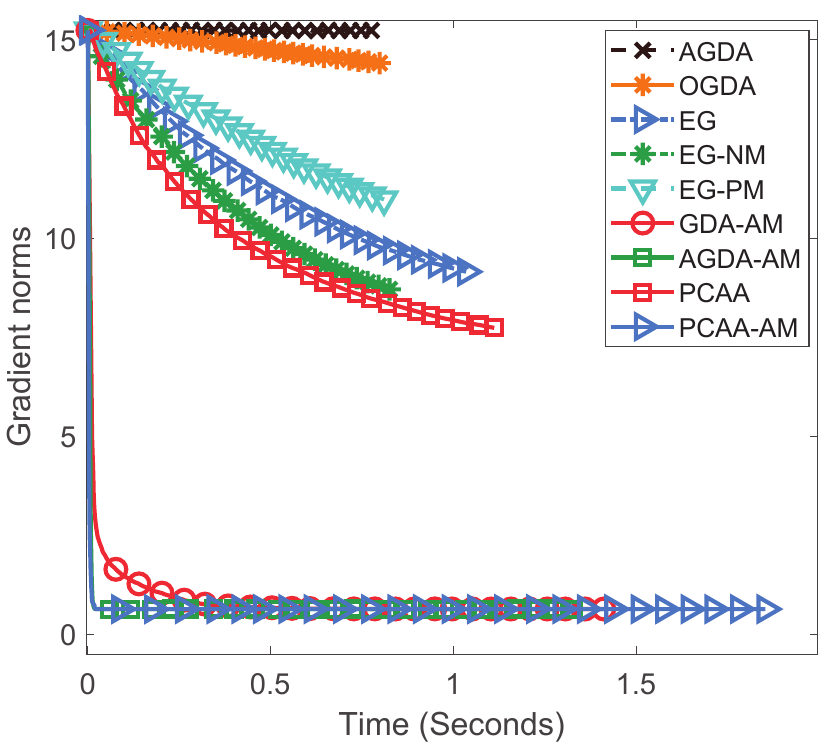}}
	\subfloat[$m=100,n=102$]{
		\includegraphics[width=150 pt, height =140 
		pt]{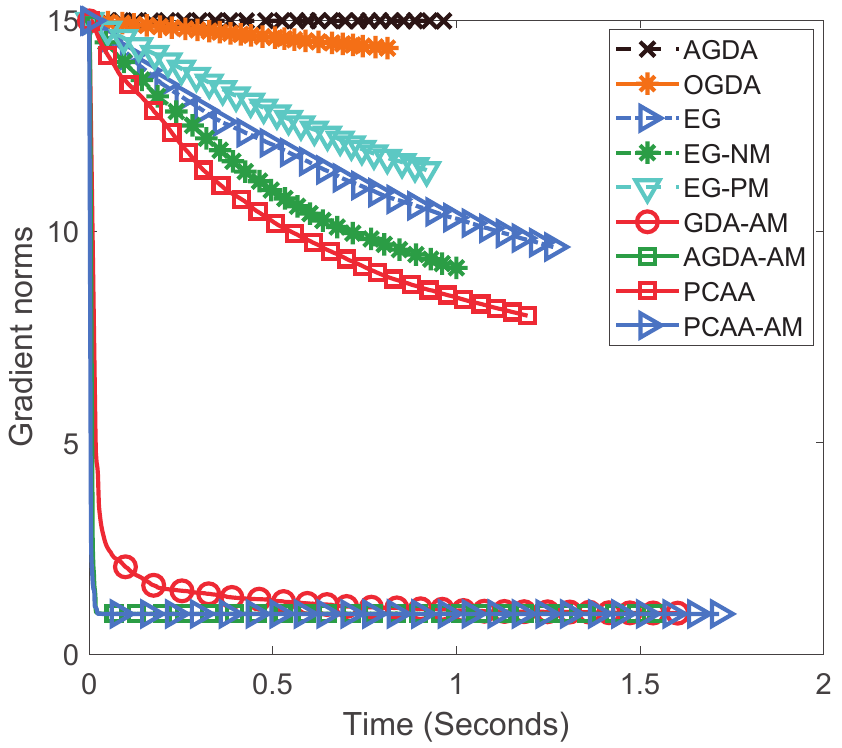}} 
	\subfloat[$m=500,n=510$]{ \includegraphics[width=150 pt, height 
	=140 
		pt]{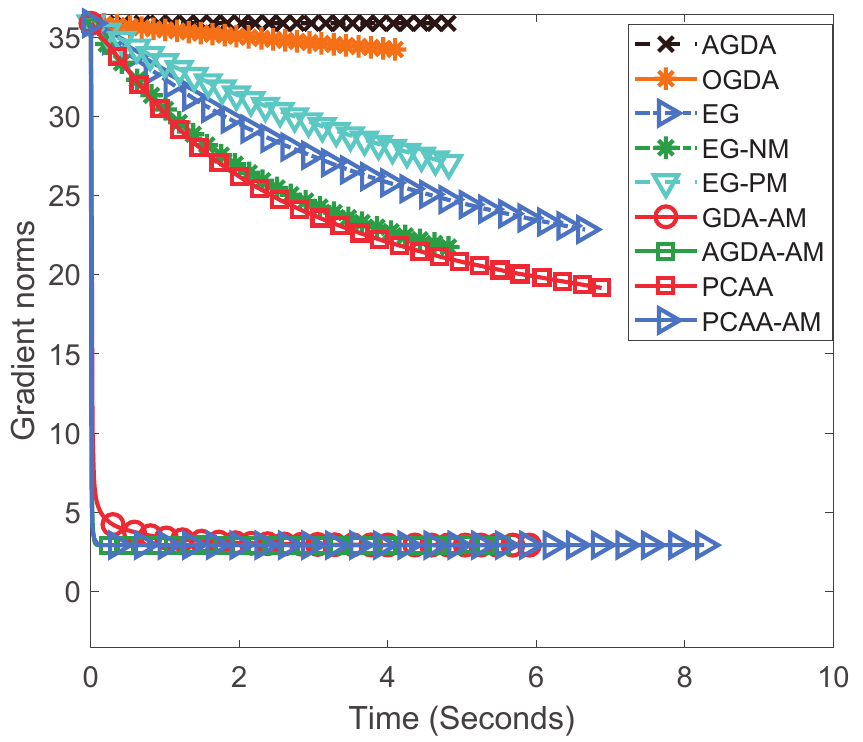}}
	\caption{Comparison between methods in terms of time. 
	}		\label{pcaa-am4}	
\end{figure}

Above, we compared the numerical performance of ten algorithms when 
the matrix $A$ is a full-rank square matrix in bilinear game $\min
_{\theta} \max_{\phi} V(\theta,\phi)
= \theta^{\mathrm{T}} A \phi+\theta^{\mathrm{T}}
b+c^{\mathrm{T}} \phi$. The results showed that 
combining PCAA with Anderson mixing is more effective for solving the 
game. Now, we continue to test the numerical performance of the ten 
algorithms on this bilinear game with matrix $A$ is not necessarily 
square. The experimental results are shown in Figure 
\ref{pcaa-am3} and Figure \ref{pcaa-am4}. It should be noted that 
according 
to Proposition \ref{proposition1}, the norm of the gradient of GDA 
iteration points increases continuously with the number of 
iterations. The experimental results in this example align with the 
theoretical analysis. However, plotting the graph of GDA together 
with the graphs of the other nine algorithms would obscure the 
differences in the latter. Therefore, in the following experiments, 
we did not include the norm graph of the GDA iteration points. 

The experimental results in Figure \ref{pcaa-am3} indicate that PCAA, 
without using Anderson mixing, achieves smaller norms of the 
gradients of the iteration points compared to AGDA, OGDA, EG, EG-NM
and EG-PM. This suggests that PCAA converges faster than AGDA, OGDA, 
EG, EG-NM and EG-PM in this case. However, the results in Figure 
\ref{pcaa-am4} show that PCAA takes relatively more time compared to 
the other five algorithms. When Anderson mixing is applied to GDA, 
AGD, and PCAA, we can compare the norms of the gradients of the 
iteration points of GDA-AM, AGDA-AM, and PCAA-AM, given the same 
number of iterations, based on the results presented in Figures 
\ref{pcaa-am3} and \ref{pcaa-am4}. It can be observed that PCAA-AM 
has smaller norms of the gradients of the iteration points compared 
to GDA-AM and AGDA-AM. We would like to add that although the results 
on the graph may appear to overlap for AGDA-AM and PCAA-AM, in 
reality, the norm values of the gradients obtained by PCAA-AM are 
slightly smaller. However, as shown in the time comparison graph in 
Figure \ref{pcaa-am4}, PCAA-AM takes slightly more time compared to 
GDA-AM and AGDA-AM.

\subsection{Wasserstein GAN-GP on CelebA} \label{WGAN-GP6.2}
\indent
\par
It should be noted that after the proposal of Extra-Adam in 
\cite{mertikopoulos2019optimistic}, it was tested using GANs 
architecture on the CelebA dataset. The effectiveness of the proposed 
algorithm in training GANs was illustrated by showing improved FID 
(Fr\'{e}chet inception distance) values compared to Adam. 
On the other hand, based on the analysis of the algorithm framework 
in subsection \ref{PCAA-Adamp}, does the empirical evidence support 
our theoretical analysis? In light of this, we conducted numerical 
experiments on the CelebA dataset to demonstrate the effectiveness of 
the algorithm. Additionally, through practical experiments, we aim to 
validate our theoretical analysis.

{\bf Experiment 1: Wasserstein GAN-GP on CelebA.}
Next, we trained Wasserstein GAN-GP using the proposed algorithm on 
the CelebA dataset to demonstrate the effectiveness of the algorithm, 
as evidenced by the FID scores. This serves as an answer to the 
aforementioned question two in subsection \ref{PCAA-Adamp}.
The network 
architecture of the Wasserstein GAN is presented in Table 
\ref{WGAN1}, 
which has been trained on the CIFAR10 dataset for GANs training in 
\cite{gidel2019a}. 
The specific experimental setup can be found in Table
\ref{experement1}, and the hyperparameter 
settings for the comparative algorithms are provided in Table
\ref{hyper1}. 

\begin{table}[h!]\footnotesize\tabcolsep 16pt
	\begin{center}
	\caption{ResNet architecture used for our CelebA 
	experiment}\label{WGAN1}\vspace{-2mm}
	\end{center}
	\begin{center}
	\begin{tabular}{c}
		\toprule   \bf Generator \\ 
		\midrule Input: $z \in \mathbb{R}^{128} \sim 
		\mathcal{N}(0, I)$ \\
		Linear $128 \rightarrow 128 \times 4 \times 4$ \\
		ResBlock \quad $128 \rightarrow 128$ \\
		ResBlock \quad $128 \rightarrow 128$ \\
		ResBlock \quad $128 \rightarrow 128$ \\
		Batch Normalization \\
		${\rm ReLU} (\cdot)$ \\
		conv. (kernel: $3 \times 3,128 \rightarrow 3$, stride: 
		1 , pad: 1 ) \\
		${\rm Tanh} (\cdot)$ \\
		\midrule \bf Discriminator \\
		\midrule Input: $x \in \mathbb{R}^{3 \times 32 \times 32}$ \\
		ResBlock \quad $~~~3 \rightarrow 128$ \\
		ResBlock \quad $128 \rightarrow 128$ \\
		ResBlock \quad $128 \rightarrow 128$ \\
		ResBlock \quad $128 \rightarrow 128$ \\
		~Linear  \quad $128 \rightarrow 1$ \\
		\bottomrule
	\end{tabular} 
	\end{center}
\end{table}

\begin{table}[h!]\footnotesize\tabcolsep 16pt
	\begin{center}
	\caption{Experimental environment on the CelebA 
	dataset}\label{experement1}\vspace{-2mm}
    \end{center}
    \begin{center}
	\begin{tabular}{c c c c}
		\toprule   
		\multicolumn{4}{c}{\bf Experimental environment}           \\
		\midrule   
		\qquad CPU:      \quad   & Inter Core i7-10700    
		&\qquad GPU:      \quad   & RTX  A4000  16G          \\
		\qquad RAM:      \quad   & 64 G  
		&\qquad Python:   \quad   & Version 3.8.0    \\    
		\qquad PyTorch: \quad   & Version 1.9.0       \\
		\bottomrule  
	\end{tabular} 
	\end{center}
\end{table}

\begin{table}[h!]\footnotesize\tabcolsep 16pt
	\begin{center}
	\caption{(ResNet) WGAN-GP 
	hyperparameters}\label{hyper1}\vspace{-2mm}
    \end{center}
    \begin{center}
	\begin{tabular}{c c}
		\toprule 
		\multicolumn{2}{c}{ Hyperparameters settings} \\
		\midrule
		Batch size & $64$ \\
		Max iterations T  & $200,000$ \\
		Adam $\beta_{1}$ & $0.5$ \\
		Adam $\beta_{2}$ & $0.9$ \\
		Gradient penalty & $10$  \\
		GAN objective     &  Wasserstein GAN-GP \\
		Learning rate for generator    & $\gamma =2 
		\times 10^{-5}$  (for Adam, Extra-Adam)\\
		\qquad\qquad\qquad\,    & $\gamma=\beta=2 \times 
		10^{-5}, \; \alpha=2.1\times 10^{-5}  $ (for PCAA-Adam) \\
		Learning rate for discriminator   & $\gamma =2 
		\times 10^{-5}$ (for Adam, Extra-Adam)\\ 
		\qquad\qquad\qquad\,   & $\gamma=\beta=2 \times 
		10^{-5}, \; \alpha=2.1\times 10^{-5}  $ (for PCAA-Adam) \\
		\bottomrule
	\end{tabular} 
	\end{center}
\end{table} 

\begin{figure}[h!]
	\centering\includegraphics[width=3.3 in]{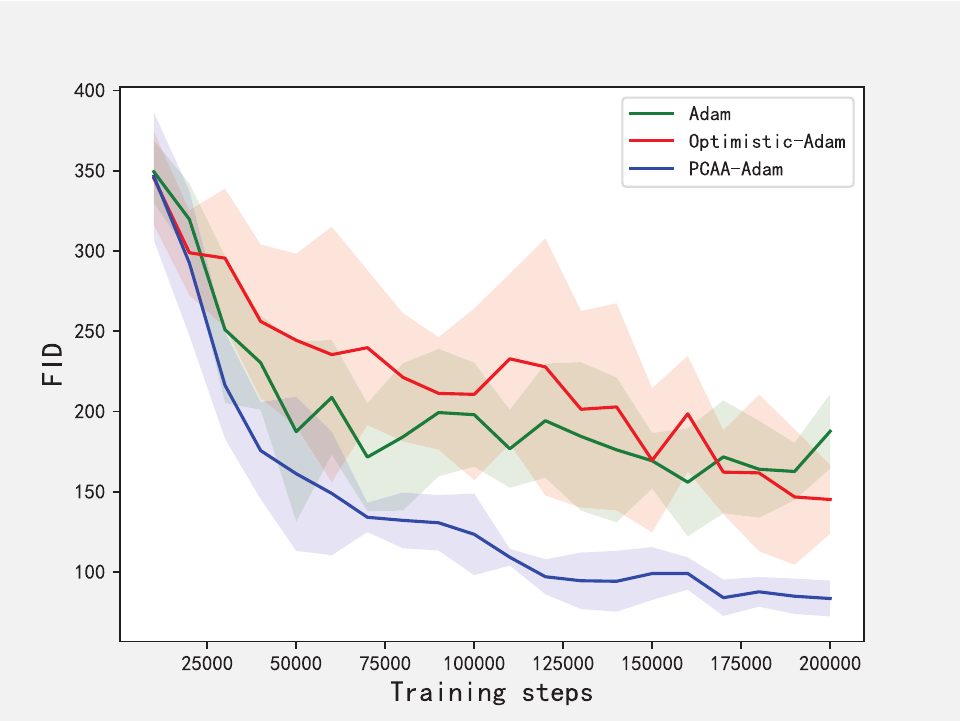}
	\caption{Mean and standard deviation of the FID computed over 5 
		runs for each method on Wasserstein GAN-GP trained on  CelebA.
	}\label{fid1}
\end{figure}

The comparison graph of FID values obtained during the training 
process using Adam, Extra-Adam, and PCAA-Adam algorithms shows that 
Wasserstein GAN-GP trained with PCAA-Adam consistently achieve 
smaller FID values 
in the middle to later stages of training. This not only demonstrates 
the effectiveness of the PCAA-Adam algorithm but also supports our 
analysis of the algorithm framework. Moreover,
The results of Experiment 1 indicate that introducing the concept of 
centripetal acceleration to Adam can stabilize the training process 
and achieve smaller FID values. This suggests that the trained GANs 
have distributions that are closer to the real data distribution.

{\bf Experiment 2: Wasserstein GAN-GP with different network 
architectures on 
	CelebA dataset.}
The purpose of this experiment is the same as experiment 1. The 
difference is that we have increased the number of pixels, making the 
differences in FID values more apparent through the clarity of the 
generated samples.
The network 
architecture of the Wasserstein GAN is presented in Table \ref{WGAN2},
the specific experimental setup can be found in 
Table \ref{experement1},
and the 
hyperparameter 
settings for the comparative algorithms are provided in 
Table \ref{hyper2}. 

\begin{table}[h!]\footnotesize\tabcolsep 16pt
	\begin{center}
	\caption{ResNet architecture used for our CelebA 
	experiment}\label{WGAN2}\vspace{-2mm}
	\end{center}
	\begin{center}
	\begin{tabular}{c}
		\toprule   \bf Generator \\ 
		\midrule Input: $z \in \mathbb{R}^{128} \sim 
		\mathcal{N}(0, I)$ \\
		Linear $128 \rightarrow 512 \times 8 \times 8$ \\
		ResBlock \quad $512 \rightarrow 256$ \\
		ResBlock \quad $256 \rightarrow 128$ \\
		ResBlock \quad $128 \rightarrow 64$ \\
		Batch Normalization \\
		${\rm ReLU} (\cdot)$ \\
		conv. (kernel: $3 \times 3 $, $64 \rightarrow 3$, 
		stride: 
		1 , pad: 1 ) \\
		${\rm Tanh} (\cdot)$ \\
		\midrule \bf Discriminator \\
		\midrule Input: $x \in \mathbb{R}^{3 \times 64 \times 64}$ \\
		ResBlock \quad $~~~3 \rightarrow 128$ \\
		ResBlock \quad $128 \rightarrow 128$ \\
		ResBlock \quad $128 \rightarrow 256$ \\
		ResBlock \quad $256 \rightarrow 512$ \\
		~Linear  \quad $512 \rightarrow 1$ \\
		\bottomrule
	\end{tabular} 
	\end{center}
\end{table}

\begin{table}[h!]\footnotesize\tabcolsep 16pt
	\begin{center}
	\caption{(ResNet) WGAN-GP 
	hyperparameters}\label{hyper2}\vspace{-2mm}
    \end{center}
    \begin{center}
	\begin{tabular}{c c}
		\toprule 
		\multicolumn{2}{c}{ Hyperparameters settings} \\
		\midrule
		Batch size & $64$ \\
		Max iterations T  & $200,000$ \\
		Adam $\beta_{1}$ & $0.0$ \\
		Adam $\beta_{2}$ & $0.9$ \\
		Gradient penalty & $10$  \\
		GAN objective     &  Wasserstein GAN-GP \\
		Learning rate for generator    & $\gamma =2 
		\times 10^{-4}$  (for Adam, Extra-Adam)\\
		\qquad\qquad\qquad\,    & $\gamma=\beta=2 \times 
		10^{-4}, \; \alpha=2.4\times 10^{-4}  $ (for PCAA-Adam) \\
		Learning rate for discriminator   & $\gamma =2 
		\times 10^{-4}$ (for Adam, Extra-Adam)\\ 
		\qquad\qquad\qquad\,   & $\gamma=\beta=2 \times 
		10^{-4}, \; \alpha=2.4\times 10^{-4}  $ (for PCAA-Adam) \\
		\bottomrule
	\end{tabular}
	\end{center}
\end{table} 

\begin{figure}[h!]  
    \subfloat[]{\centering
    	\includegraphics[width=228 pt, height =200 
		pt]{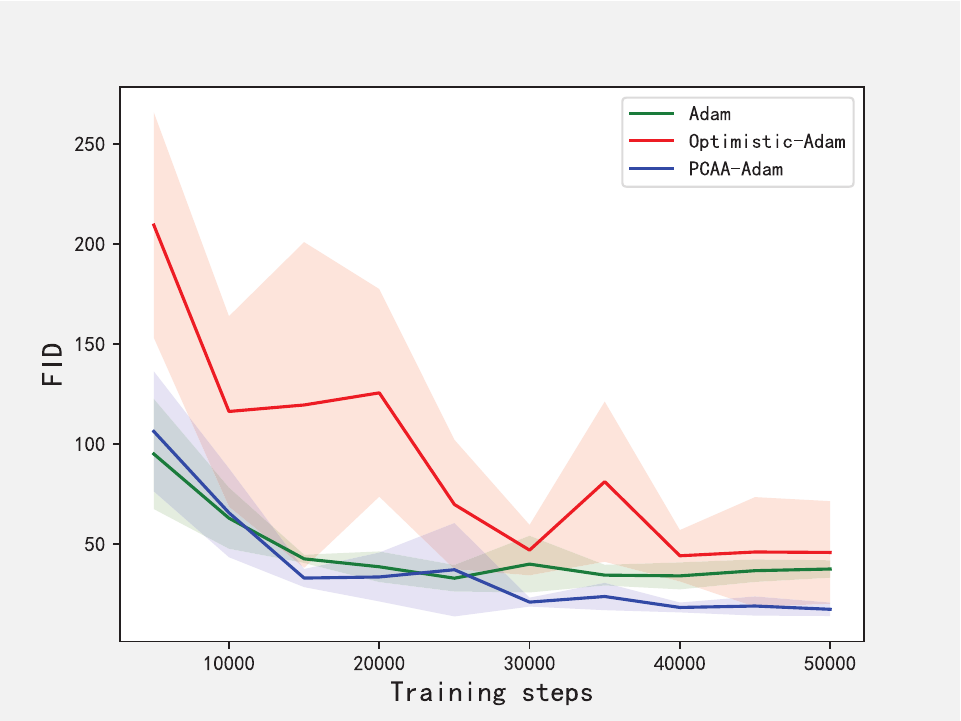}} 
	\subfloat[]{\centering
		\includegraphics[width=228 pt, height =200 
		pt]{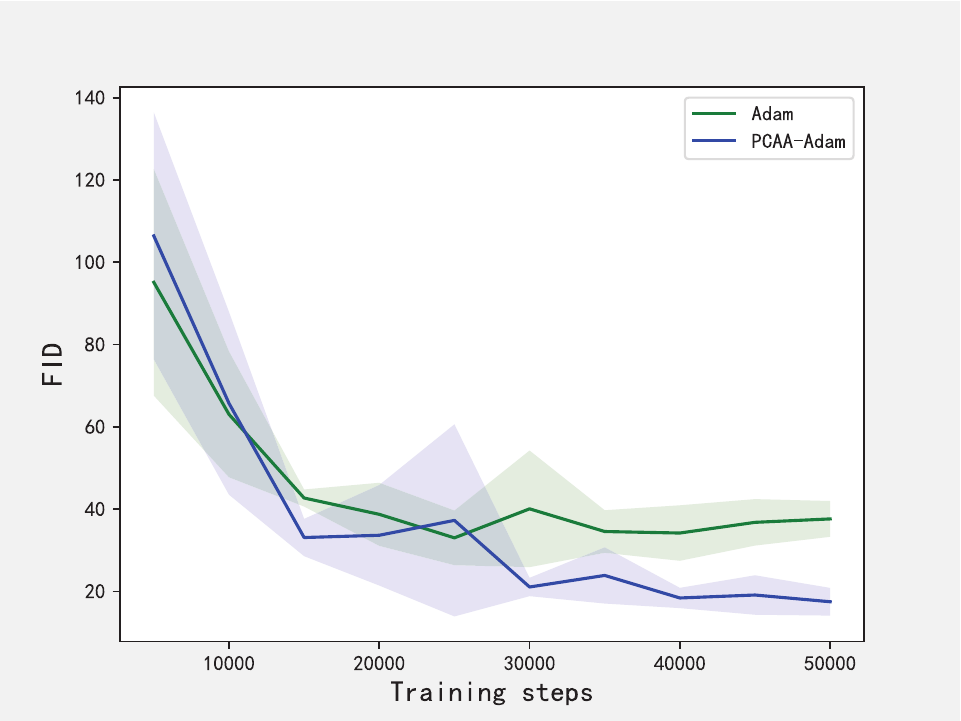}}
	\caption{Mean and standard deviation of the FID computed over 5 
		runs for each method on Wasserstein GAN-GP trained on  
		CelebA. (a): Comparison of FID values obtained from a 
		single training 
		process of Adam, Extra-Adam, and PCAA-Adam. (b):
		Comparison of FID values obtained from separate training 
		processes of Adam and PCAA-Adam in the left figure. Due to 
		the 
		significant fluctuations in the FID values obtained from 
		Extra-Adam in the left figure, it diminishes the visual 
		effect of 
		the FID values obtained from Adam and PCAA-Adam. Therefore, 
		in 
		the right figure, a separate comparison of the FID values 
		obtained from the training processes of Adam and PCAA-Adam is 
		presented.
	}		\label{pcaa-fid2}	
\end{figure}

\begin{figure}[h!]
	\centering	
	\subfloat[PCAA-Adam for CelebA]{ 
		\includegraphics[width=240 pt, height =240 
		pt]{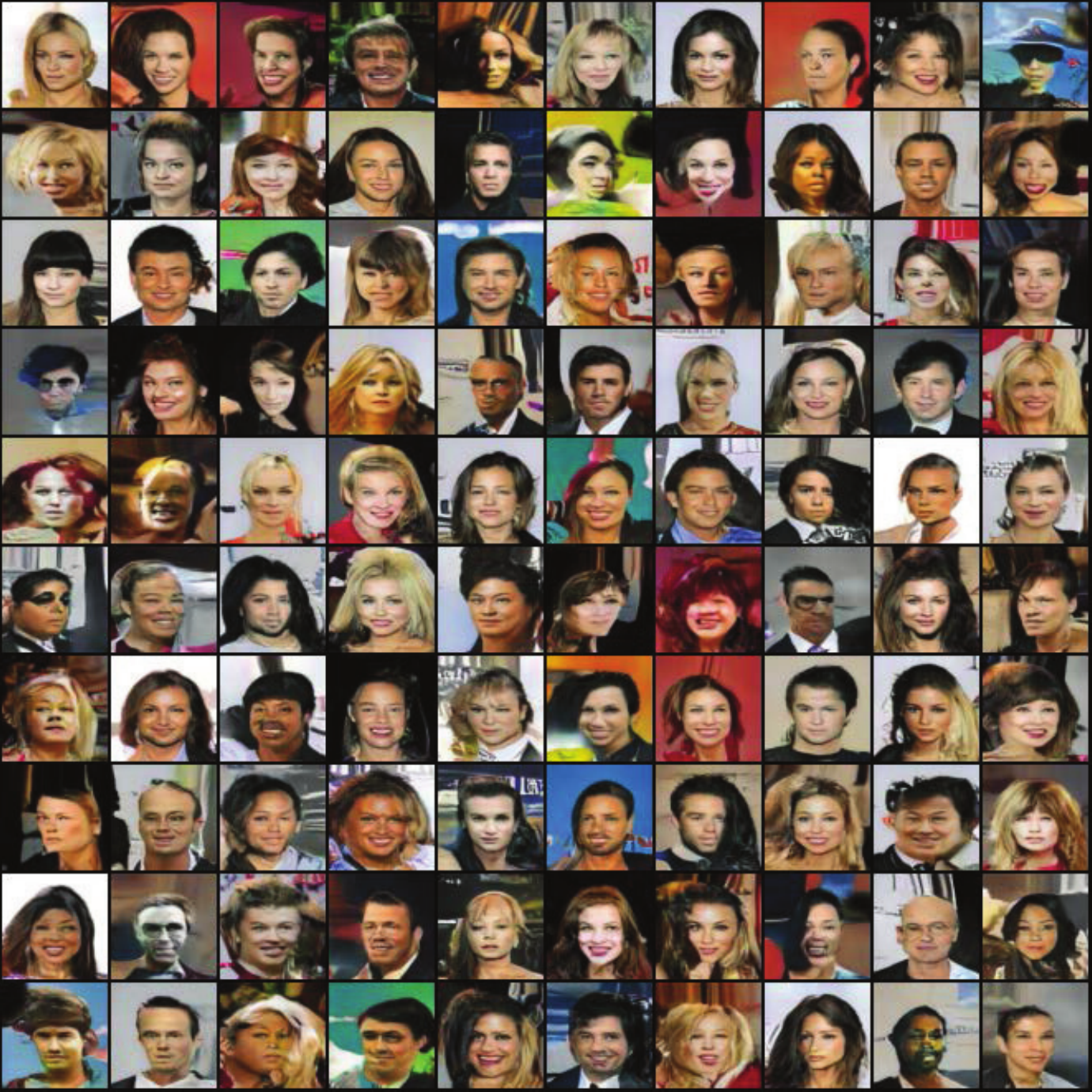}} \\
	\subfloat[Adam for CelebA]{ 
	\includegraphics[width=220 pt, height =240 
		pt]{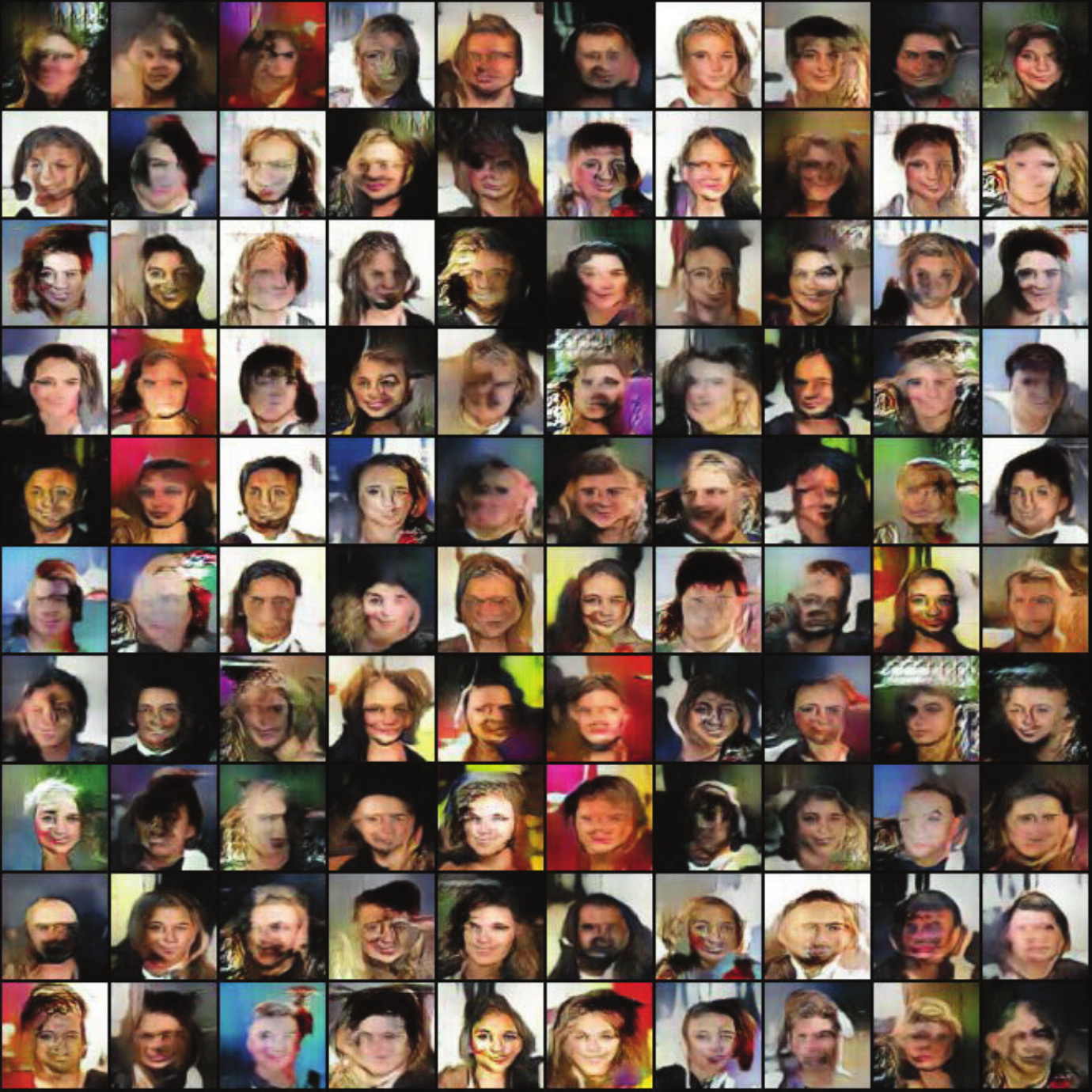}}
	\subfloat[Optimistic-Adam for CelebA]{ 
		\includegraphics[width=220 pt, height =240 
		pt]{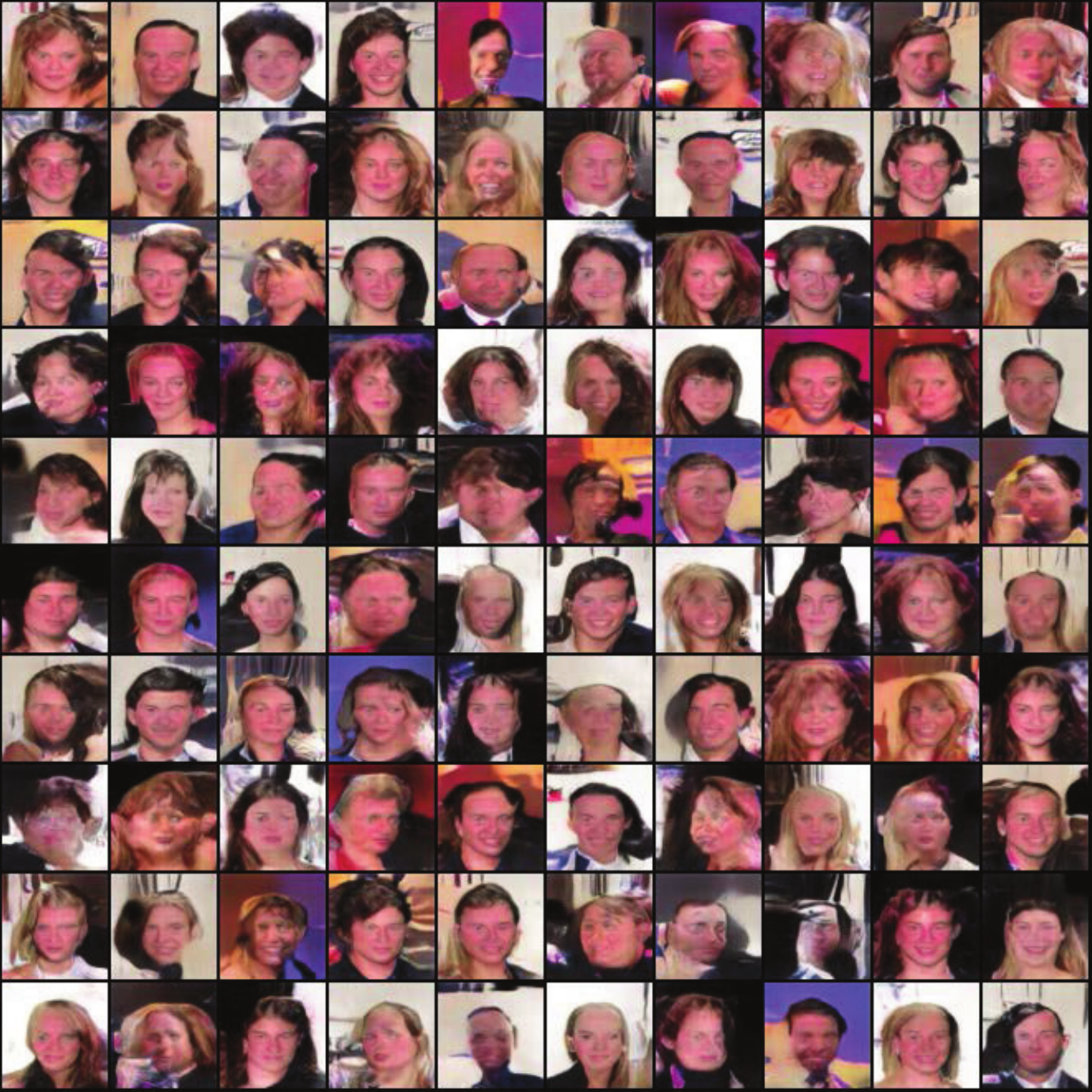}} 
	\caption{The figure shows a comparison of generated images by the 
		generator at the end of training the Wasserstein GAN using 
		Adam, Extra-Adam and PCAA-Adam.
	}		\label{pcaa-fig1}	
\end{figure}

The FID comparison graph Figure \ref{pcaa-fid2} reveals that after a 
fluctuation in the mid-training phase, the FID value obtained from 
Adam continues to increase, while the FID values obtained from 
Extra-Adam and PCAA-Adam initially increase and then rapidly 
decrease. This demonstrates that the introduction of the extra 
gradient technique and the centripetal acceleration idea can enhance 
the robustness of the FID values obtained during training. Comparing 
the FID values obtained from Extra-Adam and PCAA-Adam, it is observed 
that after the fluctuation, the FID value obtained from Extra-Adam 
exhibits a trend of increasing and then decreasing. However, the FID 
value obtained from PCAA-Adam continues to decrease. This indicates 
that the incorporation of the centripetal acceleration idea, which 
involves incorporating gradient-related information from the 
prediction step into the update step, can improve the practical 
performance of Extra-Adam. It also demonstrates the effectiveness of 
our algorithm. Figure  \ref{pcaa-fig1} depicts a comparison of the 
generated images by the generator at the end of training for Adam, 
Extra-Adam, and PCAA-Adam. The smaller the FID value, the clearer the 
generated images. PCAA-Adam achieves the smallest FID value and 
generates the clearest images, thus visually demonstrating the 
effectiveness of the PCAA-Adam algorithm.

\section{Conclusion}
In response to the open problem of how to eliminate limit cycle 
behavior in zero-sum games using optimization techniques, as proposed 
in \cite{hsieh2020convergence}, and combining the research ideas of 
GANs training algorithms. This paper first demonstrated the fact that 
GDA does not converge on the general bilinear game through geometric 
and convergence rate analysis.
Moreover, it proved that the limit point 
of PCAA iteration sequence was a Nash equilibrium point. Then, we 
obtained the last-iterate linear convergence rate of PCAA on the 
general bilinear game, which outperformed the results obtained in the 
comparative literature. Finally, we validated the effectiveness of 
our proposed PCAA-Adam algorithm for training GANs through practical 
experiments.

Nevertheless, training general GANs involves the use of optimization 
algorithms and deep learning frameworks to tackle a non-convex 
non-concave zero-sum game. The current research status in the 
international community reveals that both theoretical and algorithmic 
aspects of non-convex non-concave zero-sum games encompass various 
important and challenging problems that require attention 
\cite{xuzifei2020}. Additionally, the stability of training processes 
for many GAN variants is highly intricate and represents an area for 
further investigation in our research. 
Furthermore, recently Chae et al. \cite{Chae2023open, 
Chae2023two} has shown that two-timescale EG holds the 
potential to address the open problem studied in their paper. 
However, their results do not yet guarantee that the limit points of 
the algorithm's iterates are Local minimax optima. Therefore, they 
invite the academic community to propose new algorithms or more 
reasonable concepts of Local minimax optima to address the unresolved 
issues. Since PCAA can reduce to EG, can PCAA provide some insights 
into solving the open problem of their interest? And can PCAA almost 
surely avoid strict non-minimax points? These questions are all 
waiting to be further explored.


\Acknowledgements{This work was supported by Major Program of 
National Natural Science 
	Foundation of China (Nos. 11991020, 11991024). }




\end{document}